\documentclass[letterpaper]{article}
\usepackage[preprint]{aaai2027}
\usepackage[hyphens]{url}
\usepackage{graphicx}
\usepackage{natbib}
\usepackage{mathtools}
\usepackage{amssymb}
\usepackage{booktabs}
\usepackage{multirow}
\usepackage{amsthm}

\newtheorem{theorem}{Theorem}

\newtheorem{proposition}[theorem]{Proposition}

\title{KG-SoftMAP: Soft Knowledge-Graph Priors for Bayesian Network Structure Learning from Sparse Discrete Data}

\author{Guoliang Xu \quad James E.~Corter}
\affiliations{Columbia University}

\begin{document}
\maketitle

\begin{abstract}
Learning Bayesian network (BN) structure from sparse discrete data is hard: when each instance records only a few variables, most variable pairs lack the joint observations needed for reliable scoring, and data-only methods recover little structure.
Imperfect domain knowledge, expressible as a weighted, directed knowledge graph (KG), is often available.
KG-SoftMAP encodes such a KG as a \emph{finite-strength, confidence-weighted} edge prior and maximizes a MAP objective that adds this logit-form prior to the BDeu score.
With an informative but imperfect KG, KG-SoftMAP recovers partial directed structure even at an observation rate of $\rho{=}0.05$ (directed F1 (DF1) of $0.19$--$0.32$ across benchmarks).
At higher observation rates within this sparse grid, DF1 reaches $0.44$--$0.66$ at $\rho{=}0.20$ and $0.46$--$0.64$ at $\rho{=}0.40$.
Across the same three rates, KG-SoftMAP without the KG prior averages DF1 $0.00$, $0.19$, and $0.21$.
Stress tests that corrupt, remove, or blur the KG signal, together with checks on LLM-extracted graphs beyond canonical benchmarks, show that recovery rises and falls with KG quality.
On three real sparse educational datasets without ground-truth DAGs, we evaluate prediction, calibration, and KG-consistency.
On Short Answer Feedback (SAF), KG-SoftMAP+VE reaches F1\textsubscript{FAIL} (F1 for the \textsc{Fail} class) $0.75$ versus $0.78$ for logistic regression while also providing an inspectable concept graph, calibrated \textsc{Fail} probabilities, and posterior queries from partially observed concept evidence.
The remaining datasets sharpen the operating picture: weak heuristic KG signal leaves prediction unchanged, while an independent expert ontology moves the learned graph toward expert relatedness.
\end{abstract}
\section{Introduction}\label{sec:intro}

A Bayesian network (BN) is a directed acyclic graph (DAG) whose nodes are variables and whose edges encode direct dependencies \citep{pearl1988probabilistic}, with applications across genetics \citep{friedman2000using}, medicine \citep{mani1999causal}, and manufacturing \citep{li2007knowledge}.
Learning the DAG from data is a core but NP-hard problem \citep{chickering2004large}, addressed by two main approaches: constraint-based methods that build the graph from conditional-independence tests \citep{spirtes2000causation,colombo2014order}, and score-based methods that search for the DAG maximizing a score such as BDeu \citep{heckerman1995learning,cooper1992bayesian}.
To tame the search space, hybrid and sparsity-based methods restrict candidate parents \citep{tsamardinos2006max,friedman1999learning,huang2013sparse}, and recent work recasts the problem as continuous optimization \citep{zheng2018dags,ng2020golem,bello2022dagma}.
All of these determine the graph solely from the observed data.

In many domains, however, that data is \emph{sparse}: each instance records only a small subset of variables, so the data matrix is mostly missing.
Such sparsity appears in educational assessments, clinical registries, and sensor networks; in our SAF case study, the observation rate, the fraction of matrix entries observed and denoted $\rho$, is about $4.4\%$.
Under such sparsity, few variable pairs share enough joint observations for reliable scoring or independence testing, and classical and modern data-only methods alike recover little structure.

Useful knowledge about causal, prerequisite, or implication-like variable relations often exists outside the data and can be written as a weighted, directed knowledge graph (KG): experts may know which skills build on others or which symptoms relate to which diseases.
Existing approaches differ in how much trust they place in this knowledge.
Hard constraints fix or forbid edges \citep{meek1995causal}, which is brittle when the knowledge is itself uncertain.
Recent LLM-based approaches elicit graph structure from a language model \citep{zhang2025promptbn,srivastava2025sciencegrounded}; when used directly, the final structure inherits model errors without statistical vetting against sparse data.
Domain-specific priors exist \citep{li2007knowledge,wu2011altered}, but they are engineered for particular domains.
We instead seek a general mechanism by which any weighted KG can enter a sparse discrete BN score as bounded, confidence-weighted evidence.

KG-SoftMAP maximizes a MAP objective that combines the BDeu score with an edge-factored KG prior: each edge's inclusion log-odds is a linear function of its KG confidence weight, so the KG contributes bounded evidence rather than hard constraints.
An optional first stage constructs the KG from domain reference text via LLM-based extraction, so the method supports both curated and extracted KGs.
We test the claim that soft KG priors improve sparse discrete BN structure learning when the KG is meaningful but imperfect.
To our knowledge, KG-SoftMAP is the first MAP BN learner to explicitly couple a weighted KG prior with family-wise available-case BDeu for the family-sparse regime, where each candidate parent set may have only a handful of jointly observed cases.

We evaluate this claim in two tracks.
In the synthetic track, where ground-truth DAGs are known, data-only baselines produce near-zero directed F1 (DF1) at the sparsest masks and recover only limited directed structure as $\rho$ grows.
KG-SoftMAP keeps measurable directed recovery across all three sparse rates, while the no-prior and data-only rows remain lower or isolated; ablations isolate the KG prior as the dominant factor (Section~\ref{sec:track1}).
LLM-provenance checks address a specific confound: an LLM may have memorized canonical benchmarks.
Standard-benchmark elicitation provides a sanity check, and NovelGraphs~\citep{srivastava2025sciencegrounded} extends the test to graphs outside canonical BN repositories, where benchmark memorization is less plausible.
In the real-data track, we study three educational datasets spanning different KG-provenance regimes (Section~\ref{sec:track2}).
On SAF, the BN reaches F1\textsubscript{FAIL} $0.75$ versus $0.78$ for logistic regression while providing structure, calibration, and posterior querying that fixed discriminative predictors do not.

The paper makes three contributions:

\begin{itemize}
\item \textbf{A sparse KG-MAP scoring framework.}
We make soft external graph knowledge operational in the regime where so few cases are jointly observed per candidate family that data-only scores cannot discriminate parent sets.
The technical coupling is the key object: the KG supplies finite log-odds evidence, while family-wise available-case BDeu scores each candidate parent set on the rows where the child and proposed parents are jointly observed (Section~\ref{sec:bn}).
This family-specific scoring is not a cosmetic choice; it is the count scale on which the sparse local likelihood is defined and avoids the global-sample-size BIC inconsistency identified under heterogeneous MCAR observation rates~\citep{balov2013consistent}.
A local score analysis interprets how the logit term shifts individual edge decisions under sparse complete cases; we present it as interpretation, not a global recovery guarantee.

\item \textbf{A boundary characterization of finite-strength KG priors.}
We test when the prior helps, when it hurts, and which component transfers across search routines.
The characterization highlights the non-obvious operating rules: losing or misorienting true KG edges hurts more than adding extra false edges, over-dense KGs can still help when they retain useful recall, separated confidence bands are not required, and zero-signal KGs return to the no-prior regime.
Same-KG checks further show that the sparse KG-MAP scorer, rather than one specific search routine, is the portable component.
Provenance checks with blindly LLM-elicited KGs, including NovelGraphs~\citep{srivastava2025sciencegrounded}, show the same mechanism operating on graph signal extracted from text rather than generated from the true DAG.

\item \textbf{A posterior-query evaluation across KG-provenance regimes.}
We design a three-dataset evaluation spanning LLM-extracted, weak-heuristic, and expert-ontology KGs.
On SAF, KG-SoftMAP+VE reaches F1\textsubscript{FAIL} $0.75$ versus $0.78$ for logistic regression and supports calibrated posterior queries from partial evidence; ASSISTments confirms the predicted weak-KG boundary, and Eedi shows that an independent expert ontology improves structural relatedness more than accuracy.
\end{itemize}

\section{Related Work}\label{sec:related}

\subsection{Data-Only BN Structure Learning Under Sparsity}\label{sec:related_classical}

Classical BN structure learning estimates the graph from data alone.
Score-based methods search for a DAG that maximizes a decomposable score such as BDeu~\citep{heckerman1995learning,cooper1992bayesian} or BIC~\citep{schwarz1978estimating}; greedy hill-climbing remains a standard approximate search strategy~\citep{chickering2002optimal}.
Constraint-based methods such as PC~\citep{spirtes2000causation,colombo2014order} remove edges by conditional-independence testing, while hybrid methods such as MMHC~\citep{tsamardinos2006max} use tests to restrict a later score-based search.
Sparse Candidate~\citep{friedman1999learning} similarly narrows the parent search space.
These methods are effective when enough joint observations are available, but sparse discrete matrices leave many child-parent families with too few complete cases for reliable scoring or testing.

Missing-data BN learning usually relies on Structural EM~\citep{friedman1998bayesian}, bounded-treewidth variants~\citep{scanagatta2018learning}, or data augmentation~\citep{fernandez2010learning}.
These methods recover information through model-based imputation, but the imputation model itself must be estimated from co-observed cases; under family-level missingness with observation rates below 5\%, that signal is largely absent.
More recent data-only search methods give two useful reference points.
GRaSP~\citep{lam2022grasp} and BOSS~\citep{andrews2023boss} keep the comparison inside the discrete BN scoring family, testing whether stronger permutation search is enough under sparse observation masks (Section~\ref{sec:track1}).
GOLEM~\citep{ng2020golem}, DAGMA~\citep{bello2022dagma}, DiBS~\citep{lorch2021dibs}, and SDCD~\citep{azizi2024sdcd} target continuous structural-equation models, so we treat them as continuous-model references under assumption mismatch.
These comparisons help separate gains from the KG prior from gains due to newer data-only search.

\subsection{Prior Knowledge in Structure Learning}\label{sec:related_prior}

Prior knowledge has been injected into structure learning in several forms.
Bayesian scores have long admitted explicit priors over DAG structures, alongside the Dirichlet parameter prior that defines BDeu~\citep{heckerman1995learning}.
In many practical BN learners, however, the structure prior is uniform or reduced to a simple edge-count penalty~\citep{scutari2014bayesian}.
Other methods let the user fix or forbid certain edges \citep{meek1995causal} or impose ordering constraints \citep{cooper1992bayesian}.
These hard constraints are powerful when knowledge is certain; our setting treats domain knowledge as graded evidence.
Richer domain-specific priors exist; for example, \citet{li2007knowledge} use process knowledge in manufacturing and \citet{wu2011altered} use knowledge about brain regions in Alzheimer's disease research.
Inclusion-driven arc priors also assign finite probabilities to candidate arcs \citep{castelo2003inclusion,scutari2014bayesian}.
The missing piece is not the existence of arc priors; it is how to keep a weighted external graph useful when the data are sparse, discrete, and missing at the family level.
KG-SoftMAP stays in the inclusion-prior lineage but changes the operating regime: weighted KG edges enter as bounded log-odds and are coupled to family-wise available-case BDeu, so the score is evaluated on the same complete cases that define each candidate family.
The contribution is this available-case coupling together with an empirical operating-boundary characterization of where finite-strength priors help, where they fail, and what transfers across search routines (Section~\ref{sec:track1}).
Section~\ref{sec:track1} checks portability through \textsc{bnlearn}'s Castelo--Siebes (\texttt{cs}) arc prior.

The choice of structure prior also interacts with score equivalence: a prior that treats Markov-equivalent DAGs differently can bias learning toward one member of an equivalence class~\citep{scutari2013priorposterior}. Our KG prior deliberately uses directed edge information, so Section~\ref{sec:method} discusses the design implications and an equivalence-aware variant.

A recent line of work injects machine-readable domain knowledge, often elicited from large language models, into the structure search.
ILS-CSL~\citep{ban2023ilscsl} iteratively applies LLM-derived pairwise constraints; \citet{ban2025harmonized} soften these into a harmonized structure prior, and \citet{ban2025harnessing} source the knowledge from KGs.
This is the nearest LLM/prior-guided line to ours, and it differs on two axes: these methods consume binary or near-binary edge judgments under standard complete-data scores, whereas KG-SoftMAP maps continuous KG confidence weights into bounded log-odds and couples them to family-wise available-case BDeu in the extreme-sparse regime.
We therefore include two mechanism checks in the supplementary material: a same-KG near-hard-prior comparison and a sparse soft hill-climbing (softHC) adaptation of ILS-CSL's pairwise-constraint mechanism.

\subsection{LLM-Assisted Causal Discovery}\label{sec:related_llm}

Large language models have been used for causal discovery in three broad roles~\citep{wan2025llmcdsurvey,ma2025llmcisurvey}: generating a graph directly, refining an algorithm's output, and supplying priors to a statistical learner.
\citet{zhang2025promptbn} propose PromptBN for LLM-based BN graph elicitation and ReActBN for data-aware LLM-guided refinement.
Causal Modelling Agents~\citep{abdulaal2024causalagents} couple LLM metadata reasoning with structural causal models, and \citet{darvariu2024llmpriors} use LLM edge judgments as priors for causal graph discovery in a continuous-optimization setting.
We differ from Darvariu et al. by operating on discrete data with family-level missingness and by deriving graded prior strength from KG confidence weights rather than binary LLM votes.
\citet{srivastava2025sciencegrounded} highlight benchmark-recall risks for LLM-only discovery on standard BN repositories and introduce NovelGraphs as a cleaner test bed.
Our setup follows the hybrid direction: LLM extraction is one possible source of a logit prior inside a discrete BDeu-based learner.
The prior keeps KG edges revisable, and the corruption, zero-signal, and NovelGraphs checks quantify how performance changes as prior quality varies.

\subsection{Educational Prediction and Knowledge Tracing}\label{sec:related_kt}

Our real-data evaluation uses educational datasets, where Bayesian networks have long been used to model student knowledge and misconceptions, from Bayesian knowledge tracing~\citep{corbett1995bkt} to misconception diagnosis~\citep{lee2011subtractionbugs}.
Modern knowledge tracing instead predicts a student's next response from a time-ordered interaction sequence, with models such as DKT~\citep{piech2015dkt}, SAKT~\citep{pandey2019sakt}, AKT~\citep{ghosh2020akt}, and simpleKT~\citep{liu2023simplekt}; pyKT~\citep{liu2022pykt} provides standardized benchmarking tools.
Our task is different: we evaluate static posterior prediction for held-out concept states from co-observed concept states in the same record, using SAF~\citep{filighera-etal-2022-answer}, ASSISTments~\citep{feng2009assistments}, and Eedi~\citep{wang2020eedi} to span different KG provenances.
Because the evaluation is a static per-record posterior query rather than next-response prediction, sequence-based KT models solve a different task; we therefore compare against static discriminative predictors in Section~\ref{sec:track2}.
\section{Proposed Approach}
\label{sec:method}

KG-SoftMAP takes two inputs: a sparse $N \times p$ discrete matrix $\mathbf{D}$, with entries $D_{ic}\in\{0,\dots,r_c-1\}$ or unobserved, and a weighted directed graph $\mathbf{K}$ with edges $\{(u,v,w_{uv})\}$, $w_{uv}\in[0,1]$.
Section~\ref{sec:kg} describes an optional LLM-based step that extracts $\mathbf{K}$ when no curated KG exists; Section~\ref{sec:preprocess} describes how raw observations become $\mathbf{D}$; Section~\ref{sec:bn} presents the MAP structure learner, which is agnostic to how either input was produced.

\paragraph{Running example: educational assessment.}
We use the Short Answer Feedback (SAF) dataset~\citep{filighera-etal-2022-answer} to make the inputs concrete: variables are concepts, rows are analysis instances with observed concept evidence, and KG edges are prerequisite-style concept relations.
SAF-specific labels, voting rules, and artifact definitions are deferred to Section~\ref{sec:setup} and the supplementary material.

\subsection{Inputs and Optional KG Construction}
\label{sec:kg}

The KG can come from any source: expert annotation, structured databases, or automated extraction.
When no curated KG is available, we use an LLM-based pipeline that takes domain reference material as input and produces two outputs.
The first output is \emph{variable relations}: latent variables and directed relations among them (e.g., prerequisite dependencies), each with a confidence weight, forming the edges of $\mathbf{K}$ that enter the MAP objective in Section~\ref{sec:bn}.
Edge weights are confidence scores assigned by the LLM conditioned on the reference material; because the model also draws on what it learned during pretraining, Section~\ref{sec:track1} includes provenance checks.
The second output is optional \emph{instance-level deviations}: per-instance, concept-specific evidence with severity weights, used only in dataset-specific preprocessing and never in the structure prior.
The two outputs are independent: a domain with a curated KG and structured observations can skip extraction and supply $\mathbf{K}$ and $\mathbf{D}$ directly.
Extracted candidates pass programmatic validation, including concept-name filtering, confidence thresholding, and an acyclicity check; details are in the supplementary material.
Figure~\ref{fig:saf_pipeline} summarizes how SAF instantiates these generic inputs.

\begin{figure}[t]
\centering
\includegraphics[width=\columnwidth]{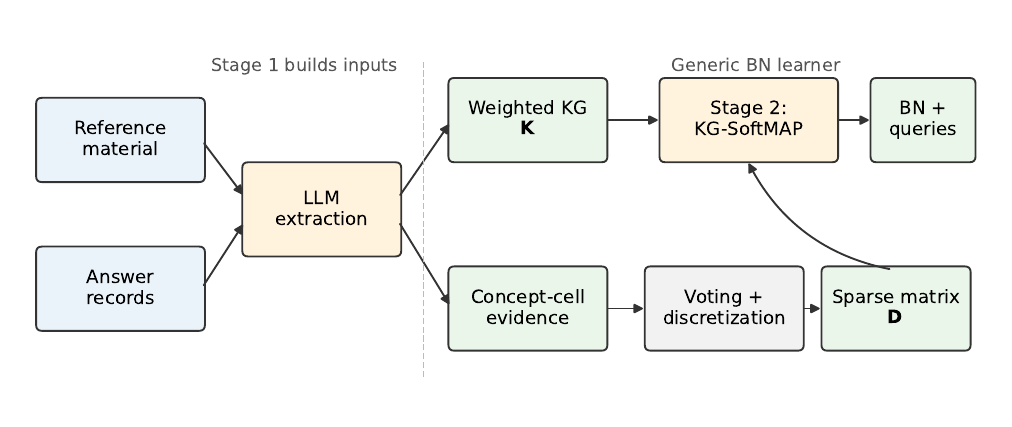}
\caption{Instantiation of the KG-SoftMAP pipeline on the Short Answer Feedback (SAF) dataset. Domain reference material yields the weighted KG $\mathbf{K}$, while answer records yield per-instance concept evidence. Dataset-specific voting and discretization turn that evidence into the sparse matrix $\mathbf{D}$. The structure learner sees only $\mathbf{K}$ and $\mathbf{D}$.}
\label{fig:saf_pipeline}
\end{figure}

\subsection{Sparse Data Representation and Preprocessing}
\label{sec:preprocess}

The matrix $\mathbf{D}$ above is application-specific.
Each entry $D_{ic} \in \{0, \dots, r_c{-}1\}$ is a categorical state of variable $c$ for instance $i$, or unobserved when missing, where $r_c$ is the number of states of variable $c$.
We define the \emph{observation rate} $\rho = |\{(i,c) : D_{ic} \text{ observed}\}| / (N \cdot p)$.
Converting raw data into $\mathbf{D}$ typically involves \emph{discretization}, mapping continuous observations such as scores, sensor readings, or biomarker levels to categorical states, and \emph{aggregation}, combining multiple raw observations for the same cell $(i,c)$ into one label by majority or weighted voting.
When the raw data is already a sparse discrete matrix (e.g., survey responses, diagnostic codes), both steps are trivial and $\mathbf{D}$ can be used directly.
For the synthetic benchmarks, we sample complete data from the true conditional probability distributions (CPDs) and mask entries uniformly at random at rate $1{-}\rho$, so no preprocessing is needed.

\subsection{KG-SoftMAP Structure Learning}
\label{sec:bn}

Given the sparse matrix $\mathbf{D}$ and the KG $\mathbf{K}$, KG-SoftMAP learns a DAG by maximizing a MAP objective that balances data fit against the KG prior.
The learning step below uses only the discrete entries of $\mathbf{D}$ and the edge weights of $\mathbf{K}$.
We present the data score, the KG prior, the search procedure, and then the local score analysis.

\paragraph{MAP Objective.}
We find the DAG $\mathbf{G}^* = (V, E)$ that maximizes the log posterior:
\begin{equation}\label{eq:maps}
\mathbf{G}^* = \arg\max_{\mathbf{G}} \Big[\, \underbrace{\log P(\mathbf{D} \mid \mathbf{G})}_{\text{data fit}} \;+\; \underbrace{\log P(\mathbf{G} \mid \mathbf{K})}_{\text{KG prior}} \,\Big].
\end{equation}
The first term measures how well the graph fits the data. The second pulls the graph toward structures consistent with the KG.

\paragraph{Data Likelihood and Scoring under Sparsity.}
Under the standard BN factorization, the marginal likelihood decomposes by node:
\begin{equation}\label{eq:bdeu}
\log P(\mathbf{D} \mid \mathbf{G}) = \sum_{v=1}^{p} S_{\text{BDeu}}\!\big(v,\, \mathrm{Pa}(v)\big),
\end{equation}
where $\mathrm{Pa}(v)$ is the parent set of node $v$ in $\mathbf{G}$, and $S_{\text{BDeu}}\!\big(v,\mathrm{Pa}(v)\big)$ is the local node score for child $v$ and its parent set. This score integrates out the local conditional probability parameters under a Dirichlet prior $\mathrm{Dir}(\alpha_{1j},\ldots,\alpha_{r_v j})$ for each parent configuration $j$:
\begin{align}\label{eq:bdeu_local}
&S_{\text{BDeu}}\!\big(v,\, \mathrm{Pa}(v)\big) \notag \\
&= \sum_{j} \bigg[\, \log \Gamma(\alpha_j) - \log \Gamma(\alpha_j \!+\! N_j) \notag \\
&\quad\; + \sum_{i} \Big(\log \Gamma(\alpha_{ij} \!+\! N_{ij}) - \log \Gamma(\alpha_{ij})\Big) \bigg],
\end{align}
where $j$ indexes parent configurations, $i$ indexes node states, $N_{ij}$ is the count of cases where node $v$ is in state $i$ with parents in configuration $j$, and $N_j=\sum_i N_{ij}$. The hyperparameters are $\alpha_j = \alpha / q$ and $\alpha_{ij} = \alpha / (q r_v)$, where $q = \prod_{z \in \mathrm{Pa}(v)} r_z$ is the total number of parent configurations and $r_v$ is the number of states of node $v$. We use the standard BDeu equivalent sample size $\alpha{=}1$~\citep{heckerman1995learning} unless otherwise stated; the supplementary material reports a sensitivity sweep.
Adding a parent increases $q$, which spreads the pseudocounts thinner, acting as an automatic complexity penalty.

With missing data, we compute each local score on rows where node $v$ and all its current parents are jointly observed. This family-wise available-case BDeu surrogate is the default for planned or Missing Completely At Random (MCAR) sparse observations; its effective sample size is the number of rows that observe the relevant child--parent family, and it avoids imputation.
We also apply a scoring gate: a candidate family with parent set $S$ is scored only when its complete-case count $m(v,S)=\sum_j N_j$ satisfies $m(v,S)\ge\max(5,q)$.
This motivates using BDeu rather than relying on a global-$N$ BIC penalty: \citet{balov2013consistent} showed that BIC penalties based on a single sample size can be inconsistent under MCAR when local families have different observation rates, while our score is computed on the family-specific complete-case counts that enter each local likelihood.

\emph{Robustness variants.}\quad Optional extensions for MAR settings and adaptive-ESS smoothing are described in the supplementary material. The main learner uses complete-case scoring and standard BDeu; under the default scoring gate, the adaptive-ESS guard is dormant in every reported experiment.

\paragraph{KG Structure Prior.}
The second term in Eq.~\eqref{eq:maps} encodes how well the graph matches the KG.
We use an edge-factored structure-prior score over acyclic DAGs.
Before imposing the acyclicity restriction, each candidate directed edge receives an independent Bernoulli inclusion probability $\theta_{uv}$.
Restricted to valid DAGs, the resulting log-prior score decomposes, up to an additive constant including the DAG-space normalizer, as a sum of $\mathrm{logit}(\theta_{uv})$ over the edges present in $\mathbf{G}$:
\begin{equation}\label{eq:prior}
\log P(\mathbf{G} \mid \mathbf{K}) = \!\!\sum_{(u,v) \in E}\!\! \mathrm{logit}(\theta_{uv}) + C,
\end{equation}
where $C$ is a constant, $\theta_{uv}$ is the prior inclusion probability of edge $(u \!\to\! v)$, and $\mathrm{logit}(\theta) = \log \frac{\theta}{1-\theta}$.
The logit is a linear function of the KG confidence:
\begin{equation}\label{eq:logit}
\mathrm{logit}(\theta_{uv}) = \beta_0 + \beta_1 \cdot w_{uv},
\end{equation}
with $\beta_0 = \mathrm{logit}(\theta_0)$ and $\beta_1 = \mathrm{logit}(\theta_{\max}) - \beta_0$.
$\theta_0$ is the base inclusion probability for edges absent from the KG ($w_{uv}{=}0$); when $\theta_0 \ll 0.5$, the logit is negative, acting as a sparsity penalty.
$\theta_{\max}$ is the inclusion probability for full-confidence KG edges ($w_{uv}{=}1$).
For example, $\theta_0 = 0.01$ and $\theta_{\max} = 0.8$ give logits of about $-4.6$ and $+1.4$, respectively.
Thus the KG contributes a finite log-odds bonus or penalty while keeping all edge decisions revisable.

\emph{Scope.}\quad The prior consumes \emph{weighted directed candidate edges} $\{(u,v,w_{uv})\}$ and leaves richer KG semantics such as relation types, edge labels, attributes, and multi-hop reasoning for future extensions. This keeps the prior \emph{source-agnostic}: a curated expert ontology, a structured database, or an LLM-extracted graph can all instantiate $\mathbf{K}$ identically, and the learner is unchanged by how the edges were obtained. The KG affects edge inclusion log-odds; the data likelihood remains the BDeu score on the sparse matrix.

\emph{Design note: Markov equivalence.}\quad The edge-factored prior in Eq.~\eqref{eq:prior} can assign different mass to Markov-equivalent DAGs~\citep{scutari2013priorposterior}. The supplementary material reports an equivalence-aware variant that scores the prior symmetrically over the skeleton and only orients compelled edges; recovered graphs remain nearly unchanged.

\paragraph{Greedy Search.}
We maximize Eq.~\eqref{eq:maps} by greedy local search (hill climbing) over edge operations, applying the legal operation with the largest positive gain $\Delta$ at each step.

\textit{Add} edge $(u \!\to\! v)$:
\begin{align}\label{eq:add}
\Delta_{\text{add}} &= S_{\text{BDeu}}\!\big(v,\, \mathrm{Pa}(v) \!\cup\! \{u\}\big) \notag \\
&\quad - S_{\text{BDeu}}\!\big(v,\, \mathrm{Pa}(v)\big) + \mathrm{logit}(\theta_{uv}).
\end{align}

Delete is the reverse move, with the prior logit entering with the opposite sign: removing an edge forfeits its prior bonus, or relieves the penalty for non-KG edges.
Swap replaces parent $u_{\text{old}}$ with $u_{\text{new}}$ at node $v$ in one step, with gain equal to the BDeu change plus $\mathrm{logit}(\theta_{u_{\text{new}},v})-\mathrm{logit}(\theta_{u_{\text{old}},v})$; this compound move helps avoid local optima caused by parent-set limits.
The search has two phases: the \emph{forward} phase repeatedly applies the best add or swap; the \emph{backward} phase removes edges that hurt the objective.
Acyclicity is enforced at every step, a score cache avoids redundant BDeu computations, and finite monotone improvement is guaranteed because each accepted move strictly increases Eq.~\eqref{eq:maps} over a finite DAG space.

\paragraph{Local Score Analysis.}
The add gain in Eq.~\eqref{eq:add} decomposes into a BDeu gain and a finite logit-prior term.
For a candidate add move $e=(u\!\to\!v)$ with current parent set $S$, let
$m_e$ be the complete-case count for the after-add family $(v,S\cup\{u\})$,
$\omega_e=\mathrm{logit}(\theta_{uv})$,
$\lambda_0=-\mathrm{logit}(\theta_0)>0$ be the absent-edge penalty, and
$q_S=\prod_{z\in S}r_z$ be the number of current parent configurations.
Then $\Delta\nu_e=q_S(r_u-1)(r_v-1)$ is the additional number of local free parameters.
Let $\gamma_e=I(X_v;X_u\mid X_S)$ denote the conditional dependence strength under the distribution of rows that jointly observe the after-add family; under MCAR this coincides with the full-population conditional mutual information.
Let $\mathcal E_e(R_e)$ denote the event that the after-add family has $m_e$ complete cases and the BDeu part of Eq.~\eqref{eq:add} admits the expansion
\[
\Delta_{\mathrm{BDeu}}(e;S)
=m_e\gamma_e-\frac{\Delta\nu_e}{2}\log m_e+\varepsilon_e,
\qquad |\varepsilon_e|\le R_e .
\]
Here $R_e$ collects the Dirichlet finite-sample remainder, empirical fluctuation, and any support-change term induced by family-wise available-case scoring.

\begin{proposition}[Local edge-wise discrimination]\label{prop:edge_discrimination}
Consider an eligible acyclic add move $e=(u\!\to\!v)$ whose after-add family satisfies the scoring gate.
On $\mathcal E_e(R_e)$, the MAP add gain in Eq.~\eqref{eq:add} is positive whenever
\[
m_e\gamma_e+\omega_e>\frac{\Delta\nu_e}{2}\log m_e+R_e,
\]
and negative whenever
\[
m_e\gamma_e+\omega_e<\frac{\Delta\nu_e}{2}\log m_e-R_e.
\]
In particular, a false KG edge with $\gamma_e=0$ is rejected once
\[
\frac{\Delta\nu_e}{2}\log m_e>\omega_e+R_e .
\]
A false non-KG edge with $\omega_e\le -\lambda_0$ is rejected once
\[
\lambda_0+\frac{\Delta\nu_e}{2}\log m_e>R_e .
\]
\end{proposition}

This is a local score-mechanism result: it characterizes one eligible edge move before acyclicity conflicts and search-order effects enter.
KG weights shift the edge log-odds upward, while unsupported edges pay a finite penalty.
The supplementary material derives the common-support expansion and states how available-case support changes enter $R_e$.

\paragraph{Operating Regime: Complete-Case Scarcity.}
Complete-case counts determine how much evidence can enter Proposition~\ref{prop:edge_discrimination}. For a family $(v,\mathrm{Pa}(v))$ with $k$ parents under MCAR,
\begin{equation}\label{eq:neff}
\mathbb{E}[N_{\mathrm{complete}}] \;=\; N \cdot \rho^{\,k+1},
\end{equation}
because a row is complete only when the child and all $k$ parents are simultaneously observed.
For a candidate add move at current parent depth $k$, the after-add family contains $k{+}1$ parents, so the expected scored count is $N\rho^{k+2}$.

\begin{proposition}[Complete-case availability threshold]\label{prop:cc_threshold}
Let $m_{\min}$ denote the complete-case threshold imposed by the scoring gate for the after-add family (default: $\max(5,q)$ for that family), and consider adding an edge to a node whose current parent set has size $k$.
Under independent MCAR observation with rate $\rho$, the after-add complete-case count is
$M_k \sim \mathrm{Binomial}(N,\rho^{k+2})$.
The nominal observation-rate threshold for the expected scored count to reach the gate is
\[
\rho_{\min}(k) = \left(\frac{m_{\min}}{N}\right)^{1/(k+2)}.
\]
If $\rho < \rho_{\min}(k)$, then $\mathbb{E}[M_k] < m_{\min}$ and
\[
\Pr(M_k \ge m_{\min})
\le
\exp\{-N\rho^{k+2} h(m_{\min}/(N\rho^{k+2}))\},
\]
where $h(x)=x\log x-x+1$.
\end{proposition}

Thus, at $N{=}1{,}000$, $\rho{=}0.05$, and the minimum component of the default gate $m_{\min}{=}5$, even adding the first parent has expected count $2.5$ and lies below $\rho_{\min}(0){\approx}0.071$.
When a candidate family falls below the gate, the move is unavailable and no edge is added regardless of the prior.
For families just above the gate, $m_e\gamma_e$ is small relative to $|\omega_e|$ in Proposition~\ref{prop:edge_discrimination}, so the KG logit can dominate the local decision.
As $\rho$ rises above the relevant $\rho_{\min}(k)$ values, more candidate families pass the gate and BDeu evidence can revise both KG-supported and unsupported edge decisions.
Whole-graph recovery then depends on many such local decisions, acyclicity, search order, and KG quality; Section~\ref{sec:track1} tests that graph-level behavior directly.
The supplementary material gives the proof of the threshold bound.

\emph{Prior calibration.}\quad All reported results use the fixed defaults $\theta_0{=}0.01$, $\theta_{\max}{=}0.8$, which sit on a performance plateau (Section~\ref{sec:track1}). When labeled validation data is available, these endpoints may be tuned by instance-level cross-validation with an edge-count penalty; the grid and formula are reported in the supplementary material.

\paragraph{Parameter Estimation.}
After the structure is fixed, we estimate conditional probability tables (CPTs) via the standard Dirichlet posterior~\citep{heckerman1995learning,cooper1992bayesian}:
\begin{equation}\label{eq:cpd}
\hat{P}(X_v \!=\! i \mid \mathrm{Pa}(v) \!=\! j) = \frac{N_{ij} + \alpha_{ij}}{N_j + \alpha_j},
\end{equation}
where $N_{ij}$ and $N_j$ are the family-wise complete-case counts used in the scoring step, and $\alpha$ is the standard (non-adaptive) equivalent sample size.
When $N_j = 0$, the estimate reduces to the uniform prior $1/r_v$.

\section{Experiments}\label{sec:experiments}

We use known-DAG synthetic benchmarks for directed structure recovery, then use real educational datasets for prediction, calibration, and KG-consistency, where no ground-truth DAG is available.
\textbf{Track~1} compares KG-SoftMAP with classical, modern data-only, continuous-DAG, and prior-guided baselines, and adds robustness and ablation analyses.
\textbf{Track~2} examines how the learned probabilistic concept model behaves on real sparse educational data.

\subsection{Setup}\label{sec:setup}

\paragraph{Synthetic benchmarks.}
We use six discrete BN benchmarks from the bnlearn repository:
\texttt{cancer}~(5/4 nodes/edges), \texttt{asia}~(8/8), \texttt{sachs}~(11/17),
\texttt{child}~(20/25), \texttt{insurance}~(27/52), and \texttt{alarm}~(37/46).
For each network, we sample $n = 1{,}000$ instances from the true CPDs and mask entries at random.
The main sparse-recovery table reports $\rho \in \{0.05, 0.20, 0.40\}$; a separate diagnostic grid in the supplementary material extends coverage to $\rho\in\{0.60,0.80,1.00\}$.
The synthetic KG is a high-recall, low-confidence-noise prior derived from the true DAG: true edges receive $w_{uv} \sim \mathrm{Uniform}(0.5, 1.0)$, plus $10\%$ false directed edges with $w_{uv} \sim \mathrm{Uniform}(0.0, 0.3)$.
This is a cleanly separable informative prior; corruption, overlapping-weight, and truth-independent KG controls in Section~\ref{sec:track1} vary its quality.

\paragraph{NovelGraphs LLM-provenance check.}
Standard BN benchmarks are useful for controlled recovery but are widely circulated.
Following the benchmark-recall concern studied by \citet{srivastava2025sciencegrounded}, we add two NovelGraphs DAGs, \texttt{alzheimers} and \texttt{covid19-small}.
Their ground-truth DAGs permit Directed-F1 evaluation, while the KG is elicited from variable descriptions rather than derived from the true graph.
Masked-data learners use $\rho\in\{0.05,0.20,0.40\}$ over five random seeds; Table~\ref{tab:novographs} reports the LLM-only, LLM-refined, prior-guided, and data-only discrete results, with continuous-DAG references in the supplementary material.

\paragraph{Real data and SAF case-study role.}
SAF is the main real-data case study: it supplies a sparse concept-state matrix and a fixed LLM-extracted concept KG built from instructor-provided reference answers.
Because SAF has no ground-truth DAG, we use it for prediction, calibration, and KG-consistency diagnostics rather than directed-structure recovery.
Observed SAF concept cells take three states, \textsc{Master}~(0), \textsc{Unsure}~(1), or \textsc{Fail}~(2), obtained from fixed score thresholds and concept-level mistake evidence: extracted per-response indications that a concept was implicated in the response-level error.
Full voting details are in the supplementary material.
We label two SAF artifacts throughout: \textbf{SAF-full} is the original 238-concept extraction used only for descriptive KG-consistency diagnostics, and \textbf{SAF-eval} is a separately rebuilt 191-concept artifact from the same SAF data and pipeline, pruned to the fixed prediction vocabulary used in Table~\ref{tab:saf_pred}.
The synthetic experiments test how effectively the method uses a given prior under known structure; SAF tests the full pipeline including KG construction.
ASSISTments~\citep{feng2009assistments} adds a weak-KG contrast based on heuristic priors, and Eedi~\citep{wang2020eedi} adds an independent expert-ontology prior over leaf-level knowledge components.

\paragraph{Compared methods and protocol.}
We organize comparisons by the question each method group answers.
The main sparse-recovery table compares KG-guided learning against representative data-only discrete BN learners, missing-data BN learners, modern permutation search, and continuous-DAG references under the same sparse masks.
The main-table data-only group includes Ours-NoPrior, Sparse-Cand~\citep{friedman1999learning}, Structural EM~\citep{friedman1998bayesian}, the Adel--de Campos missing-data BN score~\citep{adel2017learning} (Adel-dC in tables), GRaSP-BDeu~\citep{lam2022grasp}, and BOSS~\citep{andrews2023boss}.
The full supplementary grids for Table~\ref{tab:recovery} add standard GES~\citep{chickering2002optimal}, MMHC~\citep{tsamardinos2006max}, PC, and PC-Stable~\citep{spirtes2000causation} for continuity with classical BN baselines; GES and PC also appear as data-only references in the NovelGraphs check (Table~\ref{tab:novographs}).
The continuous-DAG reference group in Table~\ref{tab:recovery} includes GOLEM~\citep{ng2020golem}, DAGMA~\citep{bello2022dagma}, and SDCD~\citep{azizi2024sdcd}. These methods learn DAGs under continuous structural-equation assumptions and require complete numeric inputs, so we run them on imputed, standardized matrices and interpret their rows under model-assumption mismatch.
Matched-KG checks use KG-SoftMAP, MMHC-cand+Prior (internal search-ablation variant), \textsc{bnlearn}+\texttt{cs}, and ILS-CSL-softHC under shared KG inputs (Table~\ref{tab:mechanism}); GES+Prior and additional internal variants are reported in the supplementary material.
The NovelGraphs check adds LLM-pairwise, LLM-BFS, PromptBN, ReActBN~\citep{zhang2025promptbn,srivastava2025sciencegrounded}, and its own data-only discrete references (GES, PC); continuous-DAG references (DAGMA and NOTEARS~\citep{zheng2018dags}) are evaluated in the supplementary material.
Track~2 uses LR, MLP, and XGBoost on the same held-out concept cells as BN inference.
On SAF-full, complete-data baselines use the Dense-50 subset (the 50 most densely observed concepts), while Structural EM, Adel-dC, and Sparse-Cand run on all 238 variables; implementation sources, seeds, and runtime scope are summarized in the supplementary material.

\subsection{Evaluation Metrics}\label{sec:metrics}

\paragraph{Metrics.}
\emph{Structural.}\quad \textbf{SHD} ($\downarrow$): edge additions, deletions, and reversals needed to match the true DAG.
\textbf{Skeleton F1} ($\uparrow$): F1 on the undirected skeleton, fairer for CPDAG methods that leave some edges undirected; full skeleton-F1 tables are in the supplementary material.
\textbf{Directed-F1} ($\uparrow$): F1 on directed edges; for CPDAGs, undirected edges are excluded from both precision and recall.
\textbf{Log\,P(D$\mid$G,$\theta$)} ($\uparrow$): total log-likelihood of observed data under the learned DAG and CPDs. We report it as a \emph{data-fit diagnostic}: denser graphs can attain higher likelihood, so the metric is interpreted alongside KG-consistency. On SAF-full, it provides context for the descriptive diagnostics reported in the supplementary material.
\textbf{KG retention} ($\uparrow$): share of KG edges retained in the learned graph.
\textbf{KG conflicts} ($\downarrow$): edges whose direction contradicts the KG.
\textbf{Edge density} ($\downarrow$): learned directed edges divided by all possible ordered pairs, $|E|/[p(p-1)]$.
\emph{Predictive.}\quad These evaluate the BN's ability to infer held-out cell values: for each observed entry in $\mathbf{D}$, we hide its value, condition the BN on all other observed variables for that instance, and compare the predicted label against the held-out ground truth (details in Section~\ref{sec:track2}).
\textbf{Accuracy} ($\uparrow$): overall correct rate.
\textbf{F1\textsubscript{FAIL}} ($\uparrow$): F1 for \textsc{Fail} class in one-vs-rest.
\textbf{PR-AUC\textsubscript{FAIL}} ($\uparrow$): precision--recall AUC for \textsc{Fail}, computed from $P(\textsc{Fail})$; threshold-free complement to F1\textsubscript{FAIL}.
\textbf{Macro-F1} ($\uparrow$): class-averaged F1.
\textbf{ECE\textsubscript{FAIL}} ($\downarrow$): expected calibration error for $P(\textsc{Fail})$, 10 equal-width bins.
\emph{Feasibility constraints.}\quad The sparse score-search BN learners (KG-SoftMAP, Ours-NoPrior, MMHC-cand+Prior, Sparse-Cand, Structural EM, and Adel-dC) use acyclicity and a max-in-degree cap of~4.
Constraint-based and hybrid baselines (PC, PC-Stable, MMHC, \textsc{bnlearn}+\texttt{cs}), permutation-search baselines (GRaSP, BOSS), and continuous-DAG baselines (GOLEM, DAGMA, SDCD, NOTEARS) are run in their native form and evaluated with the same structural metrics.
We additionally report three \emph{soft} criteria as diagnostics: KG retention $\geq 0.10$, edge density $\leq 0.20$, and KG conflicts $= 0$. These diagnostics are post hoc, and baselines may violate them.
\subsection{Track~1: Controlled Structural Recovery}\label{sec:track1}

\subsubsection{Main Sparse-Recovery Result}

We first evaluate directed structure recovery in the setting where the true DAG is known.
Table~\ref{tab:recovery} reports the main sparse-recovery grid: all six bnlearn benchmarks, three levels of sparsity expressed as observation rates ($\rho\in\{0.05,0.20,0.40\}$), and the main method families under the same sparse masks.
KG-guided rows receive the same high-recall noisy KG; no-KG rows use only the masked discrete matrix.
The continuous-DAG rows assume complete continuous inputs, so we run GOLEM, DAGMA, and SDCD on imputed and standardized numeric matrices.

\begin{table*}[!t]
\centering
\footnotesize
\caption{Directed structure recovery on six synthetic benchmarks in the sparse regime (Directed-F1 means over 10 seeds). Full mean$\pm$SD and additional classical-baseline grids are in the supplement.}
\label{tab:recovery}
\setlength{\tabcolsep}{2.1pt}
\begin{tabular}{llccccccccccc}
\toprule
Net & $\rho$ & \multicolumn{2}{c}{KG-guided} & \multicolumn{6}{c}{No-KG / data-side} & \multicolumn{3}{c}{Cont. stress} \\
\cmidrule(lr){3-4}\cmidrule(lr){5-10}\cmidrule(lr){11-13}
 & & KG-SoftMAP & ILS-CSL & Ours-NoPrior & SpCand & StructEM & Adel-dC & GRaSP & BOSS & GOLEM & DAGMA & SDCD \\
\midrule
\texttt{cancer} & 0.05 & 0.19 & 0.11 & 0.00 & 0.00 & 0.36 & 0.36 & 0.04 & 0.04 & 0.00 & 0.00 & 0.00 \\
\texttt{cancer} & 0.20 & 0.62 & 0.33 & 0.22 & 0.22 & 0.12 & 0.12 & 0.04 & 0.04 & 0.00 & 0.00 & 0.00 \\
\texttt{cancer} & 0.40 & 0.54 & 0.31 & 0.26 & 0.26 & 0.11 & 0.11 & 0.12 & 0.12 & 0.00 & 0.00 & 0.15 \\
\addlinespace[1pt]
\texttt{asia} & 0.05 & 0.32 & 0.24 & 0.00 & 0.00 & 0.15 & 0.15 & 0.02 & 0.02 & 0.00 & 0.00 & 0.00 \\
\texttt{asia} & 0.20 & 0.66 & 0.60 & 0.26 & 0.26 & 0.22 & 0.19 & 0.32 & 0.32 & 0.00 & 0.00 & 0.09 \\
\texttt{asia} & 0.40 & 0.64 & 0.44 & 0.30 & 0.30 & 0.23 & 0.25 & 0.48 & 0.46 & 0.04 & 0.02 & 0.14 \\
\addlinespace[1pt]
\texttt{sachs} & 0.05 & 0.20 & 0.14 & 0.00 & 0.00 & 0.04 & 0.04 & 0.01 & 0.01 & 0.00 & 0.00 & 0.06 \\
\texttt{sachs} & 0.20 & 0.48 & 0.38 & 0.23 & 0.24 & 0.16 & 0.15 & 0.02 & 0.02 & 0.00 & 0.00 & 0.13 \\
\texttt{sachs} & 0.40 & 0.61 & 0.38 & 0.25 & 0.24 & 0.30 & 0.29 & 0.06 & 0.07 & 0.00 & 0.00 & 0.30 \\
\addlinespace[1pt]
\texttt{child} & 0.05 & 0.19 & 0.12 & 0.00 & 0.00 & 0.03 & 0.03 & 0.00 & 0.00 & 0.00 & 0.00 & 0.00 \\
\texttt{child} & 0.20 & 0.48 & 0.33 & 0.16 & 0.20 & 0.14 & 0.14 & 0.08 & 0.08 & 0.00 & 0.00 & 0.12 \\
\texttt{child} & 0.40 & 0.46 & 0.29 & 0.19 & 0.25 & 0.38 & 0.38 & 0.33 & 0.30 & 0.02 & 0.02 & 0.25 \\
\addlinespace[1pt]
\texttt{insurance} & 0.05 & 0.21 & 0.15 & 0.00 & 0.00 & 0.03 & 0.03 & 0.01 & 0.01 & 0.00 & 0.00 & 0.06 \\
\texttt{insurance} & 0.20 & 0.44 & 0.28 & 0.14 & 0.17 & 0.13 & 0.15 & 0.08 & 0.08 & 0.00 & 0.00 & 0.09 \\
\texttt{insurance} & 0.40 & 0.47 & 0.23 & 0.16 & 0.19 & 0.23 & 0.22 & 0.27 & 0.30 & 0.02 & 0.03 & 0.17 \\
\addlinespace[1pt]
\texttt{alarm} & 0.05 & 0.20 & 0.12 & 0.00 & 0.00 & 0.04 & 0.04 & 0.01 & 0.01 & 0.00 & 0.00 & 0.00 \\
\texttt{alarm} & 0.20 & 0.49 & 0.37 & 0.10 & 0.19 & 0.13 & 0.16 & 0.12 & 0.12 & 0.01 & 0.01 & 0.06 \\
\texttt{alarm} & 0.40 & 0.50 & 0.32 & 0.13 & 0.26 & 0.23 & 0.22 & 0.25 & 0.24 & 0.05 & 0.06 & 0.13 \\
\bottomrule
\end{tabular}
\end{table*}

The same broad pattern appears from \texttt{cancer} (5 nodes) through \texttt{alarm} (37 nodes).
At $\rho{=}0.05$, KG-SoftMAP recovers partial directed structure while most no-KG methods remain near zero.
At $\rho{=}0.20$ and $\rho{=}0.40$, the informative KG usually moves recovery above the data-only and continuous-model references, although strong data-only search can become competitive on some networks.
The two denser rates are close rather than strictly monotone: the top cell is $0.66$ at $\rho{=}0.20$ and $0.64$ at $\rho{=}0.40$.
We therefore report both rates instead of compressing them into a single density trend.
Ours-NoPrior does not collapse after the corrected co-observation-count rerun, but it remains well below KG-SoftMAP in cross-network means ($0.00/0.19/0.21$ vs.\ $0.22/0.53/0.54$ for $\rho{=}0.05/0.20/0.40$).
This makes the broader data-only group important for context and motivates the matched-KG mechanism check below.
ILS-CSL-softHC, the other KG-guided row, is useful but less stable across observation rates; it declines from $\rho{=}0.20$ to $\rho{=}0.40$ on five of six networks.
The supplementary ILS-CSL grid shows the same aggregate pattern and suggests that the sparse KG-MAP scorer controls unsupported edge growth more strongly at $\rho{=}0.40$.
The supplementary grids add the standard GES/MMHC/PC/PC-Stable baselines and extend the same sparse-recovery protocol to \texttt{barley}, \texttt{mildew}, and \texttt{water}, including sparse-adapted ILS-CSL variants.

We also remove the separated confidence-band cue while keeping the same high-recall KG edge set: true and false KG edges both receive weights from the same range.
In this confidence-band-removed variant, KG-SoftMAP reaches mean Directed-F1 $0.23/0.52/0.51$ at $\rho{=}0.05/0.20/0.40$, compared with Ours-NoPrior $0.00/0.18/0.21$ and GRaSP/BOSS around $0.03/0.13/0.25$.
This check shows that recovery in Table~\ref{tab:recovery} is not driven by calibrated confidence bands alone; edge-set quality is tested separately by the corruption and density-mismatch sweeps below.
Full per-network mean$\pm$SD results are in the supplementary material.

The remaining checks answer three mechanism questions.
Table~\ref{tab:mechanism} asks whether the sparse KG-MAP scorer transfers across search routines and prior interfaces; Table~\ref{tab:novographs} tests LLM-derived graph signal outside canonical BN repositories; Figure~\ref{fig:robust} varies KG quality directly.
A zero-signal \texttt{random\_disjoint} KG defines the floor of this stress test.

\subsubsection{Mechanism and KG-Quality Checks}
\begin{figure}[t]
\centering
\includegraphics[width=0.76\columnwidth]{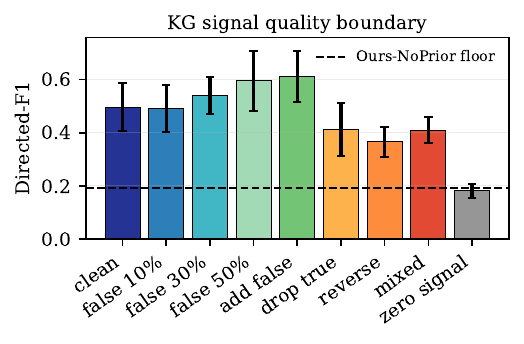}
\caption{Recovery tracks KG signal quality (mean Directed-F1 over \texttt{child}/\texttt{insurance}/\texttt{sachs}, 5 seeds, $\rho{=}0.4$). Starting from the informative synthetic KG, we vary false-edge rates and apply targeted $30\%$ drop, reverse, and mixed corruptions. False additions are milder because true-edge recall remains; drop/reverse/mixed errors remove or misorient KG signal. The zero-signal \texttt{random\_disjoint} KG returns to the Ours-NoPrior floor. Horizontal references mark clean KG, flat prior, and zero-signal KG.}
\label{fig:robust}
\end{figure}

\paragraph{Same-KG mechanism.}
Table~\ref{tab:mechanism} gives the matched-KG mechanism check behind the main recovery table.
It compares four prior-guided learners on the same six networks at $\rho{=}0.4$ under two synthetic KG false-edge rates ($10\%$ and $30\%$), distinct from the structured corruption protocols in Figure~\ref{fig:robust}.
The rows play different roles.
MMHC-cand+Prior is an \emph{internal} search-ablation variant: it keeps our sparse BDeu+logit scorer, KG prior, CPD estimation, acyclicity constraint, and max-in-degree cap, but uses MMHC-style candidate restriction before the local search.
We include it to ask whether the gain depends on KG-SoftMAP's particular search routine.
\textsc{bnlearn}+\texttt{cs} and ILS-CSL-softHC are \emph{external} prior-guided checks.
They receive the same KG signal in their native forms: Castelo--Siebes endpoint probabilities for \textsc{bnlearn}+\texttt{cs}, and pairwise soft constraints for ILS-CSL-softHC.

The close match between KG-SoftMAP and MMHC-cand+Prior shows that the sparse BDeu+logit-prior scorer, rather than one search routine, is the main portable component.
We therefore read KG-SoftMAP and MMHC-cand+Prior as two instantiations of the same sparse KG-MAP scorer family.
The same table clarifies what transfers.
\textsc{bnlearn}+\texttt{cs}, the closest existing soft-arc-prior baseline, matches the aggregate recovery of KG-SoftMAP when given the same high-recall synthetic KG.
This tie is the point of the mechanism check: finite-strength KG evidence is portable.
KG-SoftMAP contributes the family-sparse available-case coupling and the operating-boundary evaluation, rather than claiming a proprietary search advantage.
The internal rows isolate the scorer from the search restriction, while \textsc{bnlearn}+\texttt{cs} serves as a third-party soft-prior portability check rather than a direct scorer ablation because it changes both missing-data handling and prior interface.
ILS-CSL-softHC remains above the no-KG floor from Table~\ref{tab:recovery}, but is lower in this aggregate.
After aggregation over networks, the external rows change little across the two false-edge rates; both their means and SDs round to the same values.
Full per-network grids and seed-level details are in the supplementary material.

\begin{table}[t]
\centering
\caption{Same-KG mechanism check (Directed-F1, mean$\pm$SD across six network-level means; $\rho{=}0.4$). The two columns set the false-edge rate in the synthetic KG; per-network seed counts and grids are in the supplementary material.}
\label{tab:mechanism}
\small
\begin{tabular}{lcc}
\toprule
Method & 10\% false edges & 30\% false edges \\
\midrule
KG-SoftMAP (Ours)  & 0.54$\pm$0.07 & 0.58$\pm$0.05 \\
MMHC-cand+Prior    & 0.55$\pm$0.06 & 0.58$\pm$0.06 \\
\textsc{bnlearn}+\texttt{cs} & 0.58$\pm$0.15 & 0.58$\pm$0.15 \\
ILS-CSL-softHC      & 0.33$\pm$0.07 & 0.33$\pm$0.07 \\
\bottomrule
\end{tabular}
\end{table}

\paragraph{LLM-extracted KG provenance.}
On standard BN benchmarks, blind elicitation produces useful but uneven KGs: \texttt{asia} is recovered exactly (precision/recall $1.00/1.00$), while \texttt{sachs} and \texttt{child} have imperfect precision/recall ($0.62/0.76$ and $0.54/0.80$).
Feeding these KGs into KG-SoftMAP still improves over the data-only floor in the supplementary standard-benchmark check, especially once $\rho{\ge}0.2$.

NovelGraphs is the harder provenance test: the KG is elicited from variable descriptions rather than derived from the true DAG.
Table~\ref{tab:novographs} compares four families on \texttt{alzheimers} and \texttt{covid19-small}: data-free LLM graph elicitation, including PromptBN in Panel A; sparse prior-guided learners; ReActBN as a data-conditioned LLM-refined learner; and data-only discrete learners~\citep{zhang2025promptbn}.
The LLM-only rows show that variable descriptions contain usable graph signal.
KG-SoftMAP+LLM-KG remains competitive in this harder setting, nearly matching the best learner on \texttt{covid19-small} at $\rho{=}0.20$ ($0.45$ vs.\ $0.46$) and improving over Ours-NoPrior at every reported rate.
The broader pattern is heterogeneous rather than one-sided: prior-guided and LLM-refined learners lead at low observation rates, while data-only search catches up by $\rho{=}0.40$.
The LLM-only rows are not sparse-data learners: they assert a graph from descriptions alone and do not fit CPDs or a joint distribution.
The sparse learner turns graph elicitation into a fitted probabilistic model with CPDs and posterior queries; structural F1 is only one part of that comparison.
The margin is narrower than in Table~\ref{tab:recovery}, as expected: the KG here is elicited from descriptions rather than generated from the true DAG, so the prior carries less signal. This pattern matches Figure~\ref{fig:robust}: external graph signal helps sparse recovery, and the KG-MAP score converts imperfect LLM graph signal into recoverable structure.

\begin{table}[t]
\centering
\caption{NovelGraphs LLM-provenance check (Directed-F1). Panel A reports data-free graph quality for one elicited graph per method/dataset; Panel B reports sparse-learner means over 5 seeds. Each Panel B entry is $\rho{=}0.05/0.20/0.40$. Bold marks the best Panel B value within each $\rho$ column; full mean$\pm$SD and continuous-model grids are in the supplementary material.}
\label{tab:novographs}
\small
\setlength{\tabcolsep}{2.4pt}
\begin{tabular}{lcc}
\toprule
\multicolumn{3}{l}{\textbf{A. LLM-only graph elicitation}} \\
Method & \texttt{alzheimers} & \texttt{covid19-small} \\
\midrule
LLM-pairwise & 0.70 & 0.62 \\
LLM-BFS & 0.60 & 0.00 \\
PromptBN & 0.50 & 0.67 \\
\midrule
\multicolumn{3}{l}{\textbf{B. Sparse learners using observational masks}} \\
Method & \texttt{alzheimers} & \texttt{covid19-small} \\
\midrule
\multicolumn{3}{l}{\emph{Prior-guided sparse learners}} \\
KG-SoftMAP+LLM-KG & 0.13 / 0.25 / 0.23 & 0.22 / 0.45 / 0.49 \\
MMHC-cand+Prior & 0.13 / 0.25 / 0.25 & 0.22 / 0.44 / 0.54 \\
ILS-CSL-softHC & 0.14 / 0.20 / 0.15 & 0.17 / 0.31 / 0.34 \\
\addlinespace[1pt]
\multicolumn{3}{l}{\emph{LLM-refined learner}} \\
ReActBN & \textbf{0.29} / 0.25 / 0.22 & \textbf{0.36} / \textbf{0.46} / 0.54 \\
\addlinespace[1pt]
\multicolumn{3}{l}{\emph{Data-only discrete learners}} \\
BOSS-BDeu & 0.00 / 0.24 / 0.36 & 0.02 / 0.35 / \textbf{0.57} \\
GRaSP-BDeu & 0.00 / 0.24 / \textbf{0.38} & 0.02 / 0.37 / 0.55 \\
PC & 0.18 / \textbf{0.26} / 0.30 & 0.09 / 0.31 / 0.36 \\
GES & 0.05 / 0.24 / 0.35 & 0.12 / 0.33 / 0.37 \\
Sparse-Cand & 0.00 / 0.14 / 0.13 & 0.00 / 0.27 / 0.30 \\
Structural EM & 0.00 / 0.14 / 0.12 & 0.00 / 0.26 / 0.37 \\
Adel-dC & 0.02 / 0.14 / 0.10 & 0.06 / 0.26 / 0.28 \\
Ours-NoPrior & 0.00 / 0.14 / 0.13 & 0.00 / 0.29 / 0.29 \\
\bottomrule
\end{tabular}
\end{table}

\paragraph{KG-quality boundary.}
We stress the KG along two axes. The first varies false-edge additions and targeted errors: dropping true edges, reversing true edges, or mixing drop/reverse/add errors. The second is a \emph{truth-independent} KG constructed separately from the true DAG, including a \texttt{random\_disjoint} variant that excludes all true edges. Figure~\ref{fig:robust} reports mean Directed-F1 across \texttt{child}/\texttt{insurance}/\texttt{sachs} at $\rho{=}0.4$. False-edge additions are relatively benign because true-edge recall is preserved; dropping or reversing true edges is more damaging because the KG loses or misorients the signal the prior is meant to supply.
The \texttt{random\_disjoint} KG falls back to the Ours-NoPrior floor (Directed-F1 $0.18$ vs.\ $0.19$), showing that arbitrary high-confidence edges do not by themselves create a directed-recovery advantage. These controls identify the operating condition: recovery is strongest when the KG carries meaningful structural signal, and it returns to the no-prior regime when that signal is absent.

The density-mismatch sweep ($\rho{=}0.4$, 5 seeds, same six networks as Table~\ref{tab:recovery}) separates precision and recall. With only half of the true edges retained in the KG, KG-SoftMAP reaches mean Directed-F1 $0.34$; with all true edges plus an equal number of false high-confidence edges (KG precision $0.50$), it reaches $0.58$, above the Ours-NoPrior floor ($0.21$). Even a severely over-dense KG with about six times the true edge count and precision $0.17$ retains Directed-F1 $0.31$. Thus the learner benefits most from KG recall, while precision shapes how much extra structure the search must filter. Full per-network results and KG construction are in the supplementary material.

\paragraph{Technical scope checks.}
Additional scope checks are reported in the supplementary material: a Markov-equivalence-aware prior leaves the main recovery pattern largely unchanged, the controlled MAR EM extension is computationally heavier, ESS and prior-strength sweeps show no narrow tuning dependence, and runtime remains practical at the SAF scale.
These checks support scope and reproducibility; the main empirical claim rests on the recovery, mechanism, provenance, and KG-quality checks above.

\subsection{Track~2: Real-Data Evaluation}\label{sec:track2}

Track~2 examines real sparse educational data, where no ground-truth DAG exists, along three axes: prediction, calibration, and KG-consistency.
The learned BN is a reusable concept-level posterior model: it exposes an inspectable concept graph and supports calibrated held-out \textsc{Fail} prediction once observed partial evidence is available.
For BN prediction, we use variable elimination (VE): for each held-out observed concept cell, the BN conditions on the other observed concept states in the same response row and returns the posterior \textsc{Fail} probability.
The missingness patterns differ across datasets.
SAF is closer to planned missingness induced by question--concept coverage~\citep{graham2006planned}, ASSISTments involves self-selected skill practice, and Eedi has broad diagnostic coverage.
We benchmark against strong discriminative references: logistic regression (LR), a multilayer perceptron (MLP), and XGBoost.
For each target concept, each discriminative model is trained on training-fold response rows where that concept is observed; the features encode the other observed concept states in the same row, and the label is the target concept's cell state.
The three datasets span KG-provenance regimes within education.
SAF uses an end-to-end LLM-extracted KG, ASSISTments provides weak heuristic KG signal, and Eedi supplies an independent expert ontology that weakly favors edges between related concepts.

\subsubsection{SAF Full-Pipeline Evaluation}
SAF is our primary end-to-end case study: a real, LLM-extracted concept KG paired with a $\rho{\approx}4.4\%$ observation matrix.
We use two SAF artifacts for two distinct purposes.
\textbf{SAF-eval} is the fixed 191-concept artifact for prediction and calibration; its KG and evaluation artifact are frozen in advance (specification in the supplementary material) and built from training-split reference material only.
\textbf{SAF-full} is the original 238-concept extraction used for descriptive KG-consistency diagnostics.
In each response-level, row-disjoint CV fold on SAF-eval, KG-SoftMAP learns the BN on training rows.
We then apply the VE protocol above to each held-out concept cell.
LR, MLP, and XGBoost use this per-concept cell-prediction protocol on the same folds and held-out cells as VE (Table~\ref{tab:saf_pred}).
In SAF-eval each response identifier appears once, so the split measures row-disjoint response-record generalization; student-disjoint splitting is outside this fixed artifact.

The SAF prediction comparison shows what the BN adds in its intended role.
KG-SoftMAP+VE reaches F1\textsubscript{FAIL} $0.75$ versus $0.78$ for logistic regression while returning a reusable BN rather than only per-cell classifiers.
Compared with a direct-parent predictor that conditions only on the target's parents, VE conditions on all other observed concepts in the same response row; this all-evidence protocol improves F1\textsubscript{FAIL} from $0.68$ to $0.75$ at a mild calibration cost (ECE $0.015 \to 0.019$, still comparable to LR's $0.015$).
The posterior-query capability is clearest in the restricted-evidence evaluation (Table~\ref{tab:saf_partial_main}).
The same learned BN is reused without retraining while VE conditions on only $K$ observed concepts from the response row.
The clean comparison is the fixed-cell rise from $K{=}0$ to $K{=}1$ to $K{=}2$ on the same $11{,}637$ evaluation cells; it shows that a small amount of observed concept evidence carries most of the signal, and calibration remains low once any evidence is observed.
The $K{=}3$ row slightly exceeds all-evidence VE; this may reflect strong observed concepts saturating the signal, but the $K$-restricted rows use different eligible cell sets and should not be read as a monotone curve.
Two SAF-specific checks support this evaluation.
A reference-provenance pilot perturbs the reference material given to the extractor.
Replaying the correct references reproduces the KG (Jaccard $0.98$), whereas shuffled or absent references change it substantially ($0.21$ and $0.28$); VE prediction nevertheless remains stable (F1\textsubscript{FAIL} $0.75$--$0.77$).
On SAF, the KG mainly shapes the interpretable structure and KG-consistency of the learned graph, while the fitted likelihood carries the cell-prediction signal, consistent with the ASSISTments control.
A no-new-LLM voting sweep keeps the observed mask fixed and changes only voting constants; small threshold changes are within $0.002$ F1\textsubscript{FAIL}, and reducing the mistake boost leaves metrics identical.

\begin{table}[t]
\centering
\caption{SAF-eval prediction on the fixed 191-concept artifact: response-level, row-disjoint $5$-fold CV with identical evaluation cells; KG-SoftMAP structure is relearned per fold on training rows. Subscript F denotes the \textsc{Fail} class; Mac-F1 is macro-F1; Time is mean per-fold wall-clock inference time in seconds. Best values are bolded, with lower ECE and Time better.}
\label{tab:saf_pred}
\small
\setlength{\tabcolsep}{1.0pt}
\begin{tabular}{lrrrrrr}
\toprule
Predictor & Acc. & Mac-F1 & F1\textsubscript{F} & PR-AUC\textsubscript{F} & ECE\textsubscript{F} & Time \\
\midrule
BN direct-parent & 0.915 & 0.846 & 0.676 & 0.774 & \textbf{0.015} & \textbf{$<0.1$} \\
KG-SoftMAP+VE    & 0.945 & 0.895 & 0.751 & 0.833 & 0.019 & 1.6 \\
LogReg       & \textbf{0.954} & \textbf{0.908} & \textbf{0.780} & \textbf{0.866} & \textbf{0.015} & 13.6 \\
MLP                       & 0.948 & 0.900 & 0.765 & 0.846 & 0.017 & 48.6 \\
XGBoost                   & 0.886 & 0.798 & 0.650 & 0.677 & 0.038 & 26.9 \\
\bottomrule
\end{tabular}
\end{table}

\begin{table}[t]
\centering
\caption{SAF-eval posterior queries with restricted evidence using the same learned BN. VE conditions on exactly $K$ observed concepts from the same response row; ``All'' uses every other observed concept.}
\label{tab:saf_partial_main}
\small
\setlength{\tabcolsep}{4.5pt}
\begin{tabular}{lrrr}
\toprule
Evidence used & Eval cells & F1\textsubscript{FAIL} & ECE\textsubscript{FAIL} \\
\midrule
0 concepts & 11{,}637 & 0.205 & 0.025 \\
1 concept  & 11{,}637 & 0.682 & 0.019 \\
2 concepts & 11{,}637 & 0.734 & 0.014 \\
3 concepts & 11{,}430 & 0.767 & 0.012 \\
5 concepts & 9{,}936  & 0.762 & 0.013 \\
All evidence & 11{,}637 & 0.751 & 0.019 \\
\bottomrule
\end{tabular}
\end{table}

\paragraph{Interpretability and KG consistency.}
On SAF-eval, the learned BN retains local chains such as \textsc{DHCP} $\rightarrow$ \textsc{DHCP server} $\rightarrow$ \textsc{IP address} (labels shortened), giving an inspectable concept graph for posterior queries.
The quantitative evidence for this capability is the restricted-evidence evaluation above: one learned BN answers target-cell queries from different evidence subsets without retraining.
A fixed discriminative model can be trained for one target and one feature design, but it does not by itself provide the full family of posterior queries over arbitrary target concepts and evidence subsets.
On SAF-full descriptive diagnostics, KG-SoftMAP retains $72.7\%$ of KG edges with zero direction conflicts while staying parsimonious; the data-only Sparse-Cand learner attains higher data fit (Log\,P(D$\mid$G,$\theta$); full values in the supplementary material) but introduces $27$ KG-direction conflicts. These diagnostics come from the original full-scope artifact; the prediction comparison above uses SAF-eval.

\subsubsection{Real-Data Generalization Checks}
ASSISTments supplies weak heuristic KG signal through name-similarity, random, and null priors.
All KG conditions perform indistinguishably (F1\textsubscript{FAIL}$\approx0.13$), confirming the expected weak-KG boundary: the soft prior contributes little when the KG carries no structural signal.
The low absolute BN performance (VE $0.14$ vs.\ LR $0.58$) appears in the null-prior setting as well and is consistent with the self-selected practice regime, where observation is tied to practice choices rather than independent sampling.
Eedi supplies an independent, non-LLM expert subject ontology. We use it as a soft relatedness prior over leaf-level knowledge components ($249$ KCs, $\rho{\approx}13\%$): it weakly favors edges between components that share an ontology topic, without imposing prerequisite directions. This prior measurably increases the share of \emph{within-topic} edges in the learned structure (from $0.045$ under a flat prior to $0.123$), pulling the structure toward taxonomy-coherent \emph{relatedness}. Because the ontology is a containment taxonomy rather than a prerequisite graph, this validates relatedness consistency only; recovering prerequisite \emph{direction} would require richer ontology-aware priors. LR again leads on cell-level prediction (F1\textsubscript{FAIL} $0.43$ vs.\ BN $0.20$).

Together, SAF, ASSISTments, and Eedi give a coherent real-data picture: a meaningful KG supports a reusable posterior concept model, weak heuristic KGs correctly provide no artificial gain, and an independent ontology moves the learned structure toward expert relatedness, as expected from a taxonomy source rather than a prerequisite graph.
Additional partial-evidence, provenance, voting-sensitivity, SAF-full, ASSISTments, and Eedi tables are in the supplementary material.

\section{Conclusion}\label{sec:conclusion}

We presented KG-SoftMAP, a KG-MAP scoring framework for discrete Bayesian-network structure learning under extreme sparsity.
The experiments delimit the operating range of a finite-strength KG prior.
Useful edge-set signal is necessary, separated confidence bands are not, and KGs that lose or misorient true edges can fall back toward the no-prior regime.
LLM-provenance checks show the same scoring mechanism using graph signal extracted from text rather than generated from the true DAG.
The supplementary full-rate grid also shows that available-case scoring can behave non-monotonically at intermediate observation rates, so the main recovery claim is tied to the sparse grid reported in Table~\ref{tab:recovery}.
On SAF, this posterior model reaches F1\textsubscript{FAIL} $0.75$ versus $0.78$ for logistic regression while answering target-cell queries from arbitrary observed concept subsets without retraining.
External graph knowledge should be scored as weighted structural evidence whose usefulness can be tested; it should not be imposed as hard constraints or discarded when data are too sparse to speak alone.

\subsection*{Limitations}

The main structural evidence comes from synthetic benchmarks with known DAGs; the real-data studies provide within-education evidence through prediction, calibration, and KG-consistency.
The main synthetic KG is idealized by construction, with separated confidence ranges for true and false edges.
The overlapping-weight, corruption, density-mismatch, and NovelGraphs checks cover progressively less idealized KG regimes.
Our synthetic masks are missing-completely-at-random, and available-case scoring is not guaranteed unbiased under the informative missingness plausible in self-selected practice data; the controlled MAR extension in the supplement is a first step rather than a full treatment.
Natural next steps are evaluation in non-educational domains, scaling beyond a few hundred variables, and provenance-controlled KG extraction on additional real-world corpora beyond SAF.

\bibliography{kg_softmap_refs}

\end{document}


\maketitle


\section{Derivation of the KG-SoftMAP Framework}\label{app:derivation}

This appendix provides full derivations for both stages of the KG-SoftMAP framework: knowledge graph construction and MAP-based structure learning with the KG prior.

\subsection{Stage 1: Knowledge Graph Construction}

The knowledge graph (KG) encodes domain relations as weighted directed edges $\mathcal{K} = \{(u, v, w_{uv})\}$, where $w_{uv} \in [0,1]$ is a soft belief that $u$ should precede $v$ in the Bayesian network. When no curated KG is available, we extract one from domain reference material using a single \emph{structured} (schema-constrained) LLM call per document, rather than a separate query per candidate pair.

For SAF, each question's reference answer is passed to the LLM (\texttt{gpt-4o-mini}, temperature~$0$), which returns, in one JSON response, a set of concepts and a set of directed prerequisite edges, each with an extracted confidence weight $w_{uv} \in [0,1]$ (the full prompt and schema are in Appendix~\ref{app:prompts}). A second schema-constrained call per response record returns the concept-level mistakes used in SAF preprocessing.

Extracted candidates pass \emph{programmatic} validation only: syntactic concept filtering, a vocabulary constraint on mistakes, and an acyclicity check on the prerequisite edges. No manual or expert curation is applied. Edges whose confidence falls below a threshold $\delta$ (default $0.3$) are discarded. Determinism comes from temperature~$0$; the rebuilt pipeline also hashes prompts so identical requests are reused deterministically.

\subsection{SAF Data Representation and Voting Rule}\label{app:saf_voting_rule}

\paragraph{Data Representation.}
For SAF, preprocessing builds an $N \times p$ matrix $\mathbf{D}$ over $N$ instances (graded responses) and $p$ concepts.
Each entry $D_{ic} \in \{0, 1, 2, \texttt{NaN}\}$ is a three-state concept label:
\textsc{Master}~(0), \textsc{Unsure}~(1), or \textsc{Fail}~(2).
Since each concept may be assessed by multiple questions, we aggregate evidence by weighted voting.

Each question $q$ covers a set of concepts $\mathcal{C}_q$ (its \emph{scope}, extracted by the LLM in Stage~1).
When instance~$i$ has a graded answer to question~$q$ with score $s_{iq} \in [0,1]$, the score is first mapped to a base state:
\begin{equation}\label{eq:base_state}
\bar{s}_{iq} = \begin{cases}
\textsc{Fail}   & \text{if } s_{iq} < t_\ell, \\
\textsc{Master} & \text{if } s_{iq} > t_u, \\
\textsc{Unsure} & \text{otherwise},
\end{cases}
\end{equation}
with thresholds $t_\ell {=} 0.3$ and $t_u {=} 0.7$.
The LLM may also extract concept-specific mistakes; each links to one concept $c$ with a severity weight $m_{iqc} \in [0,1]$.

For each concept $c \in \mathcal{C}_q$, a vote vector $\mathbf{v}_{iqc} \in \mathbb{R}^3$ over (\textsc{M}, \textsc{U}, \textsc{F}) is constructed as follows.
Let $\mathbf{e}_k$ denote the one-hot vector for state $k$ and let $\tau_m {=} 0.5$ be the strong-mistake threshold.
If no mistake targets $c$, the vote is simply $\mathbf{v}_{iqc} = \mathbf{e}_{\bar{s}_{iq}}$ (pure score evidence).
If a mistake with weight $m_{iqc}$ is present:
\begin{equation}
\mathbf{v}_{iqc} = \begin{cases}
(0,\; 0,\; 1.5)
  & \text{if } m_{iqc} \geq \tau_m, \\[3pt]
(0.8,\; 0.2,\; 0)
  & \text{if } m_{iqc} < \tau_m \;\wedge\; \bar{s}_{iq} {=} \textsc{M}, \\[3pt]
\mathbf{e}_{\bar{s}_{iq}} + m_{iqc} \cdot \mathbf{e}_{\textsc{F}}
  & \text{if } m_{iqc} < \tau_m \;\wedge\; \bar{s}_{iq} {\neq} \textsc{M}.
\end{cases}
\end{equation}
The first case overrides the score when a strong mistake is found: the concept is labeled \textsc{Fail} with boosted weight 1.5.
The second case weakens a \textsc{Master} verdict when a mild mistake is present, shifting some mass to \textsc{Unsure}.
The third case adds \textsc{Fail} mass proportional to the mistake weight on top of the base vote.

The LLM may also detect mistakes on concepts outside the question scope ($c \notin \mathcal{C}_q$).
These receive a \textsc{Fail}-only vote: $\mathbf{v}_{iqc} {=} (0,0,1.5)$ if $m_{iqc} \geq \tau_m$, or $(0,0,m_{iqc})$ otherwise.

Votes accumulate across all questions that provide evidence for instance~$i$:
\begin{equation}\label{eq:vote_agg}
\mathbf{V}_{ic} = \sum_{q \in \mathcal{Q}_{ic}} \mathbf{v}_{iqc},
\end{equation}
where $\mathcal{Q}_{ic}$ is the set of questions that provide evidence about concept $c$ for instance~$i$ (either through scope or through a detected mistake).
The final label is $D_{ic} = \arg\max_k V_{ic,k}$ when $\mathcal{Q}_{ic} \neq \varnothing$, and $\texttt{NaN}$ otherwise.
On SAF-full (the original full-scope $238$-concept extraction) this yields an observation rate of approximately $4.5\%$, with state distribution \textsc{Fail}\,=\,12.3\%, \textsc{Master}\,=\,68.4\%, \textsc{Unsure}\,=\,19.3\%. These are original-extraction descriptive diagnostics; the prediction evaluation instead uses SAF-eval, the separately rebuilt 191-concept evaluation artifact (Appendix~\ref{app:saf_cache}).

\subsection{MAP Objective}
Given data $\mathcal{D}$ and a candidate DAG $G$, the MAP estimate maximizes:
\begin{equation}\label{eq:map}
G^* = \arg\max_{G \in \mathcal{G}} \; \log P(\mathcal{D} \mid G) + \log P(G).
\end{equation}
The first term is the BDeu score~\citep{heckerman1995learning}, the marginal likelihood that integrates out parameters under a Dirichlet prior with equivalent sample size $\alpha$ on complete data. Under missingness, we evaluate each family on rows where the child and its current parents are jointly observed, giving a family-wise available-case BDeu surrogate under MCAR:
\begin{equation}
\log P(\mathcal{D} \mid G) = \sum_{j=1}^{p} \mathrm{BDeu}(X_j, \mathrm{Pa}_G(X_j); \alpha).
\end{equation}
The second term is the structure prior $\log P(G)$, which we now define using the knowledge graph.

\subsection{Logit-Form Edge Prior}

Let $w_{uv} \in [0,1]$ be the confidence weight of edge $u \to v$ in the knowledge graph (KG).
Edges not in the KG have $w_{uv} = 0$.
We define a per-edge inclusion probability via a \emph{logit-linear} mapping: the log-odds of inclusion are linear in the KG confidence weight:
\begin{equation}
\mathrm{logit}(\theta_{uv}) = \beta_0 + \beta_1\, w_{uv}, \qquad \theta_{uv} = \sigma(\beta_0 + \beta_1\, w_{uv}),
\end{equation}
where $\sigma$ is the logistic function, $\beta_0 = \mathrm{logit}(\theta_0)$, $\beta_1 = \mathrm{logit}(\theta_{\max}) - \beta_0$, $\theta_0 \in (0, 0.5]$ is the base inclusion probability for non-KG edges ($w_{uv} = 0$), and $\theta_{\max} \in (\theta_0, 1)$ is the inclusion probability for full-confidence KG edges ($w_{uv} = 1$). This matches the logit equation in the main text.

Assuming edge inclusions are independent given the KG, the structure prior factorizes as:
\begin{equation}
\begin{aligned}
\log P(G) = \sum_{(u,v)} \big[
& \mathbb{1}[(u \to v) \in G] \log \theta_{uv} \\
&+ \mathbb{1}[(u \to v) \notin G] \log(1 - \theta_{uv})
\big].
\end{aligned}
\end{equation}

\subsection{Properties of the Parameterization}

The logit form has three useful properties.

First, when $\theta_0 = \theta_{\max} = 0.5$, the prior reduces to a uniform (flat) prior over edges, recovering the standard BDeu objective.
This corresponds to Ours-NoPrior in our experiments.

Second, as $\theta_{\max} \to 1$, the prior assigns increasingly high probability to graphs that include KG edges, approaching a hard constraint.
The setting $\theta_{\max} = 0.99$ corresponds to Ours-HardPrior.

Third, the log-odds ratio between including and excluding a KG edge with weight $w_{uv}$ is:
\begin{equation}
\log \frac{\theta_{uv}}{1 - \theta_{uv}} - \log \frac{\theta_0}{1 - \theta_0},
\end{equation}
which increases monotonically in $w_{uv}$.
This means the prior favors high-confidence KG edges more strongly than low-confidence ones, and the data likelihood can still override the prior when evidence is strong enough.

\subsection{Adaptive Equivalent Sample Size}

Under extreme sparsity, some nodes have very few complete observations relative to their number of parent configurations.
The effective per-cell count $\alpha / (q_j \cdot K_j)$ (where $q_j$ is the number of parent configurations and $K_j$ is the number of states) can become very small, causing numerical instability in the BDeu score.

We define an adaptive-ESS guard for this case:
\begin{equation}\label{eq:adaptive_ess}
\alpha_{\mathrm{eff},j} = \alpha \cdot \min\!\left(C_{\max},\; \max\!\left(1,\; \frac{q_j}{\max(n_j,1)}\right)\right),
\end{equation}
where $n_j$ is the number of complete observations for node $j$ and its parents, and $C_{\max} = 5$ is a cap that prevents the effective ESS from growing too large.
The factor is clamped to $[1, C_{\max}]$, so the ESS is only ever scaled \emph{up}: when $n_j < q_j$ it increases the prior weight to stabilize the Dirichlet (up to the cap $C_{\max}$), and when $n_j \ge q_j$ the factor is~1 and standard BDeu applies.

This is not a standard Bayesian marginal likelihood: it modifies the ESS based on the observed data count, which can change the relative scoring of different parent sets when it engages.
We use it only during structure search.
Once the DAG is fixed, CPD estimation (Eq.~\ref{eq:cpd_est}) uses the standard (non-adaptive) $\alpha_{\mathrm{pred}}$.
On SAF-eval, where the max in-degree is~2, the adaptive scaling never engages because $q_j$ is small. More generally, under the default scoring gate the scaling is dormant: a family is scored only when its complete-case count $n_j \ge \max(5,q_j) \ge q_j$, so $q_j/\max(n_j,1)\le 1$, the clamped factor $\min(C,\max(1,q_j/\max(n_j,1)))$ equals~1, and $\alpha_{\mathrm{eff}}=\alpha$. An on/off ablation is identical in Directed-F1 and SHD across the synthetic grid (including $\rho{=}0.05$) and yields the identical learned graph on SAF-eval; the scaling engages only under relaxed gates that score families with $n_j<q_j$ (Appendix~\ref{app:ess}).

\subsection{BDeu Score Under Missingness}

The standard BDeu score for node $X_j$ with parent set $\mathrm{Pa}_j$ and $K_j$ states is:
\begin{equation}\label{eq:bdeu_full}
\begin{aligned}
\mathrm{BDeu}(X_j, \mathrm{Pa}_j)
= \sum_{q=1}^{q_j}\bigg[
&\log\frac{\Gamma(\alpha'_q)}{\Gamma(\alpha'_q + n_q)} \\
&+ \sum_{k=1}^{K_j}
\log\frac{\Gamma(\alpha'_{qk} + n_{qk})}{\Gamma(\alpha'_{qk})}
\bigg],
\end{aligned}
\end{equation}
where $q_j = \prod_{X_i \in \mathrm{Pa}_j} K_i$ is the number of parent configurations, $\alpha'_q = \alpha_{\mathrm{eff},j}/q_j$, $\alpha'_{qk} = \alpha'_q / K_j$, $n_{qk}$ is the count of instances where $\mathrm{Pa}_j = q$ and $X_j = k$ among complete cases, and $n_q = \sum_k n_{qk}$.
Under sparse data, we compute counts only from rows where both $X_j$ and all variables in $\mathrm{Pa}_j$ are observed (available-case analysis).
This avoids imputation bias and reduces the effective sample size; the dormant guard in Eq.~\ref{eq:adaptive_ess} is relevant only if the scoring gate is relaxed enough to admit very low-count families.

\subsection{Greedy Search Procedure}

KG-SoftMAP uses a two-phase greedy hill-climbing search with three move types: edge addition, edge \emph{swap} (replacing one parent of a node with a different parent when its parent set is full), and edge deletion; it does not use edge reversal.
A \emph{forward} phase repeatedly applies the best-scoring legal add or swap (preserving acyclicity and the in-degree limit) until no such move increases the combined MAP score $\log P(\mathcal{D} \mid G) + \log P(G)$; a subsequent \emph{backward} phase repeatedly deletes the edge whose removal most improves the score, until none does.
The procedure terminates after no improving move remains in the implemented forward pass and then in the implemented backward pass; another search schedule could in principle find a further move, so the guarantee is finite monotone improvement for this implementation.

The combined score decomposes by node.
For an edge addition $u \to v$, only the local score of $v$ changes.
The score delta is:
\begin{equation}
\begin{aligned}
\Delta(u \to v)
=&\ \mathrm{BDeu}(X_v, \mathrm{Pa}_v \cup \{u\}) \\
&- \mathrm{BDeu}(X_v, \mathrm{Pa}_v)
+ \log\frac{\theta_{uv}}{1 - \theta_{uv}} .
\end{aligned}
\end{equation}
The last term is the prior bonus for including an edge supported by the KG.
For non-KG edges ($w_{uv} = 0$), this bonus is $\log(\theta_0 / (1 - \theta_0)) < 0$, penalizing inclusion.
For high-confidence KG edges, the bonus is large and positive, biasing the search toward KG-consistent structures.

A swap move uses the analogous delta with prior term $\log\frac{\theta_{uv}}{1 - \theta_{uv}} - \log\frac{\theta_{u_{\mathrm{old}}v}}{1 - \theta_{u_{\mathrm{old}}v}}$, replacing parent $u_{\mathrm{old}}$ by $u$. After the forward phase, the backward phase removes edges whose deletion improves the score.
This prevents over-inclusion of spurious edges during early search iterations when the graph is still sparse and the likelihood surface is flat.

\subsection{Proof of the Local Edge-Wise Discrimination Result}\label{app:edgewise_proof}

We prove the local edge-wise discrimination result stated in the main paper. The result is local: it concerns the score change for one eligible add move before global acyclicity conflicts and later search-order effects are considered.

Let $e=(u\to v)$ and let $S$ be the current parent set of $v$. The MAP add gain decomposes exactly as
\begin{equation}\label{eq:app_map_delta}
\Delta_{\mathrm{MAP}}(e;S)
= \Delta_{\mathrm{BDeu}}(e;S) + \omega_e,
\qquad
\omega_e=\log\frac{\theta_{uv}}{1-\theta_{uv}} .
\end{equation}
This identity is the main advantage of the logit parameterization: the prior contribution is the finite log-odds of including the candidate edge.
For absent KG edges, $\omega_e=\mathrm{logit}(\theta_0)=-\lambda_0$, where $\lambda_0=-\mathrm{logit}(\theta_0)>0$.

It remains to relate the BDeu gain to the local dependence signal. For a common complete-case comparison, applying Stirling's formula to the Gamma functions in the Dirichlet-multinomial score gives the standard local expansion
\begin{equation}\label{eq:app_bdeu_expansion}
\Delta_{\mathrm{BDeu}}(e;S)
=m_e\widehat I_e-\frac{\Delta\nu_e}{2}\log m_e+r_e ,
\end{equation}
where $\widehat I_e$ is the empirical conditional mutual information between $X_v$ and $X_u$ given $X_S$ on the scored rows,
$q_S=\prod_{z\in S}K_z$ is the number of current parent configurations,
$\Delta\nu_e=q_S(K_u-1)(K_v-1)$ is the number of added local parameters, and $r_e$ contains the Dirichlet finite-sample terms. More explicitly, the leading log-likelihood difference between the two local multinomial models is $m_e\widehat I_e$, and integrating the extra parameters contributes the $-(\Delta\nu_e/2)\log m_e$ term.

The implementation uses family-wise available-case scores. Adding a parent can change the set of rows used by the local score, so the comparison is not always a strict common-support comparison. The proposition therefore states the expansion on an event where
\[
\left|\Delta_{\mathrm{BDeu}}(e;S)
-\left(m_e\gamma_e-\frac{\Delta\nu_e}{2}\log m_e\right)\right|\le R_e ,
\]
with $R_e$ absorbing three finite-sample quantities: the Dirichlet remainder $r_e$, the deviation of $\widehat I_e$ from the scored-population dependence $\gamma_e=I(X_v;X_u\mid X_S)$, and the support-change remainder induced by available-case scoring. In a common-support comparison the last term is zero; under available-case scoring it is precisely the term that must be controlled.

Substituting this bound into Eq.~\eqref{eq:app_map_delta} gives
\[
\begin{aligned}
\Delta_{\mathrm{MAP}}(e;S)
&\ge m_e\gamma_e-\frac{\Delta\nu_e}{2}\log m_e-R_e+\omega_e,\\
\Delta_{\mathrm{MAP}}(e;S)
&\le m_e\gamma_e-\frac{\Delta\nu_e}{2}\log m_e+R_e+\omega_e .
\end{aligned}
\]
The sufficient condition for a positive add move is therefore
\[
m_e\gamma_e+\omega_e>\frac{\Delta\nu_e}{2}\log m_e+R_e,
\]
and the sufficient condition for a negative add move is
\[
m_e\gamma_e+\omega_e<\frac{\Delta\nu_e}{2}\log m_e-R_e.
\]
Setting $\gamma_e=0$ gives the false-edge cases in the local result. If the edge is KG-supported, $\omega_e$ can be positive and the edge is still rejected once
$\frac{\Delta\nu_e}{2}\log m_e>\omega_e+R_e$. If the edge is absent from the KG and $\omega_e\le-\lambda_0$, then
$\Delta_{\mathrm{MAP}}(e;S)<0$ whenever
$\lambda_0+\frac{\Delta\nu_e}{2}\log m_e>R_e$.
This completes the proof.
\begin{flushright}$\square$\end{flushright}

\subsection{Proof of the Complete-Case Availability Threshold}\label{app:cc_threshold_proof}

For the complete-case availability result in the main text, consider adding an edge to a node whose current parent set has size $k$.
The after-add family contains the child, the candidate parent, and the $k$ existing parents.
Under independent MCAR$(\rho)$, all $k{+}2$ variables are observed in a row with probability $\rho^{k+2}$, so
$M_k \sim \mathrm{Binomial}(N,\rho^{k+2})$ and $\mu=N\rho^{k+2}$.
The threshold $\rho_{\min}(k)$ is the solution to $\mu=m_{\min}$.
When $\rho<\rho_{\min}(k)$, $\mu<m_{\min}$, and the multiplicative Chernoff upper-tail bound gives
\[
\begin{aligned}
\Pr(M_k\ge m_{\min})
&\le \exp\{-\mu h(m_{\min}/\mu)\},\\
h(x)&=x\log x-x+1 .
\end{aligned}
\]
This completes the proof.
\begin{flushright}$\square$\end{flushright}

\subsection{CPD Estimation and Bayesian Prediction}\label{app:cpd}

Given the learned DAG $G^*$, we estimate conditional probability distributions (CPDs) using the same Dirichlet posterior as the CPD equation in the main text.
For node $X_j$ with parent configuration $q$:
\begin{equation}\label{eq:cpd_est}
P(X_j {=} k \mid \mathrm{Pa}_j {=} q) = \frac{n_{qk} + \alpha'_{qk}}{n_q + \alpha'_q},
\end{equation}
where $\alpha'_{qk}$ and $\alpha'_q$ denote the standard Dirichlet pseudo-counts for that local conditional table.

For the held-out concept-cell prediction task, we predict the state of each observed concept $X_j$ in response record $r$, given the other observed concepts in that same record, $\mathbf{X}_{\mathrm{obs} \setminus j}^{(r)}$.
We use exact variable elimination over the Bayesian network:
\begin{equation}
P(X_j {=} k \mid \mathbf{X}_{\mathrm{obs} \setminus j}^{(r)}) = \frac{P(X_j {=} k, \mathbf{X}_{\mathrm{obs} \setminus j}^{(r)})}{P(\mathbf{X}_{\mathrm{obs} \setminus j}^{(r)})},
\end{equation}
where both terms are computed by summing over the unobserved variables using the chain rule of the Bayesian network.

For \textsc{Fail} detection, we use the posterior \textsc{Fail} probability as the score and the posterior mode as the predicted class:
\begin{equation}
\hat{y}_j^{(r)} = \arg\max_k P(X_j {=} k \mid \mathbf{X}_{\mathrm{obs} \setminus j}^{(r)}).
\end{equation}
For SAF-eval, \textsc{Fail} accounts for 12.2\% of observed concept cells.

\section{Full Synthetic Benchmark Results}\label{app:synthetic}

Table~\ref{tab:app_recovery_meansd} gives the seed variation behind the main sparse-recovery table.
Panel~A (discrete/BN baselines) uses the same six networks, masks, and methods as the main recovery table over $10$ random seeds; the continuous-optimization stress tests in Panel~B use the same $10$ seeds, with the KG-SoftMAP column repeated from Panel~A as a reference. The main table reports means only to keep the grid readable.

\begin{table*}[p]
\centering
\small
\caption{Full Directed-F1 values for the main sparse-recovery table. Panel A reports discrete/BN baselines (mean$\pm$SD over 10 seeds); Panel B reports continuous-optimization stress tests (mean$\pm$SD over 10 seeds; KG-SoftMAP column repeated from Panel A as a reference).}
\label{tab:app_recovery_meansd}
\centerline{\textbf{Panel A: discrete/BN baselines}}
\vspace{2pt}
\setlength{\tabcolsep}{2.0pt}
\begin{tabular}{llcccccccccc}
\toprule
Net & $\rho$ & KG-SoftMAP & GES & MMHC & PC-S & PC & StructEM & A-dC & SpCand & GRaSP & BOSS \\
\midrule
\texttt{cancer} & 0.05 & 0.19$\pm$0.25 & 0.00$\pm$0.00 & 0.00$\pm$0.00 & 0.00$\pm$0.00 & 0.00$\pm$0.00 & 0.36$\pm$0.27 & 0.36$\pm$0.27 & 0.00$\pm$0.00 & 0.04$\pm$0.13 & 0.04$\pm$0.13 \\
\texttt{cancer} & 0.20 & 0.62$\pm$0.09 & 0.00$\pm$0.00 & 0.00$\pm$0.00 & 0.00$\pm$0.00 & 0.00$\pm$0.00 & 0.12$\pm$0.19 & 0.12$\pm$0.19 & 0.22$\pm$0.08 & 0.04$\pm$0.13 & 0.04$\pm$0.13 \\
\texttt{cancer} & 0.40 & 0.54$\pm$0.10 & 0.03$\pm$0.11 & 0.03$\pm$0.11 & 0.00$\pm$0.00 & 0.00$\pm$0.00 & 0.11$\pm$0.18 & 0.11$\pm$0.18 & 0.26$\pm$0.07 & 0.12$\pm$0.19 & 0.12$\pm$0.19 \\
\addlinespace[1pt]
\texttt{asia} & 0.05 & 0.32$\pm$0.17 & 0.00$\pm$0.00 & 0.00$\pm$0.00 & 0.00$\pm$0.00 & 0.00$\pm$0.00 & 0.15$\pm$0.08 & 0.15$\pm$0.08 & 0.00$\pm$0.00 & 0.02$\pm$0.07 & 0.02$\pm$0.07 \\
\texttt{asia} & 0.20 & 0.66$\pm$0.09 & 0.02$\pm$0.07 & 0.02$\pm$0.07 & 0.00$\pm$0.00 & 0.00$\pm$0.00 & 0.22$\pm$0.20 & 0.19$\pm$0.15 & 0.26$\pm$0.09 & 0.32$\pm$0.14 & 0.32$\pm$0.14 \\
\texttt{asia} & 0.40 & 0.64$\pm$0.05 & 0.15$\pm$0.14 & 0.12$\pm$0.14 & 0.02$\pm$0.07 & 0.02$\pm$0.07 & 0.23$\pm$0.21 & 0.25$\pm$0.23 & 0.30$\pm$0.07 & 0.48$\pm$0.12 & 0.46$\pm$0.13 \\
\addlinespace[1pt]
\texttt{sachs} & 0.05 & 0.20$\pm$0.10 & 0.00$\pm$0.00 & 0.00$\pm$0.00 & 0.00$\pm$0.00 & 0.00$\pm$0.00 & 0.04$\pm$0.06 & 0.04$\pm$0.06 & 0.00$\pm$0.00 & 0.01$\pm$0.04 & 0.01$\pm$0.04 \\
\texttt{sachs} & 0.20 & 0.48$\pm$0.10 & 0.00$\pm$0.00 & 0.00$\pm$0.00 & 0.00$\pm$0.00 & 0.00$\pm$0.00 & 0.16$\pm$0.08 & 0.15$\pm$0.07 & 0.24$\pm$0.11 & 0.02$\pm$0.04 & 0.02$\pm$0.04 \\
\texttt{sachs} & 0.40 & 0.61$\pm$0.07 & 0.02$\pm$0.05 & 0.02$\pm$0.05 & 0.01$\pm$0.04 & 0.01$\pm$0.04 & 0.30$\pm$0.07 & 0.29$\pm$0.07 & 0.24$\pm$0.07 & 0.06$\pm$0.06 & 0.07$\pm$0.05 \\
\addlinespace[1pt]
\texttt{child} & 0.05 & 0.19$\pm$0.13 & 0.00$\pm$0.00 & 0.00$\pm$0.00 & 0.00$\pm$0.00 & 0.00$\pm$0.00 & 0.03$\pm$0.04 & 0.03$\pm$0.04 & 0.00$\pm$0.00 & 0.00$\pm$0.00 & 0.00$\pm$0.00 \\
\texttt{child} & 0.20 & 0.48$\pm$0.12 & 0.00$\pm$0.00 & 0.00$\pm$0.00 & 0.00$\pm$0.00 & 0.00$\pm$0.00 & 0.14$\pm$0.07 & 0.14$\pm$0.07 & 0.20$\pm$0.06 & 0.08$\pm$0.07 & 0.08$\pm$0.07 \\
\texttt{child} & 0.40 & 0.46$\pm$0.03 & 0.01$\pm$0.02 & 0.01$\pm$0.02 & 0.01$\pm$0.02 & 0.01$\pm$0.02 & 0.38$\pm$0.10 & 0.38$\pm$0.09 & 0.25$\pm$0.05 & 0.33$\pm$0.06 & 0.30$\pm$0.07 \\
\addlinespace[1pt]
\texttt{insurance} & 0.05 & 0.21$\pm$0.07 & 0.00$\pm$0.00 & 0.00$\pm$0.00 & 0.00$\pm$0.00 & 0.00$\pm$0.00 & 0.03$\pm$0.02 & 0.03$\pm$0.02 & 0.00$\pm$0.00 & 0.01$\pm$0.02 & 0.01$\pm$0.02 \\
\texttt{insurance} & 0.20 & 0.44$\pm$0.06 & 0.00$\pm$0.00 & 0.00$\pm$0.00 & 0.00$\pm$0.01 & 0.00$\pm$0.01 & 0.13$\pm$0.06 & 0.15$\pm$0.07 & 0.17$\pm$0.07 & 0.08$\pm$0.06 & 0.08$\pm$0.06 \\
\texttt{insurance} & 0.40 & 0.47$\pm$0.06 & 0.01$\pm$0.02 & 0.01$\pm$0.02 & 0.01$\pm$0.02 & 0.01$\pm$0.02 & 0.23$\pm$0.05 & 0.22$\pm$0.07 & 0.19$\pm$0.04 & 0.27$\pm$0.04 & 0.30$\pm$0.06 \\
\addlinespace[1pt]
\texttt{alarm} & 0.05 & 0.20$\pm$0.06 & 0.00$\pm$0.00 & 0.00$\pm$0.00 & 0.00$\pm$0.00 & 0.00$\pm$0.00 & 0.04$\pm$0.03 & 0.04$\pm$0.03 & 0.00$\pm$0.00 & 0.01$\pm$0.03 & 0.01$\pm$0.03 \\
\texttt{alarm} & 0.20 & 0.49$\pm$0.05 & 0.00$\pm$0.00 & 0.00$\pm$0.00 & 0.00$\pm$0.00 & 0.00$\pm$0.00 & 0.13$\pm$0.04 & 0.16$\pm$0.05 & 0.19$\pm$0.04 & 0.12$\pm$0.05 & 0.12$\pm$0.05 \\
\texttt{alarm} & 0.40 & 0.50$\pm$0.03 & 0.00$\pm$0.00 & 0.00$\pm$0.00 & 0.00$\pm$0.00 & 0.00$\pm$0.00 & 0.23$\pm$0.05 & 0.22$\pm$0.09 & 0.26$\pm$0.05 & 0.25$\pm$0.07 & 0.24$\pm$0.04 \\
\bottomrule
\end{tabular}
\vspace{5pt}
\centerline{\textbf{Panel B: continuous-optimization stress tests}}
\vspace{2pt}
\setlength{\tabcolsep}{3.0pt}
\begin{tabular}{llcccc}
\toprule
Net & $\rho$ & KG-SoftMAP & GOLEM & DAGMA & SDCD \\
\midrule
\texttt{cancer} & 0.05 & 0.19$\pm$0.25 & 0.00$\pm$0.00 & 0.00$\pm$0.00 & 0.00$\pm$0.00 \\
\texttt{cancer} & 0.20 & 0.62$\pm$0.09 & 0.00$\pm$0.00 & 0.00$\pm$0.00 & 0.00$\pm$0.00 \\
\texttt{cancer} & 0.40 & 0.54$\pm$0.10 & 0.00$\pm$0.00 & 0.00$\pm$0.00 & 0.15$\pm$0.25 \\
\addlinespace[1pt]
\texttt{asia} & 0.05 & 0.32$\pm$0.17 & 0.00$\pm$0.00 & 0.00$\pm$0.00 & 0.00$\pm$0.00 \\
\texttt{asia} & 0.20 & 0.66$\pm$0.09 & 0.00$\pm$0.00 & 0.00$\pm$0.00 & 0.09$\pm$0.10 \\
\texttt{asia} & 0.40 & 0.64$\pm$0.05 & 0.04$\pm$0.09 & 0.02$\pm$0.07 & 0.14$\pm$0.13 \\
\addlinespace[1pt]
\texttt{sachs} & 0.05 & 0.20$\pm$0.10 & 0.00$\pm$0.00 & 0.00$\pm$0.00 & 0.06$\pm$0.10 \\
\texttt{sachs} & 0.20 & 0.48$\pm$0.10 & 0.00$\pm$0.00 & 0.00$\pm$0.00 & 0.13$\pm$0.11 \\
\texttt{sachs} & 0.40 & 0.61$\pm$0.07 & 0.00$\pm$0.00 & 0.00$\pm$0.00 & 0.30$\pm$0.07 \\
\addlinespace[1pt]
\texttt{child} & 0.05 & 0.19$\pm$0.13 & 0.00$\pm$0.00 & 0.00$\pm$0.00 & 0.00$\pm$0.00 \\
\texttt{child} & 0.20 & 0.48$\pm$0.12 & 0.00$\pm$0.00 & 0.00$\pm$0.00 & 0.12$\pm$0.08 \\
\texttt{child} & 0.40 & 0.46$\pm$0.03 & 0.02$\pm$0.03 & 0.02$\pm$0.04 & 0.25$\pm$0.09 \\
\addlinespace[1pt]
\texttt{insurance} & 0.05 & 0.21$\pm$0.07 & 0.00$\pm$0.00 & 0.00$\pm$0.00 & 0.06$\pm$0.06 \\
\texttt{insurance} & 0.20 & 0.44$\pm$0.06 & 0.00$\pm$0.00 & 0.00$\pm$0.00 & 0.09$\pm$0.07 \\
\texttt{insurance} & 0.40 & 0.47$\pm$0.06 & 0.02$\pm$0.03 & 0.03$\pm$0.03 & 0.17$\pm$0.07 \\
\addlinespace[1pt]
\texttt{alarm} & 0.05 & 0.20$\pm$0.06 & 0.00$\pm$0.00 & 0.00$\pm$0.00 & 0.00$\pm$0.00 \\
\texttt{alarm} & 0.20 & 0.49$\pm$0.05 & 0.01$\pm$0.02 & 0.01$\pm$0.02 & 0.06$\pm$0.05 \\
\texttt{alarm} & 0.40 & 0.50$\pm$0.03 & 0.05$\pm$0.03 & 0.06$\pm$0.02 & 0.13$\pm$0.03 \\
\bottomrule
\end{tabular}
\end{table*}

Table~\ref{tab:app_synthetic} reports per-network, per-observation-rate SHD for all 12 methods on the five small benchmark networks.
This extended grid comes from a separate 5-seed diagnostic run and broadens coverage beyond the main sparse grid; the pre-specified main-text rates ($\rho\in\{0.05,0.2,0.4\}$) are reported at 10 seeds in the main recovery table and Table~\ref{tab:app_recovery_meansd}.
We use the extended grid for per-rate diagnostics and method status, not for a monotonic trend claim across all observation rates.
The mid-range dips for KG-SoftMAP on some networks mark an available-case scoring regime where more candidate families pass the gate while family supports still differ across parent sets; noisy finite-sample BDeu gains can then over-add edges.
At $\rho{=}1.0$, all families share common support again and recovery rebounds.
Methods that fail the reported hard-constraint diagnostics in any run are marked ``---''.

\begin{table*}[t]
\centering
\caption{Full synthetic results: SHD (mean over 5 runs) for each network, method, and observation rate $\rho$.
Methods that fail hard constraints in any run are marked ``---''.}
\label{tab:app_synthetic}
\begin{tabular}{llrrrrrr}
\toprule
Network & Method & $\rho{=}0.05$ & $\rho{=}0.20$ & $\rho{=}0.40$ & $\rho{=}0.60$ & $\rho{=}0.80$ & $\rho{=}1.00$ \\
\midrule
\multirow{5}{*}{cancer (5/4)}
& KG-SoftMAP (Ours) & 3.2 & 4.2 & 0.8 & 6.4 & 4.4 & 1.4 \\
& Ours-HardPrior & 3.2 & --- & 0.8 & 6.2 & 4.2 & 0.4 \\
& Ours-NoPrior & 4.0 & --- & 4.0 & --- & --- & 4.0 \\
& GES & 4.0 & 4.0 & 3.8 & 4.2 & --- & --- \\
& StructEM & 4.0 & 4.0 & --- & 3.0 & --- & --- \\
\midrule
\multirow{6}{*}{asia (8/8)}
& KG-SoftMAP (Ours) & 6.6 & 6.4 & 0.8 & 12.4 & 8.8 & 0.6 \\
& Ours-HardPrior & 6.6 & 7.0 & 0.8 & --- & --- & 0.0 \\
& Ours-NoPrior & 8.0 & --- & 7.8 & --- & 15.6 & 8.0 \\
& GOLEM & --- & 7.8 & --- & --- & --- & --- \\
& MMHC & --- & --- & 7.6 & 8.4 & 7.0 & --- \\
& PC & --- & 7.8 & 7.6 & 8.0 & 7.6 & --- \\
\midrule
\multirow{6}{*}{sachs (11/17)}
& KG-SoftMAP (Ours) & 14.2 & 17.4 & 1.6 & 16.2 & 14.2 & 0.4 \\
& Ours-HardPrior & 14.2 & 18.4 & 1.6 & --- & --- & 0.4 \\
& Ours-NoPrior & 17.0 & --- & 17.0 & 25.8 & --- & 17.0 \\
& StructEM & --- & 15.6 & 14.4 & 13.4 & 8.2 & 7.8 \\
& Sparse-Cand & --- & 21.0 & --- & 22.6 & --- & 7.8 \\
& PC & --- & --- & 17.0 & --- & --- & --- \\
\midrule
\multirow{6}{*}{child (20/25)}
& KG-SoftMAP (Ours) & 23.4 & 31.2 & 2.0 & 51.6 & 45.2 & 4.0 \\
& Ours-HardPrior & 23.4 & --- & 2.0 & --- & --- & 3.4 \\
& Ours-NoPrior & 25.0 & --- & --- & --- & --- & 25.0 \\
& StructEM & --- & 25.4 & 19.4 & 16.2 & 18.6 & --- \\
& DAGMA & --- & --- & --- & --- & 32.2 & 15.4 \\
& MMHC & --- & --- & --- & 25.0 & 25.2 & 14.2 \\
\midrule
\multirow{6}{*}{insurance (27/52)}
& KG-SoftMAP (Ours) & 46.6 & 54.0 & 7.6 & 72.0 & 52.4 & 15.0 \\
& Ours-HardPrior & 46.4 & --- & 7.2 & 74.6 & 53.6 & 11.6 \\
& Ours-NoPrior & 52.0 & --- & --- & --- & --- & 52.0 \\
& StructEM & --- & 52.6 & 51.8 & 49.6 & 41.2 & --- \\
& MMHC & --- & --- & --- & --- & 54.0 & 32.8 \\
& PC & --- & --- & 52.2 & --- & 52.6 & --- \\
\bottomrule
\end{tabular}
\end{table*}

\section{Full Synthetic F1 Results (Skeleton and Directed)}\label{app:synthetic_f1}

Table~\ref{tab:app_synthetic_f1} reports both Skeleton F1 (SF1, ignoring edge direction) and Directed F1 (DF1) for representative methods across all observation rates, from the same separate 5-seed diagnostic run as Table~\ref{tab:app_synthetic}.
The three main-text rates are reported at 10 seeds in the main recovery table and Table~\ref{tab:app_recovery_meansd}; this table is retained to show orientation behavior across additional rates.
Each cell shows SF1\,/\,DF1.
DAG-producing methods (KG-SoftMAP, Ours-HardPrior) leave no edges undirected, but SF1 still exceeds DF1 whenever an edge is reversed relative to the truth (correct in the skeleton, wrong in direction); SF1\,=\,DF1 only when there are no such orientation errors.
For CPDAG or partially oriented methods (GES, PC, PC-Stable, StructEM), SF1 can substantially exceed DF1: these methods identify correct edges but leave many undirected, so the skeleton is more accurate than the directed graph.
The gap is largest at $\rho{=}1.0$, where data suffice to detect edges but orientation remains ambiguous.
The ALARM network is omitted here as Table~\ref{tab:app_alarm_full} provides a detailed SF1\,/\,DF1 breakdown.

\begin{table*}[t]
\centering
\caption{Full synthetic results: SF1\,/\,DF1 (mean over 5 runs). $^\dagger$CPDAG method (may leave edges undirected). For all methods, SF1$\,\geq\,$DF1; the gap reflects orientation errors (wrong or missing directions).}
\label{tab:app_synthetic_f1}
\begin{tabular}{llcccccc}
\toprule
Network & Method & $\rho{=}0.05$ & $\rho{=}0.20$ & $\rho{=}0.40$ & $\rho{=}0.60$ & $\rho{=}0.80$ & $\rho{=}1.00$ \\
\midrule
\multirow{4}{*}{cancer (5/4)}
& KG-SoftMAP (Ours) & .29\,/\,.29 & .60\,/\,.60 & .93\,/\,.88 & .55\,/\,.52 & .66\,/\,.61 & .77\,/\,.77 \\
& Ours-HardPrior     & .29\,/\,.29 & .59\,/\,.56 & .93\,/\,.88 & .57\,/\,.54 & .68\,/\,.64 & .94\,/\,.94 \\
& GES$^\dagger$       & .00\,/\,.00 & .00\,/\,.00 & .13\,/\,.07 & .13\,/\,.13 & .67\,/\,.19 & .89\,/\,.50 \\
& PC-Stable$^\dagger$ & .00\,/\,.00 & .00\,/\,.00 & .00\,/\,.00 & .00\,/\,.00 & .16\,/\,.00 & .65\,/\,.26 \\
\midrule
\multirow{4}{*}{asia (8/8)}
& KG-SoftMAP (Ours) & .28\,/\,.28 & .67\,/\,.67 & .95\,/\,.95 & .55\,/\,.55 & .65\,/\,.65 & .96\,/\,.96 \\
& Ours-HardPrior     & .28\,/\,.28 & .67\,/\,.67 & .95\,/\,.95 & .52\,/\,.52 & .65\,/\,.65 & 1.0\,/\,1.0 \\
& PC$^\dagger$        & .00\,/\,.00 & .04\,/\,.00 & .09\,/\,.09 & .25\,/\,.04 & .54\,/\,.20 & .72\,/\,.15 \\
& PC-Stable$^\dagger$ & .00\,/\,.00 & .04\,/\,.00 & .09\,/\,.09 & .13\,/\,.00 & .48\,/\,.11 & .78\,/\,.20 \\
\midrule
\multirow{4}{*}{sachs (11/17)}
& KG-SoftMAP (Ours) & .28\,/\,.28 & .50\,/\,.50 & .95\,/\,.95 & .65\,/\,.64 & .70\,/\,.69 & .99\,/\,.99 \\
& Ours-HardPrior     & .28\,/\,.28 & .47\,/\,.47 & .95\,/\,.95 & .64\,/\,.63 & .69\,/\,.68 & .99\,/\,.99 \\
& StructEM            & .00\,/\,.00 & .26\,/\,.16 & .58\,/\,.30 & .67\,/\,.35 & .88\,/\,.60 & .92\,/\,.59 \\
& PC-Stable$^\dagger$ & .00\,/\,.00 & .00\,/\,.00 & .04\,/\,.00 & .17\,/\,.04 & .50\,/\,.31 & .87\,/\,.27 \\
\midrule
\multirow{4}{*}{child (20/25)}
& KG-SoftMAP (Ours) & .14\,/\,.14 & .46\,/\,.46 & .96\,/\,.96 & .44\,/\,.44 & .49\,/\,.49 & .91\,/\,.91 \\
& Ours-HardPrior     & .14\,/\,.14 & .45\,/\,.45 & .96\,/\,.96 & .42\,/\,.42 & .49\,/\,.49 & .93\,/\,.93 \\
& StructEM            & .00\,/\,.00 & .27\,/\,.16 & .63\,/\,.37 & .75\,/\,.53 & .78\,/\,.43 & .85\,/\,.52 \\
& PC-Stable$^\dagger$ & .00\,/\,.00 & .00\,/\,.00 & .00\,/\,.00 & .06\,/\,.03 & .27\,/\,.12 & .87\,/\,.47 \\
\midrule
\multirow{4}{*}{insurance (27/52)}
& KG-SoftMAP (Ours) & .19\,/\,.19 & .44\,/\,.44 & .92\,/\,.92 & .48\,/\,.48 & .61\,/\,.61 & .83\,/\,.83 \\
& Ours-HardPrior     & .20\,/\,.20 & .42\,/\,.42 & .93\,/\,.93 & .47\,/\,.47 & .61\,/\,.61 & .87\,/\,.87 \\
& StructEM            & .00\,/\,.00 & .27\,/\,.13 & .42\,/\,.22 & .48\,/\,.33 & .62\,/\,.43 & .71\,/\,.47 \\
& PC-Stable$^\dagger$ & .00\,/\,.00 & .01\,/\,.00 & .01\,/\,.00 & .04\,/\,.02 & .15\,/\,.08 & .75\,/\,.46 \\
\bottomrule
\end{tabular}
\end{table*}

\section{Full ALARM Results with Skeleton F1}\label{app:alarm}

Table~\ref{tab:app_alarm_full} extends the main paper's ALARM table by adding KG retention, KG conflicts, and Skeleton F1.
The gap between Skeleton F1 (SF1, direction ignored) and Directed F1 (DF1) reveals how much of the error comes from mis-orientation versus missing edges.
For DAG-producing methods, SF1\,=\,DF1; for CPDAG methods (PC, PC-Stable$^\dagger$), SF1 can far exceed DF1 because undirected edges are counted as correct only in the skeleton.
The same pattern holds across the other five benchmark networks.

\begin{table*}[t]
\centering
\caption{Full ALARM results (37 nodes, 46 edges). Mean over 5 runs.
SF1\,=\,Skeleton F1 (undirected). DF1\,=\,Directed F1. KG Ret.\,=\,share of KG edges retained. KG Conf.\,=\,edges against KG direction. $^\dagger$CPDAG method.}
\label{tab:app_alarm_full}
\setlength{\tabcolsep}{1.0pt}
\begin{tabular}{lrrrrrrrrrrrrrrr}
\toprule
& \multicolumn{5}{c}{$\rho = 0.1$} & \multicolumn{5}{c}{$\rho = 0.5$} & \multicolumn{5}{c}{$\rho = 1.0$} \\
\cmidrule(lr){2-6}\cmidrule(lr){7-11}\cmidrule(lr){12-16}
Method & SHD & SF1 & DF1 & Ret. & Conf. & SHD & SF1 & DF1 & Ret. & Conf. & SHD & SF1 & DF1 & Ret. & Conf. \\
\midrule
Ours-NoPrior & 46.0 & .000 & .000 & .000 & 0.0 & 48.2 & .008 & .008 & .004 & 0.0 & 46.0 & .000 & .000 & .000 & 0.0 \\
Ours-HardPrior & 24.2 & .665 & .665 & .524 & 0.0 & 5.8 & .939 & .939 & .952 & 0.2 & 2.4 & .973 & .973 & .879 & 0.2 \\
KG-SoftMAP (Ours) & 24.0 & .667 & .667 & .520 & 0.2 & 5.4 & .943 & .943 & .944 & 0.2 & 2.4 & .973 & .973 & .879 & 0.2 \\
NNM+KG-SoftMAP & 46.0 & .000 & .000 & .000 & 0.0 & 13.4 & .755 & .755 & .748 & 0.2 & 2.4 & .973 & .973 & .879 & 0.2 \\
\midrule
GES & 46.0 & .000 & .000 & .000 & 0.0 & 46.2 & .017 & .009 & .004 & 0.2 & 33.6 & .834 & .502 & .509 & 17.0 \\
MMHC & 46.0 & .000 & .000 & .000 & 0.0 & 46.2 & .017 & .009 & .004 & 0.2 & 21.2 & .935 & .591 & .533 & 15.6 \\
PC$^\dagger$ & 46.0 & .000 & .000 & .000 & 0.0 & 46.4 & .017 & .000 & .000 & 0.4 & 25.3 & .808 & .608 & .530 & 8.7 \\
PC-Stable$^\dagger$ & 46.0 & .000 & .000 & .000 & 0.0 & 46.2 & .017 & .000 & .000 & 0.4 & 20.8 & .858 & .652 & .556 & 8.8 \\
StructEM & 58.6 & .106 & .050 & .032 & 2.0 & 52.2 & .602 & .257 & .236 & 16.0 & 31.6 & .870 & .458 & .432 & 19.4 \\
A-dC & 58.2 & .106 & .050 & .032 & 2.0 & 46.4 & .534 & .201 & .152 & 12.2 & 31.6 & .870 & .458 & .432 & 19.4 \\
Sparse-Cand & 70.2 & .190 & .098 & .080 & 3.8 & 86.4 & .482 & .308 & .440 & 12.6 & 32.6 & .867 & .443 & .420 & 20.0 \\
GOLEM & 46.0 & .000 & .000 & .000 & 0.0 & 47.2 & .025 & .017 & .008 & 0.2 & 56.0 & .365 & .198 & .162 & 6.6 \\
DAGMA & 46.0 & .000 & .000 & .000 & 0.0 & 47.4 & .025 & .016 & .008 & 0.2 & 41.8 & .632 & .276 & .214 & 13.6 \\
\bottomrule
\end{tabular}
\end{table*}

\section{Prediction: Full Results and Hyperparameters}\label{app:prediction}

\emph{Provenance (SAF-full descriptive diagnostics).} SAF-full is the original $238$-concept LLM extraction, used only for descriptive diagnostics. SAF-eval is an independently rebuilt $191$-concept artifact, fixed before any prediction evaluation and used in Table~4 of the main text. The two builds share the same dataset, prompts, and pipeline, so they are closely related; they are separate LLM-extraction runs and are not intended for cell-by-cell comparison. We retain the SAF-full results here only for completeness.

\subsection{Tuned Hyperparameters}

Table~\ref{tab:app_hyperparams} reports the best hyperparameters found by grid search on a validation split for each method.
The search space is: $\alpha \in \{0.1, 0.3, 1.0, 3.0, 10.0\}$, inference temperature $\tau \in \{0.5, 0.8, 1.0, 1.2\}$, and FAIL threshold $t_F \in \{0.3, 0.35, 0.4, 0.45, 0.5, \texttt{None}\}$.
When $t_F = \texttt{None}$, the argmax decision rule is used without threshold adjustment.

\begin{table*}[t]
\centering
\caption{Tuned prediction hyperparameters per method.}
\label{tab:app_hyperparams}
\begin{tabular}{lccc}
\toprule
Method & $\alpha$ & $\tau$ & $t_F$ \\
\midrule
Ours-NoPrior & 3.0 & 0.8 & 0.45 \\
Ours-HardPrior & 3.0 & 0.8 & 0.45 \\
KG-SoftMAP (Ours) & 3.0 & 0.8 & 0.50 \\
GES & 3.0 & 0.8 & None \\
MMHC & 0.3 & 1.0 & 0.45 \\
PC-Stable & 0.3 & 1.0 & 0.45 \\
PC & 3.0 & 0.8 & 0.45 \\
StructEM & 3.0 & 1.0 & None \\
A-dC & 0.3 & 1.0 & 0.45 \\
Sparse-Cand & 3.0 & 0.8 & 0.50 \\
GOLEM & 3.0 & 0.8 & 0.45 \\
DAGMA & 3.0 & 0.8 & 0.45 \\
\bottomrule
\end{tabular}
\end{table*}

\subsection{Full LOO and CV Results}

Table~\ref{tab:app_prediction_full} reports the full SAF-full predictive diagnostics, including the Lift metric, which measures the improvement in FAIL detection over a random baseline. These figures are from the original $238$-concept extraction and are a different run from the SAF-eval fixed-artifact evaluation in the main paper; see the provenance note above.
LOO is computed on the top 50 concepts with $\geq 10$ observations (7{,}451 predictions).
CV uses 5 folds at the response level (row-disjoint; 5{,}596 predictions per fold).

\begin{table*}[t]
\centering
\caption{SAF-full predictive diagnostics (original 238-concept extraction). LOO on top 50 concepts; 5-fold response-level (row-disjoint) CV. Best per column in \textbf{bold}.}
\label{tab:app_prediction_full}
\begin{tabular}{lrrrrrrrrr}
\toprule
& \multicolumn{7}{c}{Leave-One-Out (50 concepts, 7{,}451 predictions)} & \multicolumn{2}{c}{5-Fold CV (5{,}596 pred.)} \\
\cmidrule(lr){2-8}\cmidrule(lr){9-10}
Method & Acc. & F1\textsubscript{FAIL} & Macro-F1 & ECE\textsubscript{F} & Brier\textsubscript{F} & Lift & $N_c$ & Acc. & Macro-F1 \\
\midrule
\multicolumn{10}{l}{\emph{Internal variants}} \\
Ours-NoPrior & \textbf{0.855} & 0.576 & 0.702 & \textbf{0.081} & 0.070 & 0.161 & 50 & 0.632 & 0.524 \\
Ours-HardPrior & 0.850 & \textbf{0.628} & \textbf{0.717} & 0.089 & 0.069 & 0.156 & 50 & 0.678 & 0.564 \\
KG-SoftMAP (Ours) & 0.846 & 0.603 & 0.699 & 0.084 & \textbf{0.069} & 0.152 & 50 & 0.598 & 0.448 \\
\midrule
\multicolumn{10}{l}{\emph{External baselines}} \\
GES & 0.767 & 0.207 & 0.453 & 0.030 & 0.105 & 0.073 & 50 & 0.728 & 0.548 \\
MMHC & 0.746 & 0.217 & 0.427 & 0.008 & 0.106 & 0.052 & 50 & 0.689 & 0.484 \\
PC & 0.706 & 0.111 & 0.335 & 0.001 & 0.110 & 0.012 & 50 & 0.661 & 0.362 \\
PC-Stable & 0.688 & 0.079 & 0.301 & 0.001 & 0.112 & $-$0.005 & 50 & 0.657 & 0.355 \\
StructEM & 0.847 & 0.513 & 0.670 & 0.079 & 0.077 & 0.153 & 50 & \textbf{0.769} & \textbf{0.631} \\
A-dC & 0.731 & 0.202 & 0.397 & 0.010 & 0.099 & 0.037 & 50 & 0.733 & 0.556 \\
Sparse-Cand & 0.844 & 0.606 & 0.712 & 0.099 & 0.077 & 0.150 & 50 & 0.647 & 0.476 \\
GOLEM & 0.759 & 0.169 & 0.432 & 0.036 & 0.108 & 0.065 & 50 & 0.704 & 0.467 \\
DAGMA & 0.735 & 0.188 & 0.400 & 0.030 & 0.108 & 0.041 & 50 & 0.689 & 0.462 \\
\bottomrule
\end{tabular}
\end{table*}

\section{Constraint Satisfaction Rates}\label{app:constraints}

Table~\ref{tab:app_constraints_saf} reports whether each method satisfies the hard and soft constraints on SAF-full.
Hard constraints are acyclicity and max in-degree $\leq 4$.
Soft constraints add KG retention $\geq 0.10$, edge density $\leq 0.20$, and KG conflicts $= 0$.
These SAF-full figures are original-extraction descriptive diagnostics ($238$ concepts), closely related to the separately rebuilt SAF-eval artifact used for the main-text evaluation (Appendix~\ref{app:saf_cache}).

\begin{table*}[t]
\centering
\caption{Constraint satisfaction on SAF-full (original 238-concept extraction). $\checkmark$ = pass, $\times$ = fail.}
\label{tab:app_constraints_saf}
\begin{tabular}{lccccc}
\toprule
Method & Acyclic & InDeg $\leq 4$ & KG Ret. $\geq 0.10$ & Density $\leq 0.20$ & KG Conf. $= 0$ \\
\midrule
Ours-NoPrior & $\checkmark$ & $\checkmark$ & $\checkmark$ & $\checkmark$ & $\times$ \\
Ours-HardPrior & $\checkmark$ & $\checkmark$ & $\checkmark$ & $\checkmark$ & $\checkmark$ \\
KG-SoftMAP (Ours) & $\checkmark$ & $\checkmark$ & $\checkmark$ & $\checkmark$ & $\checkmark$ \\
\midrule
GES & $\checkmark$ & $\checkmark$ & $\times$ & $\checkmark$ & $\checkmark$ \\
MMHC & $\checkmark$ & $\checkmark$ & $\times$ & $\checkmark$ & $\checkmark$ \\
PC & $\checkmark$ & $\checkmark$ & $\times$ & $\checkmark$ & $\times$ \\
PC-Stable & $\checkmark$ & $\checkmark$ & $\times$ & $\checkmark$ & $\checkmark$ \\
StructEM & $\checkmark$ & $\checkmark$ & $\times$ & $\checkmark$ & $\times$ \\
A-dC & $\checkmark$ & $\checkmark$ & $\times$ & $\checkmark$ & $\checkmark$ \\
Sparse-Cand & $\checkmark$ & $\checkmark$ & $\times$ & $\checkmark$ & $\times$ \\
GOLEM & $\checkmark$ & $\checkmark$ & $\times$ & $\checkmark$ & $\checkmark$ \\
DAGMA & $\checkmark$ & $\checkmark$ & $\times$ & $\checkmark$ & $\checkmark$ \\
\bottomrule
\end{tabular}
\end{table*}

Table~\ref{tab:app_constraints_alarm} reports constraint satisfaction rates on ALARM across observation rates.
Since ALARM experiments use 5 random seeds, we report the fraction of runs that pass each constraint.

\begin{table*}[t]
\centering
\caption{Constraint satisfaction rates on ALARM (fraction of 5 runs passing). Hard = acyclicity $\wedge$ in-degree; Soft = hard $\wedge$ KG retention $\wedge$ density $\wedge$ no conflicts.}
\label{tab:app_constraints_alarm}
\begin{tabular}{lcccccc}
\toprule
& \multicolumn{2}{c}{$\rho = 0.1$} & \multicolumn{2}{c}{$\rho = 0.5$} & \multicolumn{2}{c}{$\rho = 1.0$} \\
\cmidrule(lr){2-3}\cmidrule(lr){4-5}\cmidrule(lr){6-7}
Method & Hard & Soft & Hard & Soft & Hard & Soft \\
\midrule
Ours-NoPrior & 1.0 & 0.0 & 1.0 & 0.0 & 1.0 & 0.0 \\
Ours-HardPrior & 1.0 & 1.0 & 1.0 & 1.0 & 1.0 & 1.0 \\
KG-SoftMAP (Ours) & 1.0 & 1.0 & 1.0 & 1.0 & 1.0 & 1.0 \\
\midrule
GES & 1.0 & 0.0 & 1.0 & 0.0 & 1.0 & 1.0 \\
MMHC & 1.0 & 0.0 & 1.0 & 0.0 & 1.0 & 1.0 \\
PC & 1.0 & 0.0 & 1.0 & 0.0 & 1.0 & 1.0 \\
PC-Stable & 1.0 & 0.0 & 1.0 & 0.0 & 1.0 & 1.0 \\
StructEM & 1.0 & 0.0 & 1.0 & 1.0 & 1.0 & 1.0 \\
A-dC & 1.0 & 0.0 & 1.0 & 1.0 & 1.0 & 1.0 \\
Sparse-Cand & 1.0 & 0.4 & 1.0 & 1.0 & 1.0 & 1.0 \\
GOLEM & 1.0 & 0.0 & 1.0 & 0.0 & 1.0 & 1.0 \\
DAGMA & 1.0 & 0.0 & 1.0 & 0.0 & 1.0 & 1.0 \\
\bottomrule
\end{tabular}
\end{table*}

\section{Full Ablation Results}\label{app:ablation}

Table~\ref{tab:app_ablation_full} provides the complete ablation results on SAF-full, expanding the sensitivity summary in the main paper with all 10 configurations ranked by F1\textsubscript{FAIL}.
These SAF-full ablation figures are original-extraction descriptive diagnostics ($238$ concepts), closely related to the separately rebuilt SAF-eval artifact used for the main-text evaluation (Appendix~\ref{app:saf_cache}).

\begin{table*}[t]
\centering
\caption{SAF-full ablation diagnostics, ranked by F1\textsubscript{FAIL}. Default: $\theta_0{=}0.01$, $\theta_{\max}{=}0.8$, $\alpha{=}1.0$, thresholds 0.3/0.7.}
\label{tab:app_ablation_full}
\begin{tabular}{lrrrrrr}
\toprule
Config & Edges & KG Ret. & Acc. & F1\textsubscript{F} & Time(s) & Group \\
\midrule
Aggressive (0.4/0.6) & 244 & 137 & 0.650 & 0.458 & 3.7 & Threshold \\
Default (0.01/0.8) & 248 & 143 & 0.644 & 0.444 & 4.2 & Prior \\
$\alpha{=}1.0$ (default) & 248 & 143 & 0.644 & 0.444 & 3.7 & ESS \\
Balanced (0.3/0.7) & 248 & 143 & 0.644 & 0.444 & 4.2 & Threshold \\
Weak KG (0.1/0.6) & 235 & 105 & 0.628 & 0.411 & 4.0 & Prior \\
Conservative (0.2/0.8) & 229 & 130 & 0.610 & 0.410 & 3.9 & Threshold \\
$\alpha{=}10.0$ & 376 & 180 & 0.652 & 0.387 & 5.0 & ESS \\
$\alpha{=}0.1$ & 218 & 94 & 0.629 & 0.385 & 3.7 & ESS \\
Strong KG (0.001/0.95) & 246 & 166 & 0.627 & 0.366 & 3.9 & Prior \\
No KG (0.5/0.5) & 441 & 78 & 0.622 & 0.359 & 6.3 & Prior \\
\bottomrule
\end{tabular}
\end{table*}

\section{LLM Prompts for KG Construction}\label{app:prompts}
We use \texttt{gpt-4o-mini} at temperature~$0$ for both relation extraction and mistake detection in Stage~1.
Each extraction is a single structured (schema-constrained) call; the confidence weights are the per-edge values returned in the structured output. Determinism comes from temperature~$0$; prompt hashing in the rebuilt pipeline lets identical requests be reused deterministically.

\subsection{Knowledge Graph Schema}

The pipeline operates over a predefined schema.
Node types: \texttt{Concept}, \texttt{Question}, \texttt{Mistake}, and \texttt{StudentPerformance}.
The key relation for Stage~2 is \texttt{PRECEDES} between concepts, which forms the KG edges entering the MAP objective.
\texttt{INDICATES} links mistakes to concepts and is used only during data preprocessing in the main text.
The remaining relations (\texttt{EVALUATES}, \texttt{ON\_QUESTION}, \texttt{HAS\_MISTAKE}) link questions and student performances for bookkeeping and do not enter the structure prior.
The full schema is:

\begin{arxivverbatim}
Node types:
  Concept, Question, Mistake, StudentPerformance

Relations:
  (Concept)  --PRECEDES-->      (Concept)
  (Mistake)  --INDICATES-->     (Concept)
  (Concept)  --EVALUATES-->     (Question)
  (StudentPerformance) --ON_QUESTION-->  (Question)
  (StudentPerformance) --HAS_MISTAKE-->  (Mistake)
\end{arxivverbatim}

\subsection{Prerequisite Relation Extraction}

This prompt is applied to each reference answer to extract concepts and prerequisite edges for the KG.

\begin{arxivverbatim}
You are extracting candidate educational knowledge
structures from a reference answer.

Hard requirements:
- Return STRICT JSON only. No markdown.
- Only use concepts that appear in the text or are
  standard technical terms clearly implied by it.
- Concepts must be informative (avoid placeholders
  like "A", "B", "Event A", "Node 1").
- Prerequisites must form a DAG within this reference
  (no cycles).

Output JSON schema:
{
  "concepts": ["..."],
  "prerequisites": [
    {"parent": "...", "child": "...",
     "confidence": 0.0-1.0}
  ]
}

Reference Answer Text:
{text}
\end{arxivverbatim}

\subsection{Mistake Detection}

This prompt is applied to each student response to identify concept-level mistakes.
The concept list is fixed to the vocabulary extracted in the previous step, ensuring all mistakes map to known KG variables.

\begin{arxivverbatim}
You are extracting mistakes from a student's answer
attempt, grounded in the reference answer.

Hard requirements:
- Return STRICT JSON only. No markdown.
- Do NOT output prerequisites here.
- Each mistake must be tied to EXACTLY ONE concept
  from the provided concept list.
- The "concept" field MUST be one of the strings in
  Concept List (exact match). If none apply, output
  an empty mistakes list.

Output JSON schema:
{
  "mistakes": [
    {
      "name": "...",
      "description": "...",
      "concept": "...",
      "weight": 0.0-1.0
    }
  ]
}

Concept List (allowed values):
{concept_list}

Question:
{question}

Reference Answer:
{reference}

Student Answer:
{student_answer}

Score (continuous in [0,1]):
{score}

Optional Feedback (may be empty):
{feedback}
\end{arxivverbatim}
\section{Implementation Manifest}\label{app:implementation}

Table~\ref{tab:implementation_manifest} summarizes the implementation choices used across the reported experiments. The table records the level needed to interpret the comparisons: input representation, KG usage, and where seeds or runtime are reported. Package versions and hardware are intentionally left to the released code environment rather than repeated in prose.

\begin{table*}[t]
\centering
\caption{Implementation manifest for the reported method groups. Runtime excludes KG construction and preprocessing unless a table states otherwise.}
\label{tab:implementation_manifest}
\small
\setlength{\tabcolsep}{3pt}
\begin{tabular}{p{0.16\textwidth}p{0.22\textwidth}p{0.20\textwidth}p{0.34\textwidth}}
\toprule
Method group & Implementation/source & Input representation & KG usage / reporting scope \\
\midrule
KG-SoftMAP and internal variants & Paper code & Native sparse discrete matrix & Synthetic/SAF/Eedi KG as specified; seeds in table captions \\
GES, MMHC, PC, PC-Stable, Sparse-Cand, StructEM, A-dC & Paper code following cited algorithms & Sparse or Dense-50 input as stated in the main setup & Data-only unless explicitly marked ``+Prior'' \\
GOLEM, DAGMA, NOTEARS, SDCD & Public package APIs used by experiment scripts & Mean-imputed, standardized full matrix & Data-only continuous structural-equation stress-test references \\
GRaSP and BOSS & \texttt{causal-learn} permutation-search APIs & Mode-imputed full integer matrix & Data-only BDeu references; CPDAG output converted to a DAG for scoring \\
\textsc{bnlearn}+\texttt{cs} & \textsc{bnlearn} inclusion-driven soft arc prior & Structural EM or mode-imputed complete matrix & Same KG endpoint probabilities as KG-SoftMAP; full protocol in Appendix~\ref{app:softprior} \\
ILS-CSL-softHC & Paper reimplementation of Ban et al.'s soft-constraint HC component & Native sparse discrete matrix & Same KG as pairwise soft constraints; 10-seed grid in Appendix~\ref{app:ilscsl_sparse} \\
NovelGraphs LLM-only outputs & Released NovelGraphs protocol outputs & Variable descriptions; no observational data & LLM-pairwise, LLM-BFS, and PromptBN graph quality; single graph per dataset \\
ReActBN & Paper implementation of Zhang et al.'s ReActBN workflow & NovelGraphs descriptions plus masked observational data & Data-aware LLM refinement; 5-seed sparse grid in Appendix~\ref{app:novographs} \\
NovelGraphs sparse learners & Paper scripts on NovelGraphs DAGs and masks & Native sparse or imputed inputs, matched to each learner & Same LLM-pairwise KG for KG-SoftMAP, MMHC-cand+Prior, and ILS-CSL-softHC; data-only rows receive no KG \\
SAF VE and discriminative predictors & \texttt{pgmpy} VE; scikit-learn / XGBoost classifiers & SAF-eval folds and held-out cells & Same evaluation cells; details in Appendix~\ref{app:saf_cache} \\
\bottomrule
\end{tabular}
\end{table*}

\section{Modern Baselines: Differentiable-DAG and Discrete Permutation Search}\label{app:modern}

This appendix expands the main-text recovery comparison with structural Hamming distance (SHD) and runtime/status details for the modern differentiable baselines. For a fair runnable comparison, the continuous methods (GOLEM, DAGMA, SDCD) receive a favorable representation: a \emph{mean-imputed}, standardized, full matrix (all rows and columns). KG-SoftMAP and Ours-NoPrior operate on the native discrete matrix. Means are over 10 seeds.

\begin{table*}[t]
\centering
\caption{SHD ($\downarrow$) on the modern-baseline grid (complements the Directed-F1 in the main recovery table; means over 10 seeds). SDCD runs under the adjusted preprocessing (float32 + standardization); it often adds many spurious edges while recovering few true ones.}
\label{tab:app_modern_shd}
\begin{tabular}{llccccc}
\toprule
Net & $\rho$ & KG-SoftMAP & NoPrior & GOLEM & DAGMA & SDCD \\
\midrule
\texttt{asia}  & 0.2 & \textbf{6.7}  & 13.3 & 8.1  & 8.1  & 10.3 \\
\texttt{asia}  & 0.4 & \textbf{0.7}  & 7.7  & 7.8  & 8.0  & 11.8 \\
\texttt{child} & 0.2 & 31.3 & 47.3 & \textbf{25.0} & \textbf{25.0} & 38.3 \\
\texttt{child} & 0.4 & \textbf{2.7}  & 26.0 & 24.8 & 24.7 & 38.9 \\
\texttt{alarm} & 0.2 & 61.0 & 102.3 & \textbf{45.8} & \textbf{45.8} & 60.7 \\
\texttt{alarm} & 0.4 & \textbf{8.0}  & 48.7 & 45.2 & 45.5 & 101.3 \\
\bottomrule
\end{tabular}
\end{table*}

On \texttt{alarm} at $\rho{=}0.2$, GOLEM and DAGMA attain lower SHD through near-empty graphs (Directed-F1 $\approx 0.01$ in the main recovery table), recovering little directed structure. Runtime is modest throughout (KG-SoftMAP $<1$\,s; DAGMA $\approx 1$\,s; GOLEM $\approx 2$--$4$\,s).

\paragraph{SDCD preprocessing and status in the sparse-discrete regime.}
With the default pipeline, SDCD~\citep{azizi2024sdcd} encounters two preprocessing issues on this discrete-relaxed data: on the sparse complete-case subset its internal train/validation split has an empty validation set, and on the mean-imputed full matrix its model raises a tensor dtype mismatch (float vs.\ double). Under an adjusted preprocessing pipeline, we mean-impute, drop constant columns, z-score standardize, cast to \texttt{float32}, and label all rows observational. With this adjustment, SDCD runs cleanly on every seed in the main sparse grid: for \texttt{asia}/\texttt{child}/\texttt{alarm} it attains Directed-F1 $0.09/0.12/0.06$ at $\rho{=}0.2$ and $0.14/0.25/0.13$ at $\rho{=}0.4$ (SHD $10.3/38.3/60.7$ and $11.8/38.9/101.3$; means over 10 seeds). These results characterize SDCD in this sparse-discrete preprocessing regime.

\emph{DiBS.} DiBS~\citep{lorch2021dibs} constructs correctly under the proper API (a graph model plus a \texttt{BGe} Gaussian likelihood model), then stops at the first JAX call: the installed JAX backend invokes \texttt{np.asarray(..., copy=...)} (a NumPy~2.0 API) that is incompatible with the environment's NumPy~1.26.4. Running DiBS would require environment-level dependency changes; it is runnable in principle (e.g., under NumPy${\,\ge\,}2.0$) and is the one modern baseline we leave attempted/deferred. These outcomes support including a discrete, sparsity-aware learner.

\subsection{Modern discrete permutation search: GRaSP and BOSS}\label{app:grasp}
To complement the differentiable baselines, which all assume continuous structural-equation models, we add two modern \emph{discrete} permutation score-search methods, GRaSP~\citep{lam2022grasp} and BOSS~\citep{andrews2023boss}, as \emph{data-only} baselines (no KG prior). We run the \texttt{causal-learn} implementations with the \texttt{local\_score\_BDeu} score. Because they require complete data, we give them the discrete analog of the differentiable baselines' input: a \emph{mode-imputed full integer matrix} (all rows and the full node set, categorical states preserved, no standardization), on the same masks and 10 seeds as the main recovery table. The CPDAG output (\texttt{causal-learn} \texttt{GeneralGraph}) is converted to a consistent acyclic DAG: directed edges are kept, and undirected edges are oriented greedily so as not to introduce a cycle. We then score the result with the same SHD/Directed-F1 evaluator. Table~\ref{tab:app_grasp} reports the full six-network support grid. GRaSP and BOSS are the strongest data-only baselines in several cells, especially at higher observation rates, but the informative-KG rows in the main recovery table remain well separated across the grid.

\begin{table*}[t]
\centering
\small
\caption{Modern discrete permutation score-search baselines GRaSP~\citep{lam2022grasp} and BOSS~\citep{andrews2023boss} (data-only, BDeu; means over 10 seeds). Both run on the mode-imputed full integer matrix over the full node set, with CPDAG output extended to a consistent acyclic DAG. This table gives the Directed-F1 and SHD support grid behind the GRaSP/BOSS columns in the main recovery table.}
\label{tab:app_grasp}
\begin{tabular}{llcccc}
\toprule
 & & \multicolumn{2}{c}{GRaSP} & \multicolumn{2}{c}{BOSS} \\
\cmidrule(lr){3-4}\cmidrule(lr){5-6}
Net & $\rho$ & Directed-F1 & SHD & Directed-F1 & SHD \\
\midrule
\texttt{cancer} & 0.05 & 0.04 & 4.0  & 0.04 & 4.0 \\
\texttt{cancer} & 0.20 & 0.04 & 3.9  & 0.04 & 3.9 \\
\texttt{cancer} & 0.40 & 0.12 & 3.7  & 0.12 & 3.7 \\
\addlinespace[1pt]
\texttt{asia} & 0.05 & 0.02 & 8.1  & 0.02 & 8.1 \\
\texttt{asia} & 0.20 & 0.32 & 7.0  & 0.32 & 7.0 \\
\texttt{asia} & 0.40 & 0.48 & 6.4  & 0.46 & 6.7 \\
\addlinespace[1pt]
\texttt{sachs} & 0.05 & 0.01 & 17.1 & 0.01 & 17.1 \\
\texttt{sachs} & 0.20 & 0.02 & 17.1 & 0.02 & 17.1 \\
\texttt{sachs} & 0.40 & 0.06 & 17.4 & 0.07 & 17.1 \\
\addlinespace[1pt]
\texttt{child} & 0.05 & 0.00 & 25.1 & 0.00 & 25.1 \\
\texttt{child} & 0.20 & 0.08 & 24.5 & 0.08 & 24.5 \\
\texttt{child} & 0.40 & 0.33 & 20.7 & 0.30 & 21.2 \\
\addlinespace[1pt]
\texttt{insurance} & 0.05 & 0.01 & 52.4 & 0.01 & 52.4 \\
\texttt{insurance} & 0.20 & 0.08 & 50.6 & 0.08 & 50.6 \\
\texttt{insurance} & 0.40 & 0.27 & 48.6 & 0.30 & 46.8 \\
\addlinespace[1pt]
\texttt{alarm} & 0.05 & 0.01 & 47.5 & 0.01 & 47.5 \\
\texttt{alarm} & 0.20 & 0.12 & 47.9 & 0.12 & 47.7 \\
\texttt{alarm} & 0.40 & 0.25 & 49.8 & 0.24 & 49.6 \\
\bottomrule
\end{tabular}
\end{table*}

\subsection{ILS-CSL-softHC Sparse Adaptation}\label{app:ilscsl_sparse}
ILS-CSL~\citep{ban2023ilscsl} provides the closest recent prior-guided BN structure-learning family. Its public workflow uses complete-data score caches and cached GPT constraints, so we implement the same softHC mechanism in the sparse setting used throughout this paper: the hill-climbing operator set is add/reverse/delete; local family scores are the same family-wise available-case BDeu scores used by KG-SoftMAP; and the same synthetic KG supplies pairwise soft constraints. The weighted variant maps each KG confidence $w$ to an ILS constraint confidence $\lambda=0.50+0.49w$; the near-hard variant uses $\lambda=0.99$ for every KG constraint. In both variants, a KG edge $u\!\to\!v$ is an obligatory soft constraint and the reverse direction is a forbidden soft constraint when the reverse edge is absent from the KG.

\begin{table*}[t]
\centering
\small
\caption{ILS-CSL-softHC sparse adaptation on the same synthetic masks and same $10\%$ noisy KG as the main recovery table (Directed-F1, mean$\pm$SD over 10 seeds). The adaptation receives the KG as pairwise soft constraints and uses the same sparse available-case BDeu local scores.}
\label{tab:app_ilscsl_sparse}
\begin{tabular}{llccc}
\toprule
Net & $\rho$ & KG-SoftMAP & ILS-CSL weighted & ILS-CSL-softHC \\
\midrule
\texttt{cancer} & 0.05 & 0.19$\pm$0.25 & 0.11$\pm$0.18 & 0.11$\pm$0.18 \\
\texttt{cancer} & 0.20 & 0.62$\pm$0.09 & 0.29$\pm$0.09 & 0.33$\pm$0.08 \\
\texttt{cancer} & 0.40 & 0.54$\pm$0.10 & 0.31$\pm$0.01 & 0.31$\pm$0.01 \\
\addlinespace[1pt]
\texttt{asia} & 0.05 & 0.32$\pm$0.17 & 0.21$\pm$0.15 & 0.24$\pm$0.14 \\
\texttt{asia} & 0.20 & 0.66$\pm$0.09 & 0.48$\pm$0.13 & 0.60$\pm$0.08 \\
\texttt{asia} & 0.40 & 0.64$\pm$0.05 & 0.40$\pm$0.05 & 0.44$\pm$0.07 \\
\addlinespace[1pt]
\texttt{sachs} & 0.05 & 0.20$\pm$0.10 & 0.09$\pm$0.08 & 0.14$\pm$0.08 \\
\texttt{sachs} & 0.20 & 0.48$\pm$0.10 & 0.31$\pm$0.07 & 0.38$\pm$0.11 \\
\texttt{sachs} & 0.40 & 0.61$\pm$0.07 & 0.32$\pm$0.09 & 0.38$\pm$0.08 \\
\addlinespace[1pt]
\texttt{child} & 0.05 & 0.19$\pm$0.13 & 0.10$\pm$0.08 & 0.12$\pm$0.10 \\
\texttt{child} & 0.20 & 0.48$\pm$0.12 & 0.22$\pm$0.05 & 0.33$\pm$0.11 \\
\texttt{child} & 0.40 & 0.46$\pm$0.03 & 0.22$\pm$0.05 & 0.29$\pm$0.05 \\
\addlinespace[1pt]
\texttt{insurance} & 0.05 & 0.21$\pm$0.07 & 0.12$\pm$0.08 & 0.15$\pm$0.07 \\
\texttt{insurance} & 0.20 & 0.44$\pm$0.06 & 0.20$\pm$0.04 & 0.28$\pm$0.05 \\
\texttt{insurance} & 0.40 & 0.47$\pm$0.06 & 0.19$\pm$0.03 & 0.23$\pm$0.04 \\
\addlinespace[1pt]
\texttt{alarm} & 0.05 & 0.20$\pm$0.06 & 0.09$\pm$0.03 & 0.12$\pm$0.05 \\
\texttt{alarm} & 0.20 & 0.49$\pm$0.05 & 0.28$\pm$0.06 & 0.37$\pm$0.05 \\
\texttt{alarm} & 0.40 & 0.50$\pm$0.03 & 0.27$\pm$0.06 & 0.32$\pm$0.06 \\
\bottomrule
\end{tabular}
\end{table*}

The ILS-CSL sparse adaptation benefits from the same KG signal and is consistently above the data-only floor. ILS-CSL-softHC is usually stronger than the confidence-weighted variant, especially at $\rho{=}0.2$, matching the expectation that high-quality pairwise constraints are useful when some data are available. KG-SoftMAP remains higher in aggregate but with a smaller high-$\rho$ margin after the corrected co-observation-count rerun: averaged over the six networks, its Directed-F1 is $0.22/0.53/0.54$ at $\rho{=}0.05/0.2/0.4$, compared with $0.15/0.38/0.33$ for ILS-CSL-softHC.

\paragraph{Additional benchmark-family networks.}
We also run the same sparse-recovery protocol on three additional bnlearn benchmark-family networks: \texttt{barley}, \texttt{mildew}, and \texttt{water}. These networks broaden the known-DAG benchmark coverage beyond the six networks in the main table and include the same sparse-adapted ILS-CSL variants. Table~\ref{tab:app_ilscsl_extended} shows the same pattern as the main grid: KG-SoftMAP gives partial recovery at $\rho{=}0.05$ and separates clearly by $\rho{=}0.4$, while ILS-CSL-softHC and the strongest data-only baselines recover less directed structure under the same sparse masks.

\begin{table*}[t]
\centering
\small
\caption{Additional benchmark-family networks under the same sparse-recovery protocol (Directed-F1, mean$\pm$SD over 3 seeds). All KG-guided methods receive the same $10\%$ noisy synthetic KG; NoPrior, Sparse-Cand, GRaSP/BOSS, and SDCD receive no KG.}
\label{tab:app_ilscsl_extended}
\setlength{\tabcolsep}{1.0pt}
\begin{tabular}{llcccccccc}
\toprule
Net & $\rho$ & SoftMAP & ILS-soft & ILS-wt & NoPrior & SpCand & GRaSP & BOSS & SDCD \\
\midrule
\texttt{barley} & 0.05 & 0.062$\pm$0.013 & 0.034$\pm$0.001 & 0.034$\pm$0.001 & 0.000$\pm$0.000 & 0.000$\pm$0.000 & 0.000$\pm$0.000 & 0.000$\pm$0.000 & 0.000$\pm$0.000 \\
\texttt{barley} & 0.20 & 0.259$\pm$0.027 & 0.133$\pm$0.041 & 0.100$\pm$0.037 & 0.062$\pm$0.006 & 0.091$\pm$0.028 & 0.016$\pm$0.014 & 0.016$\pm$0.014 & 0.093$\pm$0.022 \\
\texttt{barley} & 0.40 & 0.771$\pm$0.006 & 0.119$\pm$0.008 & 0.115$\pm$0.011 & 0.000$\pm$0.000 & 0.157$\pm$0.034 & 0.198$\pm$0.009 & 0.205$\pm$0.017 & 0.159$\pm$0.017 \\
\addlinespace[1pt]
\texttt{mildew} & 0.05 & 0.095$\pm$0.060 & 0.045$\pm$0.013 & 0.027$\pm$0.002 & 0.000$\pm$0.000 & 0.000$\pm$0.000 & 0.000$\pm$0.000 & 0.000$\pm$0.000 & 0.000$\pm$0.000 \\
\texttt{mildew} & 0.20 & 0.279$\pm$0.028 & 0.127$\pm$0.049 & 0.096$\pm$0.025 & 0.121$\pm$0.046 & 0.135$\pm$0.048 & 0.000$\pm$0.000 & 0.000$\pm$0.000 & 0.210$\pm$0.085 \\
\texttt{mildew} & 0.40 & 0.756$\pm$0.006 & 0.173$\pm$0.055 & 0.161$\pm$0.047 & 0.000$\pm$0.000 & 0.166$\pm$0.066 & 0.104$\pm$0.022 & 0.104$\pm$0.022 & 0.271$\pm$0.062 \\
\addlinespace[1pt]
\texttt{water} & 0.05 & 0.240$\pm$0.024 & 0.181$\pm$0.035 & 0.146$\pm$0.026 & 0.000$\pm$0.000 & 0.000$\pm$0.000 & 0.010$\pm$0.017 & 0.010$\pm$0.017 & 0.000$\pm$0.000 \\
\texttt{water} & 0.20 & 0.374$\pm$0.046 & 0.273$\pm$0.074 & 0.191$\pm$0.009 & 0.122$\pm$0.028 & 0.131$\pm$0.042 & 0.010$\pm$0.017 & 0.010$\pm$0.017 & 0.000$\pm$0.000 \\
\texttt{water} & 0.40 & 0.765$\pm$0.013 & 0.382$\pm$0.033 & 0.254$\pm$0.047 & 0.000$\pm$0.000 & 0.222$\pm$0.071 & 0.282$\pm$0.016 & 0.273$\pm$0.027 & 0.069$\pm$0.065 \\
\bottomrule
\end{tabular}
\end{table*}

\begin{table*}[t]
\centering
\caption{ILS-CSL-softHC same-KG mechanism slice at $\rho{=}0.4$ (Directed-F1, mean$\pm$SD over 10 seeds). This is the per-network support for the ILS-CSL-softHC row in the main mechanism table.}
\label{tab:app_ilscsl_mechanism}
\begin{tabular}{lcc}
\toprule
Net & 10\% KG noise & 30\% KG noise \\
\midrule
\texttt{cancer} & 0.31$\pm$0.01 & 0.31$\pm$0.01 \\
\texttt{asia} & 0.44$\pm$0.07 & 0.44$\pm$0.07 \\
\texttt{sachs} & 0.38$\pm$0.08 & 0.37$\pm$0.08 \\
\texttt{child} & 0.29$\pm$0.05 & 0.29$\pm$0.05 \\
\texttt{insurance} & 0.23$\pm$0.04 & 0.23$\pm$0.04 \\
\texttt{alarm} & 0.32$\pm$0.06 & 0.32$\pm$0.06 \\
\bottomrule
\end{tabular}
\end{table*}

\subsection{Near-Hard Prior Mechanism Check (HP-style on \texttt{sachs})}\label{app:hp_context}
This same-KG comparison isolates soft versus near-hard prior strength. All methods receive the same sparse masks and the same imperfect KG on \texttt{sachs}: true edges have weights $w\sim U(0.5,1.0)$, and $10\%$ false directed edges are added with weights $w\sim U(0,0.3)$. The HP-style variant uses the same search code as KG-SoftMAP and replaces the soft prior endpoints with near-hard endpoints $(\theta_0,\theta_{\max})=(10^{-4},0.9999)$.

\begin{table*}[t]
\centering
\caption{Near-hard prior mechanism check on \texttt{sachs} (Directed-F1 / SHD, mean over 5 seeds, $10\%$ KG noise). The near-hard HP-style variant is comparable to or slightly above KG-SoftMAP in this low-noise slice, showing that a high-quality KG can support harder constraints once enough data is available.}
\label{tab:app_hp_context}
\begin{tabular}{lcc}
\toprule
$\rho$ & KG-SoftMAP & HP-style near-hard \\
\midrule
0.05 & 0.28 / 14.2 & 0.28 / 14.2 \\
0.20 & 0.50 / 17.4 & \textbf{0.68} / \textbf{10.8} \\
0.40 & 0.95 / 1.6  & \textbf{0.96} / \textbf{1.2} \\
0.60 & 0.64 / 16.2 & \textbf{0.69} / \textbf{14.0} \\
0.80 & 0.69 / 14.2 & \textbf{0.72} / \textbf{12.6} \\
\bottomrule
\end{tabular}
\end{table*}

At $\rho{=}0.05$, the two prior-guided variants are identical: the data are too sparse to distinguish prior strengths. From $\rho{=}0.2$ to $0.8$, the HP-style near-hard variant is slightly ahead or close, which is expected when the KG is high quality and the data can check it. The result places KG-SoftMAP near constraint-style priors under a shared informative KG, and the soft formulation is most useful when KG fidelity is uncertain. This check isolates soft versus near-hard prior strength under one KG; a protocol-matched Ban et al.\ replication is a separate comparison.

\section{MAR-EM Extension: Full Grid}\label{app:marem}

This appendix gives the full grid behind the main-text MAR summary. We compare complete-case (CC) scoring with the EM extension, both with the soft KG prior, across MCAR and a controlled (known) MAR mechanism (a root variable kept observed; non-driver missingness a sigmoid in the driver's value, calibrated to $\rho$). Means over 5 seeds; Directed-F1.

\begin{table*}[t]
\centering
\caption{EM extension full grid (Directed-F1, mean over 5 seeds). CC = complete-case; EM = EM scoring; both with the soft KG prior. EM helps most at extreme sparsity and under MAR; it is $\approx$neutral where complete-case already suffices.}
\label{tab:app_marem_full}
\begin{tabular}{lcccc}
\toprule
 & \multicolumn{2}{c}{MCAR} & \multicolumn{2}{c}{MAR (controlled)} \\
\cmidrule(lr){2-3}\cmidrule(lr){4-5}
Net~/~$\rho$ & CC & EM & CC & EM \\
\midrule
\texttt{asia} ~0.2  & 0.74 & \textbf{0.88} & 0.83 & 0.82 \\
\texttt{asia} ~0.4  & 0.92 & 0.92 & 0.65 & \textbf{0.89} \\
\texttt{sachs}~0.2  & 0.45 & \textbf{0.70} & 0.51 & \textbf{0.85} \\
\texttt{sachs}~0.4  & 0.92 & 0.92 & 0.78 & \textbf{0.96} \\
\texttt{child}~0.2  & 0.45 & \textbf{0.68} & 0.53 & \textbf{0.83} \\
\texttt{child}~0.4  & 0.97 & 0.97 & 0.96 & 0.93 \\
\bottomrule
\end{tabular}
\end{table*}

EM is a robustness option with higher computational cost: it costs $15$--$40\times$ more compute (e.g., \texttt{child} EM $\approx$\,9--13\,s/run vs.\ $<1$\,s for complete-case) and uses hard-EM (MAP imputation). A soft expected-count Structural-EM is a natural extension. The MAR mechanism is synthetic and known, so this table characterizes controlled MAR-aware scoring. With a flat prior, both CC and EM collapse to the empty graph at these rates, indicating that the EM step still needs informative structural guidance in this sparse regime.

\paragraph{Structural-EM reference.}
As an additional reference, we ran \textsc{bnlearn}'s Structural EM with BDe scoring and a flat structure prior on \texttt{asia}, using the same MCAR/MAR masks and seeds as above. This comparison measures standard Structural EM in the same sparse setting without external KG signal. Table~\ref{tab:app_marem_sem_ref} shows DF1 $0.13$--$0.32$, while the KG-guided CC/EM variants remain substantially higher.

\begin{table*}[t]
\centering
\caption{Structural-EM reference on \texttt{asia} (Directed-F1, mean over 5 seeds). KG-SoftMAP-CC and KG-SoftMAP-EM use the same 10\% noisy informative KG as Table~\ref{tab:app_marem_full}; \textsc{bnlearn}-StructuralEM is a standard no-KG reference.}
\label{tab:app_marem_sem_ref}
\begin{tabular}{lccc}
\toprule
$\rho$ / mechanism & KG-SoftMAP-CC & KG-SoftMAP-EM & \textsc{bnlearn}-StructuralEM \\
\midrule
0.2 / MCAR & 0.74 & \textbf{0.88} & 0.31 \\
0.4 / MCAR & 0.92 & \textbf{0.92} & 0.30 \\
0.2 / MAR  & \textbf{0.83} & 0.82 & 0.13 \\
0.4 / MAR  & 0.65 & \textbf{0.89} & 0.32 \\
\bottomrule
\end{tabular}
\end{table*}

\section{KG Robustness: High-Confidence Filtering and Noise}\label{app:robust}

The main robustness figure reports structured-error robustness. Here we add a complementary sensitivity check: filtering the KG to high-confidence edges \emph{hurts} recovery in this diagnostic grid because it discards useful soft signal (Table~\ref{tab:app_highconf}). This table is used only to show the value of retaining graded weights; the corrected main sparse-recovery grid is Table~\ref{tab:app_recovery_meansd}.

\begin{table*}[t]
\centering
\caption{High-confidence KG filtering (Directed-F1, mean over 5 seeds). Using the full noisy KG is preferable to discarding low-confidence edges.}
\label{tab:app_highconf}
\begin{tabular}{lcccc}
\toprule
Net & Full KG & $\geq 0.7$ & $\geq 0.8$ & No prior \\
\midrule
\texttt{cancer}    & \textbf{0.76} & 0.69 & 0.48 & 0.00 \\
\texttt{asia}      & \textbf{0.87} & 0.78 & 0.59 & 0.04 \\
\texttt{sachs}     & \textbf{0.84} & 0.68 & 0.49 & 0.00 \\
\texttt{child}     & \textbf{0.86} & 0.72 & 0.59 & 0.00 \\
\texttt{insurance} & \textbf{0.80} & 0.70 & 0.55 & 0.00 \\
\bottomrule
\end{tabular}
\end{table*}

\section{Overlapping-Weight KG Check}\label{app:overlap_weights}

The main synthetic KG assigns true edges higher weights than false added edges. To isolate whether recovery depends on this separated confidence-band cue, we repeat the sparse-recovery experiment with the same high-recall KG edge set but draw both true and false edge weights from the same range. Thus the KG membership still carries structural signal, while the confidence weight no longer separates true from false edges. Table~\ref{tab:app_overlap_weights} shows that KG-SoftMAP and MMHC-cand+Prior remain far above the no-prior and data-only permutation baselines, especially once $\rho{\ge}0.20$. This check supports the main-text claim that calibrated confidence bands are not the only source of recovery; separate corruption and density-mismatch experiments test edge-set quality.

\begin{table*}[t]
\centering
\small
\setlength{\tabcolsep}{2.2pt}
\caption{Overlapping-weight KG check (Directed-F1, mean$\pm$SD over six network-level means). True and false KG edges receive weights from the same range, so weights do not distinguish true from false edges; the edge set still carries high-recall structural signal.}
\label{tab:app_overlap_weights}
\begin{tabular}{lccccc}
\toprule
\(\rho\) & KG-SoftMAP & MMHC-cand+Prior & Ours-NoPrior & GRaSP-BDeu & BOSS-BDeu \\
\midrule
0.05 & 0.23$\pm$0.07 & 0.23$\pm$0.07 & 0.00$\pm$0.00 & 0.02$\pm$0.03 & 0.03$\pm$0.03 \\
0.20 & 0.52$\pm$0.09 & 0.52$\pm$0.09 & 0.18$\pm$0.07 & 0.13$\pm$0.13 & 0.13$\pm$0.13 \\
0.40 & 0.51$\pm$0.08 & 0.52$\pm$0.07 & 0.21$\pm$0.06 & 0.26$\pm$0.14 & 0.25$\pm$0.13 \\
\bottomrule
\end{tabular}
\end{table*}

\section{Markov-Equivalence-Aware Prior: Details}\label{app:me}

Table~\ref{tab:app_me_twins} details the Markov-equivalence prior-fit check from the main KG-prior section. It lists the log-prior on Markov-equivalent twins for all networks with a covered edge (\texttt{cancer} has none); the edge-factored (EF) prior differs across a twin, while the ME-aware prior is exactly equal, as the proposition predicts (twins were confirmed Markov-equivalent by the pgmpy \texttt{is\_iequivalent} oracle). Three supporting experiments test whether this cleaner prior fit changes recovery: (A) on clean KGs across five networks at $\rho\in\{0.05,0.2,0.4\}$, ME and EF recovery are identical or within $|\Delta\mathrm{DF1}|<0.03$; (B) under reversed/mixed KG corruption, ME and EF remain within $|\Delta\mathrm{DF1}|<0.02$; and (C) the orientation weight $\lambda$ is inert in the clean regime (Directed-F1 flat at $0.953$ for \texttt{asia}, $0.952$ for \texttt{sachs} across $\lambda\in\{0,0.5,1,2,4\}$). The guarantee concerns the \emph{prior}; the posterior is not invariant (adaptive ESS and complete-case scoring are not score-equivalent).

\begin{table*}[t]
\centering
\caption{Log-prior on Markov-equivalent DAG twins. EF differs across a twin; ME-aware is equal.}
\label{tab:app_me_twins}
\begin{tabular}{lcccc}
\toprule
 & \multicolumn{2}{c}{EF prior} & \multicolumn{2}{c}{ME-aware} \\
\cmidrule(lr){2-3}\cmidrule(lr){4-5}
Net (covered edge) & $G_1$ & $G_2$ & $G_1$ & $G_2$ \\
\midrule
\texttt{asia}      & $-3.15$ & $-7.90$  & $4.20$  & $4.20$ \\
\texttt{sachs}     & $-2.70$ & $-6.25$  & $-2.70$ & $-2.70$ \\
\texttt{child}     & $-5.13$ & $-9.87$  & $13.27$ & $13.27$ \\
\texttt{insurance} & $-6.47$ & $-11.55$ & $43.91$ & $43.91$ \\
\bottomrule
\end{tabular}
\end{table*}

\section{ASSISTments: Full Results}\label{app:assist}

ASSISTments-2009 (skill-builder, corrected) is a negative control for KG provenance. It has 4{,}151 students over 84--110 skills (after a $\geq$50-student filter) at $\rho\approx 8.6\%$, with \emph{no} expert knowledge graph; the prior is therefore instantiated from null/random/name-similarity heuristics. Student-level $5$-fold CV.

\begin{table*}[t]
\centering
\caption{ASSISTments: KG-condition comparison (BN, direct-parent) and a discriminative reference. KG conditions show no meaningful predictive lift over the null prior; logistic regression is substantially stronger for prediction.}
\label{tab:app_assist}
\begin{tabular}{lccc}
\toprule
Condition / model & Acc. & F1\textsubscript{FAIL} & PR-AUC\textsubscript{FAIL} \\
\midrule
BN, null prior      & 0.608 & 0.132 & 0.200 \\
BN, random prior    & 0.609 & 0.127 & 0.202 \\
BN, name-similarity & 0.611 & 0.132 & 0.200 \\
\midrule
BN, corrected VE    & 0.597 & 0.143 & 0.208 \\
Logistic regression & \textbf{0.731} & \textbf{0.583} & \textbf{0.616} \\
\bottomrule
\end{tabular}
\end{table*}

A marginal base-rate predictor reaches only F1\textsubscript{FAIL}~$=0.088$, while logistic regression reaches $0.575$, confirming the task is learnable; the BN's limited gain from the prior is therefore attributable to the absence of a meaningful KG signal.

\section{Eedi Pilot: Full Results}\label{app:eedi}

Eedi (NeurIPS-2020 Education Challenge) provides an independent, non-LLM expert subject ontology. We use the leaf level of the ontology as knowledge components (KCs): $20{,}000$ students $\times$ $249$ KCs, $\rho\approx 12.8\%$, with a relatively balanced label distribution (\textsc{Master}~50.6\%, \textsc{Unsure}~25.5\%, \textsc{Fail}~23.9\%). The ontology prior connects KCs that share a Level-2 topic, so it tests taxonomy-coherent relatedness rather than prediction gain or prerequisite direction.

\begin{table*}[t]
\centering
\caption{Eedi KG-consistency: fraction of learned edges that fall within the same expert topic. The ontology prior pulls the structure toward taxonomy-coherent (relatedness) edges.}
\label{tab:app_eedi_consistency}
\begin{tabular}{lcc}
\toprule
KG condition & Within-topic edge fraction & Learned edges \\
\midrule
\textbf{ontology} & \textbf{0.123} & 480 \\
name-similarity   & 0.048 & 477 \\
null (data-only)  & 0.045 & 491 \\
random            & 0.024 & 435 \\
\bottomrule
\end{tabular}
\end{table*}

\begin{table*}[t]
\centering
\caption{Eedi prediction (student-level CV). KG conditions are similar in magnitude, and LR leads. Exact VE is intractable on the dense $249$-node data-only DAG, so the BN uses direct-parent inference here.}
\label{tab:app_eedi_pred}
\begin{tabular}{lccc}
\toprule
Predictor / condition & Acc. & Macro-F1 & F1\textsubscript{FAIL} \\
\midrule
BN (direct), ontology & 0.463 & 0.356 & 0.203 \\
BN (direct), null     & 0.458 & 0.357 & 0.201 \\
BN (direct), random   & 0.455 & 0.342 & 0.216 \\
Logistic regression   & \textbf{0.554} & \textbf{0.504} & \textbf{0.432} \\
\bottomrule
\end{tabular}
\end{table*}

\paragraph{Caveat (relatedness only).} Eedi supports a focused relatedness claim: an independent expert ontology used as a soft prior increases structural consistency with the taxonomy (within-topic edges). The ontology is a containment (is-a/part-of) taxonomy, so prerequisite direction and prediction gains remain outside this pilot; the ontology, null, and random priors are close, and LR leads. We report Eedi as a feasibility pilot for the expert-ontology-as-prior idea.

\begin{table*}[t]
\centering
\caption{Compact real-data summary moved from the main text. SAF prediction uses SAF-eval; SAF KG-consistency uses SAF-full. Eedi KG-consistency is the within-topic edge fraction under the ontology prior versus a flat prior.}
\label{tab:app_realdata_summary}
\small
\begin{tabular}{llll}
\toprule
Dataset / role & KG source & Prediction summary & KG-consistency summary \\
\midrule
SAF & LLM-extracted KG & VE 0.75 / LR 0.78 & retains KG; 0 conflicts \\
ASSISTments & heuristic weak KG & VE 0.14 / LR 0.58 & weak KG signal \\
Eedi & expert ontology & direct-parent 0.20 / LR 0.43 & within-topic 0.12 vs.\ 0.05 \\
\bottomrule
\end{tabular}
\end{table*}

\section{SAF-full KG-Consistency Diagnostics}\label{app:saf_structure}

Table~\ref{tab:saf_structure} reports descriptive KG-consistency diagnostics on SAF-full, the original 238-concept extraction.
The available structural reference is the extracted KG rather than a ground-truth DAG, so these diagnostics are used only to inspect graph alignment with the extracted knowledge source.

\begin{table*}[t]
\centering
\caption{SAF-full KG-consistency and fit (238-concept original extraction). KG Ret.\ = share of KG edges retained; KG Conf.\ = edges contradicting KG direction. Log-likelihood is shown for context; the structural diagnostic emphasizes KG-consistent, parsimonious structure.}
\label{tab:saf_structure}
\small
\begin{tabular}{lrrrr}
\toprule
Method & Edges & KG Ret. & KG Conf. & $\log\ell$ \\
\midrule
Ours-NoPrior      & 421 & 0.111 & 25 & $-$1{,}585 \\
Ours-HardPrior    & 477 & \textbf{0.823} & \textbf{0} & $-$1{,}496 \\
KG-SoftMAP (Ours) & 249 & 0.727 & \textbf{0} & $-$1{,}700 \\
Sparse-Cand       & 486 & 0.091 & 27 & \textbf{$-$1{,}380} \\
\bottomrule
\end{tabular}
\end{table*}

\section{SAF Evaluation Artifact and Inference Validation}\label{app:saf_cache}

\paragraph{Provenance (two SAF extraction runs).} SAF-full is the original $238$-concept extraction, used only for descriptive diagnostics. SAF-eval is an independently rebuilt $191$-concept artifact, fixed before any prediction evaluation and used in Table~4 of the main text. The two builds share the same dataset, prompts, and pipeline, so they are closely related; they are separate LLM-extraction runs and are not intended for cell-by-cell comparison. The SAF-full tables in this appendix (KG-consistency, prediction, constraints, and ablation in Appendices~\ref{app:saf_structure},~\ref{app:prediction},~\ref{app:constraints},~\ref{app:ablation}) are retained as \emph{descriptive diagnostics} only. No table mixes the two runs.

\paragraph{SAF-eval artifact counts.} The rebuilt graph (train split, gpt-4o-mini, temperature 0) has $26$ questions, $229$ extracted concepts ($191$ after pruning), $1{,}392$ graded responses (answer-level rows; score $\in[0,1]$ filter), and $184$ extracted \texttt{PRECEDES} edges ($159$ retained among the $191$ pruned concepts); the resulting matrix has $\rho\approx 4.4\%$ with label distribution \textsc{Master}~70.5\%, \textsc{Unsure}~17.3\%, \textsc{Fail}~12.2\%.

\paragraph{Evaluation-artifact manifest.} The artifact is self-contained (no database dependency at run time) and comprises: the response-by-concept matrix (3-state $+$ NaN); the state encoding with thresholds $0.3/0.7$; the concept list; the \texttt{PRECEDES} edges with confidence weights; the $5$-fold response-level, row-disjoint split indices (seed $42$); a full record of preprocessing parameters and provenance; summary counts; and SHA-256 checksums of every component. A second preprocessing mode (per-question score normalization that keeps all rows) is implemented but not run here; the evaluation uses the faithful replication mode.

\paragraph{Inference validation.} The corrected full-evidence variable-elimination predictor used for SAF-eval was validated independently of the BN learner: on a 3-node chain with random CPDs, its marginal $P(C\mid A)$ matches brute-force enumeration over the latent variable to within $10^{-6}$, and a d-separation blocking test confirms that conditioning on a mediator $B$ makes $P(C\mid A,B)$ equal to the CPD $P(C\mid B)$. This rules out implementation error in the VE predictor, isolating the direct-parent to VE change reported in the main text.

\paragraph{Partial-evidence posterior queries.} Table~\ref{tab:saf_partial_evidence} restricts the evidence available to the same learned BN. For each held-out concept cell, exact VE conditions on only $K$ observed concepts sampled from the same response row, with no retraining. The sharp gain from $K{=}0$ to $K{=}1$ shows that a small amount of observed concept evidence carries most of the signal; performance then improves through $K{=}3$, and calibration error is low once any evidence is observed. The $K{=}3$ and $K{=}5$ settings exclude rows with too few observed concepts, so their evaluation counts are smaller.

\begin{table*}[t]
\centering
\small
\caption{SAF-eval partial-evidence posterior queries. The same learned BN is used for all rows; only the number of observed evidence concepts changes.}
\label{tab:saf_partial_evidence}
\begin{tabular}{lrrrrr}
\toprule
Evidence $K$ & Eval cells & Acc. & F1\textsubscript{FAIL} & PR-AUC\textsubscript{F} & ECE\textsubscript{F} \\
\midrule
0 & 11637 & 0.747 & 0.205 & 0.350 & 0.025 \\
1 & 11637 & 0.933 & 0.682 & 0.785 & 0.019 \\
2 & 11637 & 0.944 & 0.734 & 0.819 & 0.014 \\
3 & 11430 & 0.952 & 0.767 & 0.850 & 0.012 \\
5 & 9936  & 0.959 & 0.762 & 0.827 & 0.013 \\
All & 11637 & 0.945 & 0.751 & 0.833 & 0.019 \\
\bottomrule
\end{tabular}
\end{table*}

\section{SAF Reference-Provenance Pilot}\label{app:saf_provenance}

This pilot varies the reference material used in Stage~1 while keeping the SAF-eval prediction protocol fixed.
The correct-reference replay nearly reproduces the current KG, whereas shuffled references and no-reference extraction produce visibly different graphs.
Prediction remains stable across these variants, which is useful for interpreting SAF: the prediction evaluation measures held-out concept-cell performance, while KG overlap measures how much the graph itself depends on the provided reference material.

\begin{table*}[t]
\centering
\small
\caption{SAF reference-provenance pilot. KG overlap is measured against the current SAF-eval KG; prediction uses exact VE on the same SAF-eval held-out cells.}
\label{tab:saf_provenance}
\begin{tabular}{lrrrrr}
\toprule
Condition & KG edges & Jaccard & F1\textsubscript{FAIL} & PR-AUC\textsubscript{FAIL} & ECE\textsubscript{FAIL} \\
\midrule
Current SAF-eval KG & 159 & 1.000 & 0.751 & 0.833 & 0.019 \\
Correct-reference replay & 156 & 0.981 & 0.751 & 0.833 & 0.019 \\
Shuffled-reference extraction & 119 & 0.209 & 0.764 & 0.848 & 0.016 \\
No-reference extraction & 119 & 0.275 & 0.767 & 0.849 & 0.015 \\
\bottomrule
\end{tabular}
\end{table*}

\section{SAF Stage-1 Voting Sensitivity}\label{app:saf_voting_sens}

This appendix tests whether the SAF-eval deployment result depends on the hand-specified Stage-1 discretization/voting constants: the score thresholds $0.3/0.7$, the strong-mistake threshold $\tau_m{=}0.5$, and the vote weights $1.5/0.8/0.2$ (full rule in Appendix~\ref{app:saf_voting_rule}). It is fully LLM-free: it re-runs only the Stage-1 vote on the fixed extracted observations and tunes nothing.

\paragraph{No-LLM raw export and reproduction gate.} The mistake annotations are already present in the rebuilt graph (extracted once during artifact construction), so we export the raw per-response observations, including grader score, question concept-scope, and concept-level mistake severities, with \emph{no} new LLM calls. Re-running the exact voting with the default constants on this export reproduces the fixed evaluation matrix (Appendix~\ref{app:saf_cache}) \emph{cell-for-cell} (NaN-aware) and reproduces the main-text SAF evaluation numbers (exact-VE F1\textsubscript{FAIL} $=0.751$). This gate confirms the offline re-implementation is faithful, so the variants below are directly comparable. All variants share an \emph{identical} observed mask ($11{,}638$ cells, $\rho=0.0438$): the constants change cell \emph{labels}, not which cells are observed.

\begin{table*}[t]
\centering
\small
\caption{SAF-eval Stage-1 voting sensitivity: label distribution over observed cells. All variants share the same observed mask ($11{,}638$ cells, $\rho=0.0438$); only labels change. ``\% changed'' is the fraction of co-observed cells whose label differs from the default. \textsf{score\_only} is degenerate (see findings), shown only for completeness.}
\label{tab:saf_voting_labeldist}
\begin{tabular}{lcccc}
\toprule
Variant & \textsc{Master} & \textsc{Unsure} & \textsc{Fail} & \% labels changed \\
\midrule
\textbf{current} (default) & 0.705 & 0.173 & 0.122 & --- \\
reduced\_boost ($1.5{\to}1.0$) & 0.705 & 0.173 & 0.122 & 0.00 \\
thresh $0.25/0.75$ & 0.571 & 0.340 & 0.090 & 16.65 \\
thresh $0.33/0.67$ & 0.706 & 0.171 & 0.123 & 0.27 \\
$\tau_m{=}0.4$ & 0.686 & 0.165 & 0.150 & 2.78 \\
$\tau_m{=}0.6$ & 0.710 & 0.175 & 0.115 & 0.71 \\
\textit{score\_only} (no mistakes; degenerate) & 0.740 & 0.179 & 0.081 & 4.07 \\
\bottomrule
\end{tabular}
\end{table*}

\begin{table*}[t]
\centering
\small
\caption{SAF-eval Stage-1 voting sensitivity: corrected full-evidence VE prediction (KG-SoftMAP structure, response-level row-disjoint $5$-fold CV; same eval cells across variants). The default ($0.3/0.7$, $\tau_m{=}0.5$, $1.5/0.8/0.2$) reproduces the main-text evaluation (F1\textsubscript{FAIL} $0.751$). Direct-parent and logistic-regression predictors follow the same pattern. \textsf{score\_only} is a degeneracy artifact: each answer-level row collapses to one label.}
\label{tab:saf_voting_pred}
\begin{tabular}{lcccccc}
\toprule
Variant & Acc & Macro-F1 & F1\textsubscript{FAIL} & PR-AUC\textsubscript{FAIL} & ECE\textsubscript{FAIL} & Brier\textsubscript{FAIL} \\
\midrule
\textbf{current} (default) & 0.945 & 0.895 & 0.751 & 0.833 & 0.019 & 0.041 \\
reduced\_boost ($1.5{\to}1.0$) & 0.945 & 0.895 & 0.751 & 0.833 & 0.019 & 0.041 \\
thresh $0.25/0.75$ & 0.946 & 0.861 & 0.649 & 0.746 & 0.017 & 0.039 \\
thresh $0.33/0.67$ & 0.945 & 0.896 & 0.752 & 0.835 & 0.019 & 0.041 \\
$\tau_m{=}0.4$ & 0.913 & 0.851 & 0.663 & 0.769 & 0.025 & 0.064 \\
$\tau_m{=}0.6$ & 0.953 & 0.908 & 0.779 & 0.850 & 0.016 & 0.036 \\
\textit{score\_only} (degenerate) & 0.998 & 0.995 & 0.989 & 0.999 & 0.006 & 0.001 \\
\bottomrule
\end{tabular}
\end{table*}

\paragraph{Findings.}
(i) \emph{Vote weights are inert.} Reducing the strong-mistake boost $1.5{\to}1.0$ (\textsf{reduced\_boost}) changes $0\%$ of labels and leaves every prediction metric identical to the default. In this $\rho\approx4\%$ regime, a \textsc{Fail}-only vote arg-maxes to \textsc{Fail} regardless of magnitude.
(ii) \emph{Thresholds: locally stable.} A small change ($0.33/0.67$) is within $0.002$ F1\textsubscript{FAIL} of the default; a large symmetric rebinning ($0.25/0.75$) lowers F1\textsubscript{FAIL} by $\approx0.10$ (it reshuffles $17\%$ of labels, mostly \textsc{Master}$\leftrightarrow$\textsc{Unsure}, and thins the \textsc{Fail} class).
(iii) \emph{$\tau_m$: mild, with the default in the mid-range.} $\tau_m{=}0.6$ slightly \emph{outperforms} the default ($0.779$ vs.\ $0.751$ F1\textsubscript{FAIL}) and $\tau_m{=}0.4$ is worse; the chosen $\tau_m{=}0.5$ sits mid-range, away from the F1 optimum in this small sweep.
(iv) \emph{\textsf{score\_only} is degenerate.} Dropping mistakes makes prediction appear near-perfect (F1\textsubscript{FAIL} $0.99$). This is an artifact: because SAF is answer-level, every concept in a row then inherits that one question's score bin, so each row collapses to a \emph{single} label (mean distinct labels per row $1.00$ vs.\ $1.27$ with mistakes; row homogeneity $1.000$ vs.\ $0.948$) and a held-out cell is trivially copied from its row-mates. The mistake refinement is what gives concept cells concept-specific signal; its specific boost \emph{magnitude} does not matter.

\paragraph{Conclusion.} The SAF-eval deployment result is robust to the specific hand-picked magnitudes: the vote weights are inert, small threshold/$\tau_m$ changes move F1\textsubscript{FAIL} by $\le0.03$, and the default sits in the stable mid-range. We treat $0.3/0.7$, $\tau_m{=}0.5$, and $1.5/0.8/0.2$ as one reasonable fixed SAF-specific heuristic; the Stage-2 method consumes only the resulting discrete matrix $\mathbf{D}$.

\section{Hyperparameter Sensitivity and Runtime}\label{app:sensitivity_extra}

This appendix collects the sensitivity and runtime details behind the main technical-scope checks. The full $(\theta_0,\theta_{\max})$ grid and the per-configuration ablation are in Appendix~\ref{app:ablation}; the defaults ($0.01/0.8$) sit on a stable region of that grid.
When labeled validation data is available, the prior endpoints can be tuned by $k$-fold cross-validation with an edge-count penalty:
\[
\mathrm{CV}(\theta_0,\theta_{\max})
=\frac{1}{k}\sum_{f=1}^{k}\left(\bar{\ell}_f-\lambda |E_f|\right),
\]
where $\bar{\ell}_f$ is the average per-cell held-out log-likelihood on fold $f$, $|E_f|$ is the learned edge count, and $\lambda$ is a small sparsity constant.
The predictor evaluation's exact VE runs in $\approx 1.6$\,s on SAF-eval but is intractable on the dense $249$-node Eedi data-only DAG (Appendix~\ref{app:eedi}), where we fall back to direct-parent inference.

\subsection{BDeu ESS sensitivity}\label{app:alpha_sensitivity}

Table~\ref{tab:alpha_sensitivity} varies the BDeu equivalent sample size $\alpha$ over $\{0.1,0.5,1,5,10\}$ on \texttt{asia}/\texttt{child}/\texttt{alarm}, with the same KG-SoftMAP prior defaults and $10\%$ KG noise.
The result is stable at the two ends of the sparse regime: at $\rho{=}0.05$ every tested $\alpha$ gives the same Directed-F1, and at $\rho{=}0.4$ the largest spread is $0.013$.
At $\rho{=}0.2$, larger smoothing gives a modest lift on some networks, with maximum spread $0.087$.
Thus $\alpha{=}1$ is a fixed working default rather than a fragile tuning point; the qualitative separation from the data-only floor is unchanged across the sweep.

\begin{table*}[t]
\centering
\small
\caption{BDeu equivalent-sample-size sensitivity for KG-SoftMAP (Directed-F1, mean over 5 seeds; $10\%$ KG noise). ``Range'' and ``spread'' are computed over $\alpha\in\{0.1,0.5,1,5,10\}$.}
\label{tab:alpha_sensitivity}
\begin{tabular}{llccc}
\toprule
Net & $\rho$ & DF1 at $\alpha{=}1$ & Range & Spread \\
\midrule
\texttt{asia}  & 0.05 & 0.278 & 0.278--0.278 & 0.000 \\
\texttt{child} & 0.05 & 0.138 & 0.138--0.138 & 0.000 \\
\texttt{alarm} & 0.05 & 0.208 & 0.208--0.208 & 0.000 \\
\midrule
\texttt{asia}  & 0.20 & 0.680 & 0.665--0.752 & 0.087 \\
\texttt{child} & 0.20 & 0.450 & 0.424--0.485 & 0.061 \\
\texttt{alarm} & 0.20 & 0.488 & 0.477--0.520 & 0.043 \\
\midrule
\texttt{asia}  & 0.40 & 0.963 & 0.963--0.976 & 0.013 \\
\texttt{child} & 0.40 & 0.946 & 0.946--0.946 & 0.000 \\
\texttt{alarm} & 0.40 & 0.936 & 0.936--0.938 & 0.002 \\
\bottomrule
\end{tabular}
\end{table*}

\subsection{Runtime scaling}\label{app:scaling}

Table~\ref{tab:scaling} reports wall-clock time for the greedy MAP structure search, excluding KG construction and data preprocessing.
Time grows roughly quadratically in $p$ under the max-in-degree-4 constraint, remaining under $15$ seconds for all tested configurations including SAF-eval ($p{=}191$, $\rho{\approx}4.4\%$).
The random 100-node and 200-node entries use synthetic sparse matrices to fill the gap between bnlearn benchmarks and SAF.
These timings establish that the search is practical for moderate $p$; scaling to $p{>}500$ would likely require candidate pruning or decomposition and is untested.

\begin{table*}[t]
\centering
\small
\caption{Runtime scaling for the greedy MAP structure search. Times exclude KG construction, data preprocessing, and CPD fitting. bnlearn-network rows are means over two seeds; random and SAF-eval rows are single runs. The synthetic random rows fill the scale gap between the standard benchmarks and SAF-eval.}
\label{tab:scaling}
\begin{tabular}{lrrrrr}
\toprule
Dataset & $p$ & $N$ & $\rho$ & KG edges & Search time (s) \\
\midrule
\texttt{cancer} & 5 & 1000 & 0.393 & 5 & 0.024 \\
\texttt{asia} & 8 & 1000 & 0.393 & 9 & 0.048 \\
\texttt{sachs} & 11 & 1000 & 0.396 & 18 & 0.106 \\
\texttt{child} & 20 & 1000 & 0.398 & 27 & 0.160 \\
\texttt{insurance} & 27 & 1000 & 0.399 & 56.5 & 0.378 \\
\texttt{alarm} & 37 & 1000 & 0.400 & 50 & 0.366 \\
\texttt{random\_100} & 100 & 800 & 0.050 & 106 & 1.383 \\
\texttt{random\_200} & 200 & 800 & 0.050 & 209 & 5.182 \\
SAF-eval & 191 & 1392 & 0.044 & 159 & 13.613 \\
\bottomrule
\end{tabular}
\end{table*}

\section{Same-KG Mechanism Ablation: Full Grid}\label{app:fairkg}

Table~\ref{tab:app_fairkg} reports the per-network values behind the main-text mechanism check. It separates the effect of the KG prior from the effect of the search restriction: all ``+Prior'' methods receive the same imperfect KG, while the search procedure changes.

\begin{table*}[t]
\centering
\caption{Same-KG mechanism ablation full grid (Directed-F1, mean$\pm$SD over 5 seeds; $\rho{=}0.4$). All ``+Prior'' methods receive the same imperfect KG. MMHC-cand+Prior uses MMHC-style candidate restriction followed by the same sparse BDeu+logit local search.}
\label{tab:app_fairkg}
\begin{tabular}{llccccc}
\toprule
Net & noise & GES & MMHC & GES+Prior & MMHC-cand+Prior & KG-SoftMAP \\
\midrule
\texttt{cancer} & 0.1 & 0.07$\pm$0.15 & 0.07$\pm$0.15 & 0.30$\pm$0.01 & 0.55$\pm$0.08 & 0.55$\pm$0.08 \\
\texttt{cancer} & 0.3 & 0.07$\pm$0.15 & 0.07$\pm$0.15 & 0.30$\pm$0.01 & 0.55$\pm$0.08 & 0.55$\pm$0.08 \\
\addlinespace[1pt]
\texttt{asia} & 0.1 & 0.08$\pm$0.12 & 0.08$\pm$0.12 & 0.47$\pm$0.10 & 0.65$\pm$0.05 & 0.65$\pm$0.05 \\
\texttt{asia} & 0.3 & 0.08$\pm$0.12 & 0.08$\pm$0.12 & 0.47$\pm$0.10 & 0.65$\pm$0.06 & 0.65$\pm$0.06 \\
\addlinespace[1pt]
\texttt{sachs} & 0.1 & 0.00$\pm$0.00 & 0.00$\pm$0.00 & 0.37$\pm$0.10 & 0.59$\pm$0.07 & 0.59$\pm$0.07 \\
\texttt{sachs} & 0.3 & 0.00$\pm$0.00 & 0.00$\pm$0.00 & 0.37$\pm$0.10 & 0.58$\pm$0.06 & 0.58$\pm$0.06 \\
\addlinespace[1pt]
\texttt{child} & 0.1 & 0.00$\pm$0.00 & 0.00$\pm$0.00 & 0.26$\pm$0.05 & 0.49$\pm$0.05 & 0.49$\pm$0.05 \\
\texttt{child} & 0.3 & 0.00$\pm$0.00 & 0.00$\pm$0.00 & 0.26$\pm$0.05 & 0.56$\pm$0.13 & 0.56$\pm$0.13 \\
\addlinespace[1pt]
\texttt{insurance} & 0.1 & 0.00$\pm$0.00 & 0.00$\pm$0.00 & 0.23$\pm$0.02 & 0.51$\pm$0.11 & 0.46$\pm$0.05 \\
\texttt{insurance} & 0.3 & 0.00$\pm$0.00 & 0.00$\pm$0.00 & 0.22$\pm$0.02 & 0.66$\pm$0.07 & 0.62$\pm$0.08 \\
\addlinespace[1pt]
\texttt{alarm} & 0.1 & 0.00$\pm$0.00 & 0.00$\pm$0.00 & 0.35$\pm$0.04 & 0.50$\pm$0.03 & 0.49$\pm$0.03 \\
\texttt{alarm} & 0.3 & 0.00$\pm$0.00 & 0.00$\pm$0.00 & 0.35$\pm$0.04 & 0.51$\pm$0.04 & 0.51$\pm$0.05 \\
\bottomrule
\end{tabular}
\end{table*}

\section{Third-Party Soft-Prior Baseline: Full Grid and Protocol}\label{app:softprior}

\paragraph{Protocol.} We give \textsc{bnlearn}~\citep{scutari2014bayesian} the \emph{same} masked discrete data and the \emph{same} KG as KG-SoftMAP, with the same $\theta_0/\theta_{\max}$ endpoints, through its inclusion-driven \texttt{cs} soft arc prior \citep{castelo2003inclusion}. Because the \texttt{cs} prior consumes inclusion \emph{probabilities} directly, we instantiate them by linear interpolation, $\mathrm{prob}(u\!\to\!v)=\theta_0+(\theta_{\max}-\theta_0)\,w_{uv}$ for KG arcs and $\theta_0$ for every other ordered pair ($\theta_0{=}0.01$, $\theta_{\max}{=}0.8$); this agrees with KG-SoftMAP's logit-linear map at the endpoints $w_{uv}{\in}\{0,1\}$ and interpolates linearly in probability between them. \textsc{bnlearn} additionally shrinks probabilities away from $0$ and $1$. KG edges enter as finite inclusion probabilities, so all edge decisions remain revisable by the learner. Because \textsc{bnlearn}'s hill-climber requires complete data, it uses its own missing-data handling, in two variants:
\begin{itemize}
\item \texttt{bnlearn-StructEM+cs} (native): \texttt{structural.em(data,} \texttt{maximize="hc",} \texttt{maximize.args=list(score="bde",} \texttt{prior="cs", beta=beta),} \texttt{fit="bayes")};
\item \texttt{bnlearn-impute+cs} (fallback): mode-impute the missing entries, then \texttt{hc(data, score="bde", prior="cs", beta=beta)}.
\end{itemize}
KG-SoftMAP and MMHC-cand+Prior use per-local complete-case BDeu directly on the sparse matrix. The two \textsc{bnlearn} variants agree closely (Table~\ref{tab:app_softprior}), and the remaining margin in favour of KG-SoftMAP reflects how the learners handle the sparse-data regime. Learned DAGs are scored with the same SHD / Directed-F1 code as every other experiment.

\begin{table*}[t]
\centering
\caption{Full third-party soft-prior grid (Directed-F1, mean$\pm$SD over 5 seeds; $\rho{=}0.4$). All methods receive the same KG and matched $\theta_0/\theta_{\max}$ endpoint probabilities; \textsc{bnlearn}+\texttt{cs} interpolates linearly in probability because its prior consumes inclusion probabilities. The table shows that the same KG signal transfers to a third-party soft-prior learner; differences between rows also reflect each learner's missing-data handling and prior interface.}
\label{tab:app_softprior}
\begin{tabular}{llcccc}
\toprule
Net & noise & KG-SoftMAP & MMHC-cand+Prior & bnlearn-StructEM+cs & bnlearn-impute+cs \\
\midrule
\texttt{cancer} & 0.1 & 0.55$\pm$0.08 & 0.55$\pm$0.08 & 0.47$\pm$0.13 & 0.47$\pm$0.13 \\
\texttt{cancer} & 0.3 & 0.55$\pm$0.08 & 0.55$\pm$0.08 & 0.47$\pm$0.13 & 0.47$\pm$0.13 \\
\addlinespace[1pt]
\texttt{asia} & 0.1 & 0.65$\pm$0.05 & 0.65$\pm$0.05 & 0.79$\pm$0.14 & 0.81$\pm$0.12 \\
\texttt{asia} & 0.3 & 0.65$\pm$0.06 & 0.65$\pm$0.06 & 0.79$\pm$0.14 & 0.81$\pm$0.12 \\
\addlinespace[1pt]
\texttt{sachs} & 0.1 & 0.59$\pm$0.07 & 0.59$\pm$0.07 & 0.53$\pm$0.05 & 0.54$\pm$0.03 \\
\texttt{sachs} & 0.3 & 0.58$\pm$0.06 & 0.58$\pm$0.06 & 0.53$\pm$0.06 & 0.54$\pm$0.04 \\
\addlinespace[1pt]
\texttt{child} & 0.1 & 0.49$\pm$0.05 & 0.49$\pm$0.05 & 0.67$\pm$0.07 & 0.66$\pm$0.05 \\
\texttt{child} & 0.3 & 0.56$\pm$0.13 & 0.56$\pm$0.13 & 0.67$\pm$0.07 & 0.66$\pm$0.05 \\
\addlinespace[1pt]
\texttt{insurance} & 0.1 & 0.46$\pm$0.05 & 0.51$\pm$0.11 & 0.50$\pm$0.05 & 0.47$\pm$0.04 \\
\texttt{insurance} & 0.3 & 0.62$\pm$0.08 & 0.66$\pm$0.07 & 0.50$\pm$0.05 & 0.47$\pm$0.04 \\
\addlinespace[1pt]
\texttt{alarm} & 0.1 & 0.49$\pm$0.03 & 0.50$\pm$0.03 & 0.63$\pm$0.07 & 0.56$\pm$0.03 \\
\texttt{alarm} & 0.3 & 0.51$\pm$0.05 & 0.51$\pm$0.04 & 0.64$\pm$0.07 & 0.56$\pm$0.03 \\
\bottomrule
\end{tabular}
\end{table*}

\section{Adaptive Equivalent Sample Size: Dormant Under the Default Gate}\label{app:ess}

\paragraph{Definition.} The adaptive-ESS guard is defined in Eq.~\ref{eq:adaptive_ess}: it scales the local ESS upward only when a scored family has fewer complete cases than parent configurations. It is a numerical safeguard for very small local counts, not a replacement for the standard BDeu marginal likelihood.

\paragraph{Why it is dormant.} The structure search scores a local family only when its complete-case count $n_j$ satisfies $n_j \ge \max(\texttt{min\_obs\_base},\, \lceil \texttt{min\_obs\_per\_config}\cdot q\rceil)$. With the defaults ($\texttt{min\_obs\_base}{=}5$, $\texttt{min\_obs\_per\_config}{=}1$) this is $n_j \ge \max(5,q) \ge q$, so $q/\max(n_j,1)\le 1$, the clamped factor $\min(C,\max(1,q/\max(n_j,1)))$ equals~1, and $\alpha_{\mathrm{eff}}{=}\alpha$ for \emph{every} scored family. The guard can engage only under relaxed gates that admit families with $n_j<q$.

\paragraph{Ablation.} We toggle the guard on ($C{=}5$, default) versus off ($C{=}1$, forcing $\alpha_{\mathrm{eff}}{=}\alpha$). Under the default gate, all reported outputs are unchanged: across all $45$ synthetic cells (asia/child/alarm $\times$ $\rho\in\{0.05,0.2,0.4\}$ $\times$ 5 seeds) the Directed-F1 and SHD are identical, the learned SAF-eval graph is identical ($194$ edges, symmetric difference $0$), and the trigger never fires. Under deliberately relaxed gates the trigger \emph{does} fire ($8$--$31\%$ of scored families, with cap-hits up to $5\%$), confirming the mechanism is live rather than dead code; even then the learned structure is unchanged in the regimes tested (Table~\ref{tab:app_ess}).

\begin{table*}[t]
\centering
\caption{Adaptive-ESS on/off ablation. ``active'' = scored families with scaling factor $>1$; ``cap-hit'' = families at the cap $C$. $\Delta$DF1/$\Delta$SHD = on $-$ off. Under the default gate the guard is dormant ($0\%$ trigger, identical results); under relaxed gates it fires but does not change the learned structure here. SAF-eval has no ground-truth DAG, so only graph identity is reported.}
\label{tab:app_ess}
\begin{tabular}{lcccc}
\toprule
Setting (gate: base,\,per-cfg) & $\Delta$DF1 & $\Delta$SHD & active & cap-hit \\
\midrule
Default, all synthetic ($45$ cells) & $0.000$ & $0.0$ & $0\%$ & $0\%$ \\
Default, SAF (graph identical)       & ---     & ---   & $0\%$ & $0\%$ \\
\texttt{asia} $\rho{=}0.05$ ($5,\,0.5$) & $0.000$ & $0.0$ & $0\%$ & $0\%$ \\
\texttt{asia} $\rho{=}0.05$ ($1,\,0.5$) & $0.000$ & $0.0$ & $8\%$ & $0\%$ \\
\texttt{child} $\rho{=}0.2$ ($1,\,0.1$) & $0.000$ & $0.0$ & $31\%$ & $5\%$ \\
\bottomrule
\end{tabular}
\end{table*}

\section{Mis-specified and Truth-Independent KGs: Full Results}\label{app:misspec}

\paragraph{KG construction.} All KGs use high-confidence weights $w_{uv}\sim\mathrm{Uniform}(0.7,1.0)$ and are evaluated at $\rho{=}0.4$ over 5 seeds on \texttt{child}/\texttt{insurance}/\texttt{sachs}. We vary false-edge additions at $10/30/50\%$ and apply targeted $30\%$ drop, reverse, and mixed corruptions. The \texttt{random\_disjoint} KG is sampled separately from the true DAG while excluding every true directed edge, giving zero correct directed signal.

\begin{table*}[t]
\centering
\caption{Mis-specified / truth-independent KG: aggregate Directed-F1 and SHD (mean over network-level means; 5 seeds, $\rho{=}0.4$). False additions are relatively benign because true-edge recall is preserved; dropping or reversing true edges is more damaging. The zero-signal \texttt{random\_disjoint} KG returns to the no-prior regime.}
\label{tab:app_misspec}
\begin{tabular}{lccc}
\toprule
KG condition & KG precision & Directed-F1 & SHD \\
\midrule
clean true KG & 1.00 & 0.497 & 41.3 \\
false edges 10\% & 0.93 & 0.492 & 41.3 \\
false edges 30\% & 0.79 & 0.540 & 35.0 \\
false edges 50\% & 0.69 & 0.595 & 28.3 \\
add-false 30\% & 0.77 & 0.611 & 29.3 \\
drop 30\% & 1.00 & 0.412 & 45.4 \\
reverse 30\% & 0.69 & 0.366 & 42.7 \\
mixed 30\% & 0.59 & 0.410 & 40.0 \\
\midrule
\texttt{random\_disjoint} & 0.00 & 0.181 & 57.3 \\
Ours-NoPrior floor & --- & 0.191 & --- \\
\bottomrule
\end{tabular}
\end{table*}

\paragraph{KG density mismatch.}
Table~\ref{tab:app_density_mismatch} broadens the corruption test to all six synthetic networks and changes the KG density itself.
Sparse KGs retain only a fraction of the true edges and add no false edges; dense KGs keep all true edges and add high-confidence false directed edges at a multiple of the true edge count.
Recovery increases with KG recall in the sparse-KG cases. Dense KGs remain useful when recall is complete, but precision controls how much extra structure the learner must filter. The $50\%$-recall sparse KG reaches mean Directed-F1 $0.34$, while the $2\times$-density KG with precision $0.50$ reaches $0.58$, above the no-prior floor of $0.21$.

\begin{table*}[t]
\centering
\small
\caption{KG density mismatch on all six synthetic networks (mean over network-level means; $\rho{=}0.4$, 5 seeds). Density is KG edge count divided by true edge count. Sparse KGs drop true edges; dense KGs keep all true edges and add high-confidence false edges.}
\label{tab:app_density_mismatch}
\begin{tabular}{lccccc}
\toprule
KG condition & Density & Precision & Recall & DF1 & SHD \\
\midrule
No prior & 0.00 & --- & 0.00 & 0.206 & 52.87 \\
Sparse: keep $25\%$ true & 0.25 & 1.00 & 0.25 & 0.266 & 48.23 \\
Sparse: keep $50\%$ true & 0.49 & 1.00 & 0.49 & 0.336 & 44.60 \\
Sparse: keep $75\%$ true & 0.75 & 1.00 & 0.75 & 0.420 & 40.23 \\
Clean true KG & 1.00 & 1.00 & 1.00 & 0.536 & 34.67 \\
Dense: add $0.5{\times}$ false & 1.49 & 0.67 & 1.00 & 0.557 & 29.07 \\
Dense: add $1{\times}$ false & 2.00 & 0.50 & 1.00 & 0.579 & 26.50 \\
Dense: add $2{\times}$ false & 3.00 & 0.33 & 1.00 & 0.469 & 30.67 \\
Dense: add $5{\times}$ false & 5.83 & 0.17 & 1.00 & 0.311 & 40.97 \\
\bottomrule
\end{tabular}
\end{table*}

\section{Standard-Benchmark LLM-Extracted KG: Protocol}\label{app:llmkg}

This appendix details the blind elicitation used as the standard-benchmark sanity check in the main LLM-KG section. It tests whether a realistically imperfect, LLM-extracted KG recovers structure on networks with a \emph{known} ground-truth DAG.

\paragraph{Blind elicitation.} For each of \texttt{asia}, \texttt{sachs}, and \texttt{child}, the variables were presented to an independent language-model agent under \emph{neutral identifiers} (\texttt{V1..Vn} / \texttt{M1..Mn} / \texttt{C1..Cn}) with one-line, paraphrased descriptions of each variable's real-world meaning. The benchmark name, the variable names, and the true edges were \emph{withheld}; the agent was instructed to produce a weighted directed dependency graph from domain knowledge alone and to avoid assuming any specific published dataset. Each agent returned a JSON edge list with a confidence in $[0,1]$ per edge, which we map back to node names and use directly as the weighted KG $\mathbf{K}$ (with $w_{uv}$ the returned confidence). The prompt did not include the true DAG; the supplied information was the paraphrased variable semantics.

\paragraph{Measured KG quality.} We score each elicited KG by directed precision/recall against the true DAG (Table~\ref{tab:app_llmkg_quality}). Two of the three KGs are clearly imperfect (\texttt{sachs}, \texttt{child}: precision $0.54$--$0.62$, with reversed and spurious edges), so the elicited graph differs from the benchmark DAG in those cases. The small, canonical \texttt{asia} domain is the favorable extreme, recovered exactly. KG quality is therefore an \emph{observed, imperfect input} to the recovery experiment.

\begin{table*}[t]
\centering
\caption{Directed quality of the blind LLM-elicited KGs vs.\ the true DAG. Elicited = number of directed edges returned; Correct = directed edges matching the true DAG; Reversed = elicited edges whose reverse is a true edge.}
\label{tab:app_llmkg_quality}
\small
\begin{tabular}{lcccccc}
\toprule
Net & True edges & Elicited & Correct & Reversed & Prec. & Rec. \\
\midrule
\texttt{asia}  & 8  & 8  & 8  & 0 & 1.00 & 1.00 \\
\texttt{sachs} & 17 & 21 & 13 & 3 & 0.62 & 0.76 \\
\texttt{child} & 25 & 37 & 20 & 1 & 0.54 & 0.80 \\
\bottomrule
\end{tabular}
\end{table*}

\paragraph{Recovery and reproducibility.} The elicited KG is fed through the same KG-SoftMAP pipeline as the main recovery table (defaults $\theta_0{=}0.01$, $\theta_{\max}{=}0.8$, $\alpha{=}1.0$, max in-degree $4$), at $\rho\in\{0.05,0.20,0.40\}$ over $5$ seeds, against the data-only floor (\textsc{Ours-NoPrior}) and the true-DAG-derived synthetic KG. Results are in Table~\ref{tab:app_llmkg_recovery}. The verbatim elicited edge lists, the neutral-ID--to--node maps, and the runner are in \texttt{exp\_llm\_kg\_recovery.py}. The recovery from the imperfect LLM KG sits below the synthetic KG and the gap scales with KG quality, consistent with the corruption sweep (Appendix~\ref{app:misspec}) and the central claim that recovery tracks KG quality.

\begin{table*}[t]
\centering
\caption{Recovery from an LLM-extracted KG on standard known-DAG benchmarks (Directed-F1, mean over 5 seeds). Floor = data-only \textsc{Ours-NoPrior}; Synth-KG = true-DAG-derived KG from the main recovery table, same pipeline.}
\label{tab:app_llmkg_recovery}
\small
\begin{tabular}{llccc}
\toprule
Net & $\rho$ & LLM-KG & Floor & Synth-KG \\
\midrule
\texttt{asia}  & 0.05 & 0.28 & 0.00 & 0.28 \\
\texttt{asia}  & 0.20 & 0.72 & 0.27 & 0.67 \\
\texttt{asia}  & 0.40 & \textbf{1.00} & 0.04 & 0.96 \\
\addlinespace[1pt]
\texttt{sachs} & 0.05 & 0.24 & 0.00 & 0.28 \\
\texttt{sachs} & 0.20 & 0.41 & 0.25 & 0.49 \\
\texttt{sachs} & 0.40 & 0.72 & 0.00 & 0.95 \\
\addlinespace[1pt]
\texttt{child} & 0.05 & 0.10 & 0.00 & 0.14 \\
\texttt{child} & 0.20 & 0.43 & 0.15 & 0.43 \\
\texttt{child} & 0.40 & 0.66 & 0.00 & 0.96 \\
\bottomrule
\end{tabular}
\end{table*}

\section{NovelGraphs LLM-Provenance Check}\label{app:novographs}

This appendix expands the main NovelGraphs table. NovelGraphs~\citep{srivastava2025sciencegrounded} provides known DAGs designed for evaluating LLM-based causal discovery outside the standard BN repository. We use the LLM-pairwise KG as KG-SoftMAP's LLM-derived prior, and evaluate sparse learners under $\rho\in\{0.05,0.20,0.40\}$ over five seeds.

\begin{table*}[t]
\centering
\caption{NovelGraphs LLM-only graph quality (Directed-F1). PromptBN is from \citet{zhang2025promptbn}; LLM-pairwise and LLM-BFS follow the NovelGraphs protocol. ReActBN is evaluated in Table~\ref{tab:app_novographs_sparse}.}
\label{tab:app_novographs_llmonly}
\small
\begin{tabular}{lcc}
\toprule
Method & \texttt{alzheimers} & \texttt{covid19-small} \\
\midrule
LLM-pairwise & 0.70 & 0.62 \\
LLM-BFS & 0.60 & 0.00 \\
PromptBN & 0.50 & 0.67 \\
\bottomrule
\end{tabular}
\end{table*}

\begin{table*}[t]
\centering
\caption{NovelGraphs sparse-learning grid (Directed-F1, mean$\pm$SD over 5 seeds). KG-SoftMAP+LLM-KG, MMHC-cand+Prior, and ILS-CSL-softHC receive the same LLM-pairwise KG; ReActBN uses its LLM-refinement procedure; data-only rows receive only the masked discrete matrix.}
\label{tab:app_novographs_sparse}
\small
\begin{tabular}{llccc}
\toprule
Dataset & Method & $\rho{=}0.05$ & $\rho{=}0.20$ & $\rho{=}0.40$ \\
\midrule
\texttt{alzheimers} & KG-SoftMAP+LLM-KG & 0.13$\pm$0.13 & 0.25$\pm$0.06 & 0.47$\pm$0.07 \\
\texttt{alzheimers} & MMHC-cand+Prior & 0.13$\pm$0.13 & 0.25$\pm$0.06 & 0.47$\pm$0.07 \\
\texttt{alzheimers} & ILS-CSL-softHC & 0.14$\pm$0.08 & 0.20$\pm$0.06 & 0.15$\pm$0.10 \\
\texttt{alzheimers} & ReActBN & 0.29$\pm$0.07 & 0.25$\pm$0.13 & 0.22$\pm$0.03 \\
\texttt{alzheimers} & BOSS-BDeu & 0.00$\pm$0.00 & 0.24$\pm$0.05 & 0.40$\pm$0.10 \\
\texttt{alzheimers} & GRaSP-BDeu & 0.00$\pm$0.00 & 0.27$\pm$0.09 & 0.36$\pm$0.06 \\
\texttt{alzheimers} & PC & 0.18$\pm$0.07 & 0.26$\pm$0.05 & 0.30$\pm$0.07 \\
\texttt{alzheimers} & GES & 0.05$\pm$0.05 & 0.24$\pm$0.12 & 0.35$\pm$0.06 \\
\texttt{alzheimers} & Sparse-Cand & 0.00$\pm$0.00 & 0.14$\pm$0.06 & 0.13$\pm$0.06 \\
\texttt{alzheimers} & Structural EM & 0.00$\pm$0.00 & 0.14$\pm$0.07 & 0.12$\pm$0.06 \\
\texttt{alzheimers} & A-dC & 0.00$\pm$0.00 & 0.14$\pm$0.07 & 0.10$\pm$0.11 \\
\texttt{alzheimers} & Ours-NoPrior & 0.00$\pm$0.00 & 0.14$\pm$0.03 & 0.00$\pm$0.00 \\
\texttt{alzheimers} & DAGMA & 0.00$\pm$0.00 & 0.00$\pm$0.00 & 0.04$\pm$0.05 \\
\texttt{alzheimers} & NOTEARS & 0.00$\pm$0.00 & 0.00$\pm$0.00 & 0.04$\pm$0.09 \\
\addlinespace[1pt]
\texttt{covid19-small} & KG-SoftMAP+LLM-KG & 0.22$\pm$0.09 & 0.45$\pm$0.10 & 0.56$\pm$0.04 \\
\texttt{covid19-small} & MMHC-cand+Prior & 0.22$\pm$0.09 & 0.44$\pm$0.11 & 0.54$\pm$0.05 \\
\texttt{covid19-small} & ILS-CSL-softHC & 0.17$\pm$0.06 & 0.31$\pm$0.09 & 0.34$\pm$0.14 \\
\texttt{covid19-small} & ReActBN & 0.36$\pm$0.10 & 0.46$\pm$0.05 & 0.54$\pm$0.07 \\
\texttt{covid19-small} & BOSS-BDeu & 0.02$\pm$0.04 & 0.36$\pm$0.07 & 0.57$\pm$0.05 \\
\texttt{covid19-small} & GRaSP-BDeu & 0.02$\pm$0.04 & 0.36$\pm$0.07 & 0.52$\pm$0.08 \\
\texttt{covid19-small} & PC & 0.09$\pm$0.07 & 0.31$\pm$0.08 & 0.36$\pm$0.06 \\
\texttt{covid19-small} & GES & 0.12$\pm$0.08 & 0.33$\pm$0.09 & 0.37$\pm$0.05 \\
\texttt{covid19-small} & Sparse-Cand & 0.00$\pm$0.00 & 0.27$\pm$0.07 & 0.30$\pm$0.12 \\
\texttt{covid19-small} & Structural EM & 0.00$\pm$0.00 & 0.26$\pm$0.10 & 0.37$\pm$0.05 \\
\texttt{covid19-small} & A-dC & 0.00$\pm$0.00 & 0.26$\pm$0.10 & 0.17$\pm$0.10 \\
\texttt{covid19-small} & Ours-NoPrior & 0.00$\pm$0.00 & 0.29$\pm$0.06 & 0.00$\pm$0.00 \\
\texttt{covid19-small} & DAGMA & 0.00$\pm$0.00 & 0.00$\pm$0.00 & 0.07$\pm$0.07 \\
\texttt{covid19-small} & NOTEARS & 0.00$\pm$0.00 & 0.00$\pm$0.00 & 0.07$\pm$0.10 \\
\bottomrule
\end{tabular}
\end{table*}

\bibliography{kg_softmap_refs}